\documentclass[12pt,oneside]{thesis}

\usepackage{lipsum}
\usepackage{graphicx}
\usepackage{amsmath}
\usepackage{amssymb}
\usepackage{multirow}
\usepackage{bm}
\usepackage{caption}
\usepackage{subcaption}

\usepackage{color}

\PassOptionsToPackage{hyphens}{url}\usepackage[hidelinks]{hyperref}

\begin{document}

\pagenumbering{roman} 
\title{Gaussian Process-Based\\Active Exploration Strategies\\in Vision and Touch}

\author{Ho Jin Choi}
\department{Mechanical Engineering and Applied Mechanics}

\degree{Master of Science}
\degreeyear{2024}

\supervisor{Nadia Figueroa}{Shalini and Rajeev Misra Presidential Assistant Professor of Mechanical Engineering and Applied Mechanics}

\chairman{Prashant K. Purohit}{Graduate Group Chair}

\maketitle

\section*{Acknowledgment}
I express my sincere gratitude to all individuals and organizations who helped building this thesis. Their support and encouragement have been invaluable, without which this work would not have been possible, nor as fulfilling.

I am deeply grateful to my supervisor, Dr. Nadia Figueroa, for her consistent guidance and feedback throughout the year-long process of preparing this work. Her insightful discussions and shared experiences have often provided the breakthrough moments needed to overcome challenges. Working alongside her has been a source of immense joy and progress. Furthermore, I extend my gratitude to my thesis committee members, Dr. Cynthia Sung and Dr. Michael Posa, for generously dedicating their time to discuss and provide invaluable feedback on the thesis. Their realistic and precise insights prompted me to reconsider my approach and make significant improvements, offering invaluable lessons in the research process. I also wish to acknowledge Dr. Ruzena Bajcsy for her insightful perspectives, particularly her pioneering work in the field of Active Perception. Special thanks for the 3D printed object, courtesy of the University of Pennsylvania Libraries' Holman Biotech Commons, and for the tools and space, courtesy of the University of Pennsylvania Libraries' Education Commons, which were essential for developing custom tactile sensors. I am grateful to Alec Lanter for sharing his invaluable experience from his Master's thesis, offering guidance even after his graduation. Additionally, I appreciate the contributions of Gregory Campbell and Jessica Yin from Dr. Mark Yim's lab, as well as Siming He from Dr. Pratik Chaudhari's lab, for their assistance and technical insights. 

My heartfelt thanks also go to my family for their unwavering love and support, which has been a constant source of strength throughout my life. Their perspectives outside of engineering have encouraged me to think beyond boundaries. I am also thankful to my colleagues in the lab for their willingness to engage in discussions, no matter how crazy my ideas may have seemed, and for their words of encouragement, which have kept me motivated.

\section*{Abstract}

Robots struggle to understand object properties like shape, material, and semantics due to limited prior knowledge, hindering manipulation in unstructured environments. In contrast, humans learn these properties through interactive multi-sensor exploration. This work proposes fusing visual and tactile observations into a unified Gaussian Process Distance Field (GPDF) representation for active perception of object properties. While primarily focusing on geometry, this approach also demonstrates potential for modeling surface properties beyond geometry.
The GPDF encodes signed distance using point cloud, analytic gradient and Hessian, and surface uncertainty estimates, which are attributes that common neural network shape representation lack. By utilizing a point cloud to construct a distance function, GPDF does not need extensive pretraining on large datasets and can incorporate observations by aggregation. Starting with an initial visual shape estimate, the framework iteratively refines the geometry by integrating dense vision measurements using differentiable rendering and tactile measurements at uncertain surface regions. By quantifying multi-sensor uncertainties, it plans exploratory motions to maximize information gain for recovering precise 3D structures. For the real-world robot experiment, we utilize the Franka Research 3 robot manipulator, which is fixed on a table and has a customized DIGIT tactile sensor and an Intel Realsense D435 RGBD camera mounted on the end-effector. In these experiments, the robot explores the shape and properties of objects assumed to be static and placed on the table. To improve scalability, we investigate approximation methods like inducing point method for Gaussian Processes. This probabilistic multi-modal fusion enables active exploration and mapping of complex object geometries, extending potentially beyond geometry.



\tableofcontents
\newpage
\listoftables
\newpage
\listoffigures

\pagenumbering{arabic} 
\chapter{Introduction}\label{sec:introduction}
\section{Motivation}\label{sec:motivation}

Accurate understanding of an object's properties, encompassing its shape, material, and semantics, is important for robotic manipulators to execute prehensile and non-prehensile manipulations successfully. For instance, consider a scenario where the robot possesses precise knowledge regarding the shape and friction coefficient of an object. In such cases, rather than relying solely on shape closure for grasping, the robot can use force closure, thereby enabling a wider range of grasping motion. In more complex situations, such as when the robot is aware of the temperature of the object's surface, it can strategically avoid grasping hot surfaces to prevent damage to its own hardware. Alternatively, it can intelligently handover the object so that humans do not grasp hot surfaces, ensuring their safety and comfort.

\begin{figure}[tb]
    \centering
    \includegraphics[width=0.35\linewidth]{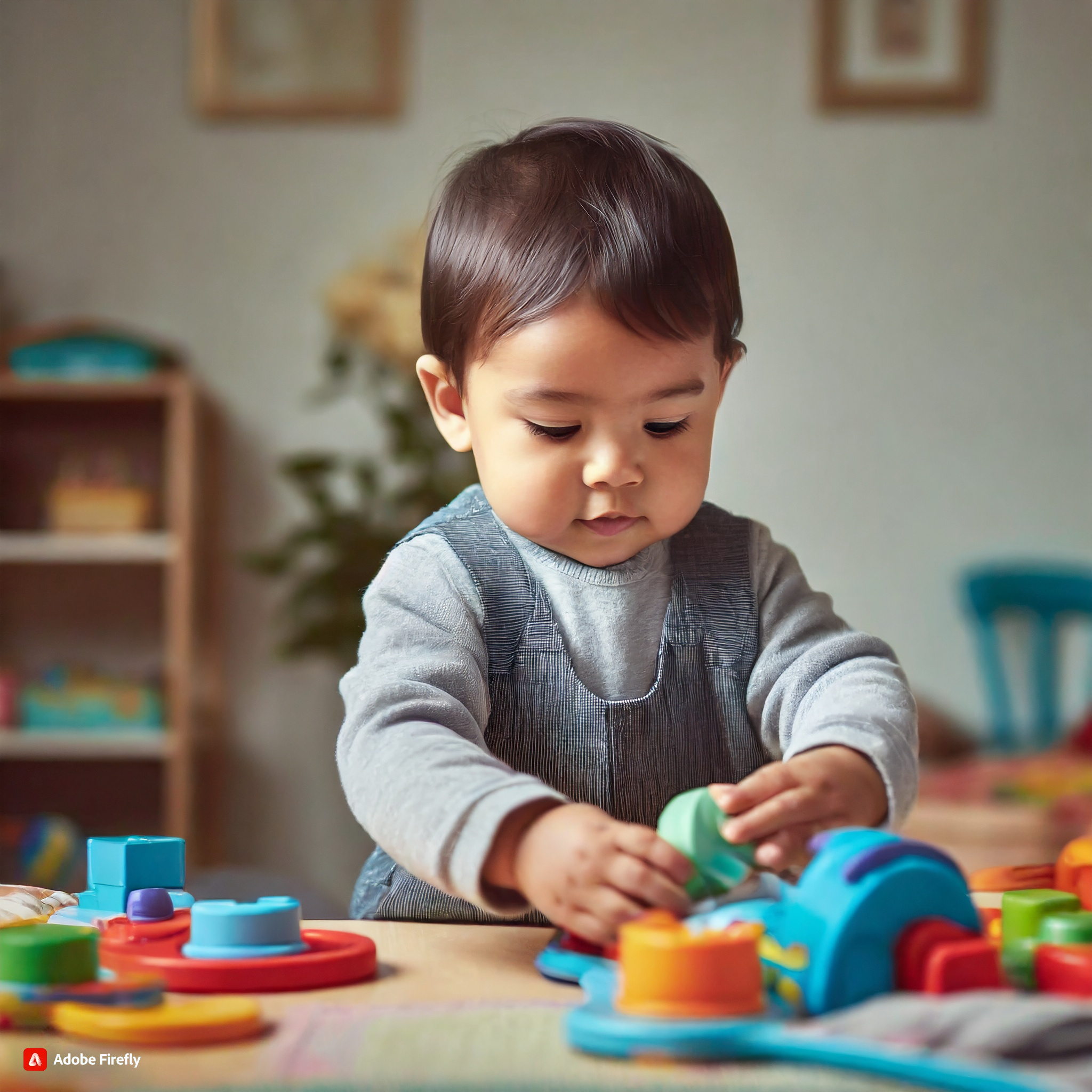}
    \caption{Babies enhance cognitive abilities through active exploration, relying on senses and statistical learning for pattern comprehension \cite{Smith2005TheDO}.}
    \label{fig:manipulative_toys}
\end{figure}
In real-world scenarios featuring complex and unstructured environments, robots often struggle with incomplete information about objects, posing a challenge for them to match the capabilities of a 3-year-old child \cite{Smith2005TheDO}. Unlike a child, who learns through interaction with the environment (see Figure \ref{fig:manipulative_toys}), robots encounter difficulties in adaptation to new environments or objects, even with pre-programmed knowledge. This underscores the importance of incorporating human-like active perception and exploration using multiple sensors, enabling robots to collect meaningful data points about their surroundings based on initial information \cite{DBLP:journals/corr/BajcsyAT16} and learn about them.

In robotics, active perception refers to ``an intelligent data acquisition process" \cite{Bajcsy1988ActiveP}, requiring a sensor model, an environmental model, and control strategies to maximize information gain while minimizing errors in the environmental model. Allen and Bajcsy \cite{Allen1985ObjectRU} provide an exemplary demonstration of active perception by combining passive vision and active tactile perception for object detection. They utilize binary images to detect object edges, initializing parametric spline surfaces, which are then refined using tactile sensors on a 6DOF PUMA robot to discern object holes and concavities. Subsequently, they infer objects by comparing the obtained surface model with a hierarchical model database containing multiple object models. This work successfully integrates various sensors and interdisciplinary techniques. We aim to address the following intriguing questions raised by this seminal work with state-of-the-art data-driven techniques:
\begin{itemize}
    \item How can we incorporate \textbf{\textit{visual structure}} beyond geometric structure?
    \item How can we model objects with \textbf{\textit{complex surfaces or scenes}}?
    \item How can we integrate \textbf{\textit{multiple camera views}}?
    \item How can we simulate human \textbf{\textit{tactile processing}}?
\end{itemize}
\begin{figure}[tb]
    \centering
    \includegraphics[width=1.0\textwidth]{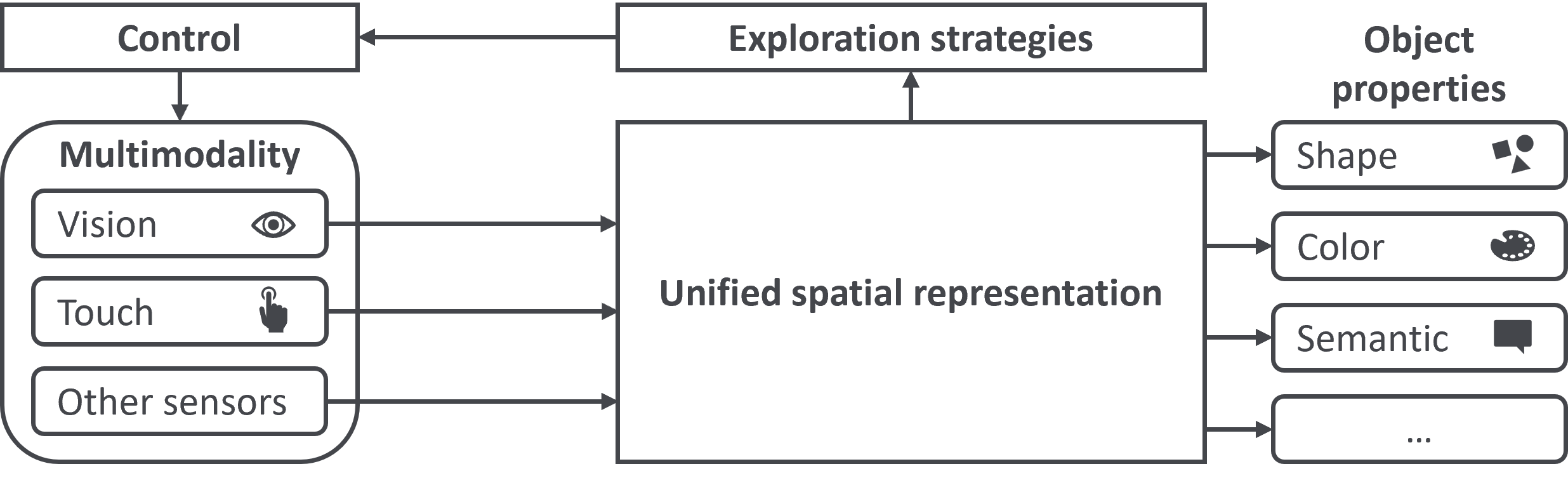}
    \caption{Active perception pipeline for different modalities and object properties}
    \label{fig:summary}
\end{figure}
In this work, we will dive into these questions and investigate how current researches try to answer them. In doing so, we propose a closed loop pipeline with unified world model based on Gaussian Process (GP) as in Figure \ref{fig:summary}, which integrates different modalities from a RGBD camera and a tactile sensor, facilitating the integration of different approaches and researches.
This work not only addresses the aforementioned questions \cite{Allen1985ObjectRU}, but also tackles two key requirements identified for integration:
\begin{enumerate}
    \item \textbf{Unified Spatial Data Representation:} The primary challenge lies in establishing a unified spatial data representation capable of \textbf{\textit{incremental updates}} based on new sensor observations and easily incorporating \textbf{\textit{multi-modal information}} such as shape, texture, semantics, and beyond. This task serves as the foundation for the other requirements, ensuring interoperability across diverse observations and exploration strategies.
    \item \textbf{Control Strategies for Different Modalities:} Sensor models for vision and tactile are different, as cameras are modeled with light rays or light transport, whereas tactile sensors can usually be modeled as point-based or vision-based tactile sensors. Therefore, they require \textbf{\textit{distinct control strategies}} and \textbf{\textit{quantification of information}} obtained from motion.
\end{enumerate}

In our research, we begin with actively reconstructing object shapes, following previous work's \cite{Allen1985ObjectRU} sequential approach from vision to tactile. This is a crucial step that provides valuable insights into geometry, which has relationship with contact points and object collisions. Unlike approaches \cite{DBLP:journals/corr/abs-1912-00280,DBLP:journals/corr/abs-1712-07262,DBLP:journals/corr/abs-1808-00671} that predict unseen parts of known object categories using partial point clouds, our work takes a similar path to works like Neural Radiance Field (NeRF) \cite{Mildenhall2020NeRFRS}, incremental Signed Distance Function (iSDF) \cite{Ortiz2022iSDFRN}, and Gaussian Splatting \cite{Kerbl20233DGS} that can incorporate multiple observations using differentiable rendering, which can avoid object class assumptions for generality and segmentation problem.

\section{Problem formulation}
\begin{figure}[tb]
    \centering
    \includegraphics[width=0.8\textwidth]{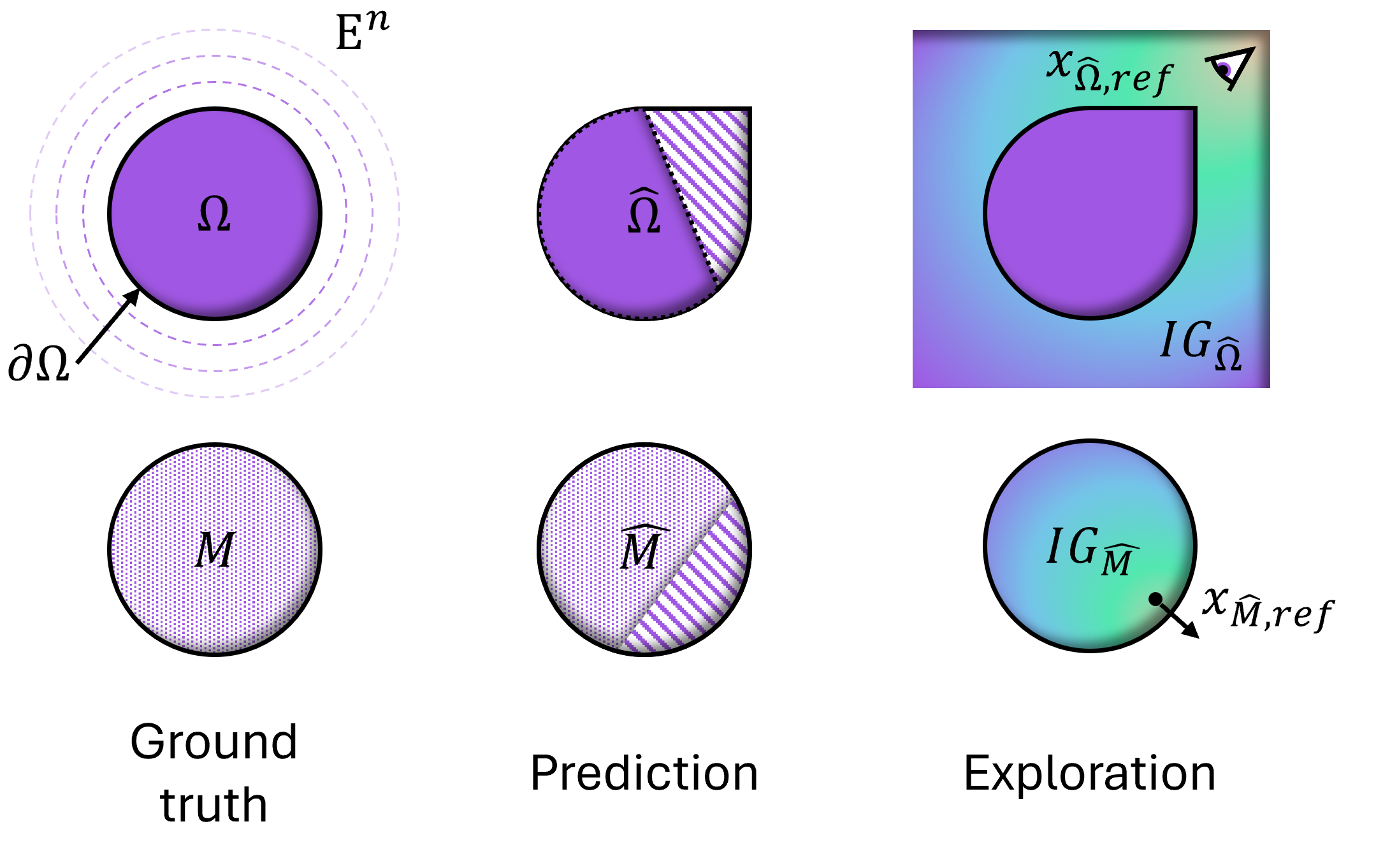}
    \caption{An object or scene, illustrated as a sphere, is a set, $\Omega$, in n-dimensional Euclidean space. Object surface properties illustrated as $M$. The symbol $\hat{\cdot}$ denotes the prediction of the property, and the image on the right illustrates how information gain guides the exploration.}
    \label{fig:environment_model}
\end{figure}
In our pursuit of solving geometry reconstruction challenges, we adopt a combined sequential approach involving both visual and tactile exploration, inspired by previous efforts.
Our methodology operates under the assumption of a predefined workspace or 3D search space for exploration. We also assume comprehensive information about the robot, including its geometry, joint configurations, and kinematic details for both the camera and tactile sensor. Currently, our focus lies on fixed, rigid objects within the reachable range of the robot. The object or scene shape, denoted as $\Omega$ and representing a set in Euclidean space as in Figure \ref{fig:environment_model}, along with its surface, $\partial\Omega$, and surface properties $M$, are partially or fully unknowns that the robot tries to explore given initial observation. Based on the estimation of shape, $\hat{\Omega}$, or material, $\hat{M}$, we calculate the information gain, $IG_{\hat{\cdot}}(x)$, which guides the exploration strategy to determine the optimal camera pose or point of contact, denoted as $x_{ref}$, for maximizing the information gain from the sensor. Subsequently, the controller guides the robot to actually reach the pose $x_{\hat{\cdot},ref}$. 



\section{Background}
We explore how sensor observations can be approximated and utilized for active perception. We delve into previous research on both visual and tactile exploration, which are fundamental for understanding this work.

\subsection{Camera modeling and modern surface reconstruction}
The widespread availability of commercial personal digital cameras led to extensive studies on camera models and perception through vision, making it a significant topic of research in robotics. Recent advancements in differentiable rendering have spurred researches into surface reconstruction tasks, notably in the context of Neural Radiance Fields (NERF) \cite{Mildenhall2020NeRFRS} and Gaussian Splatting \cite{Kerbl20233DGS}. Before getting in depth how those works, we first get into the modelling of the camera system, which is fundamental for most differentiable rendering.

\begin{figure}[tb]
    \begin{minipage}[c]{0.48\linewidth}
        \centering
        \includegraphics[width=0.7\linewidth]{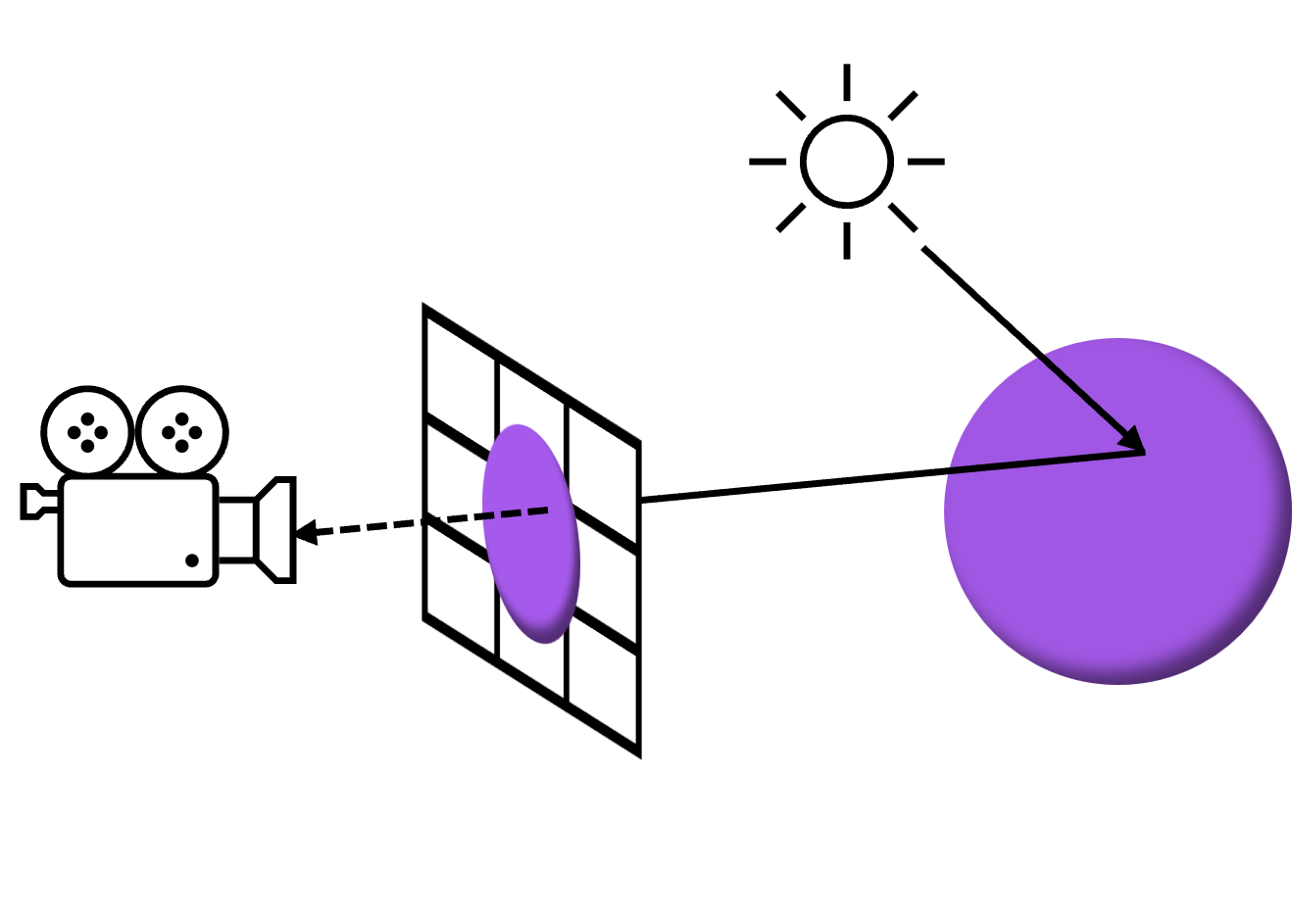}
        \caption{Illustration of real camera model where light ray comes into the lens}
        \label{fig:real_camera_model}
    \end{minipage}
    \hfill
    \begin{minipage}[c]{0.48\linewidth}
        \centering
        \includegraphics[width=0.8\linewidth]{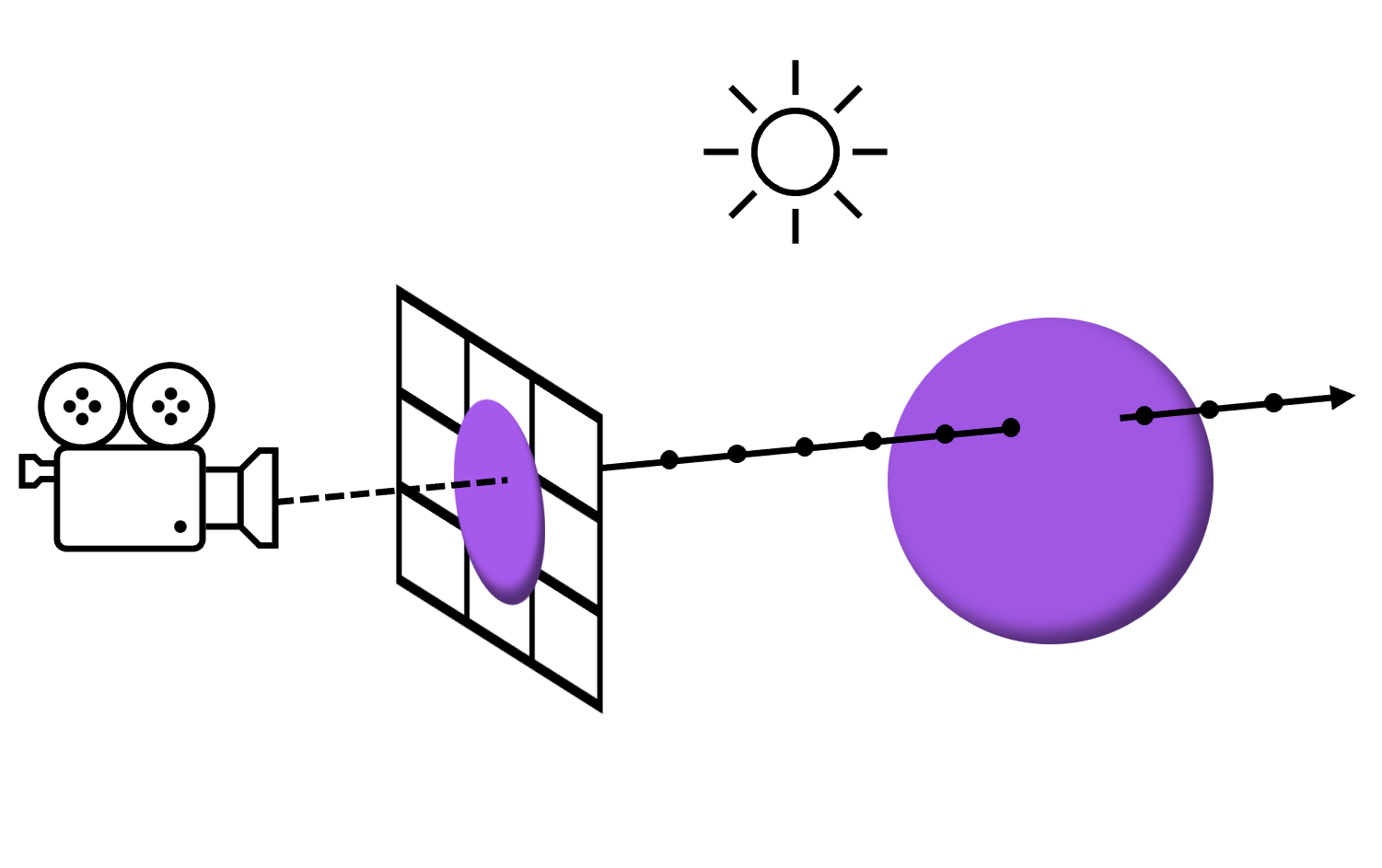}
        \caption{Illustration of ray casting for NERF \cite{Mildenhall2020NeRFRS} rendering where ray comes out from the camera and discrete samples along the ray}
        \label{fig:nerf_camera_model}
    \end{minipage}%
\end{figure}

In camera reconstruction, we consider intrinsic and extrinsic parameters, as well as skewness and distortions, as modeling factors for the camera. Intrinsic parameters determine pixel center and scale, while extrinsic parameters define the camera's 3D position. Skewness adjusts for errors in the UV axis, and distortion corrects lens issues. If there's no distortion or skewness, which is common in rectilinear projection used in commercial robotics cameras, the equations for converting 2D image to 3D world coordinates become simpler, which is known to be a pin-hole camera model:
\begin{equation}
    s
    \begin{bmatrix}
        u \\ v \\ 1
    \end{bmatrix}=
    \begin{bmatrix}
        f_x & 0 & c_x \\
        0 & f_y & c_y \\
        0 & 0 & 1 \\
    \end{bmatrix}
    \begin{bmatrix}
        R & t
    \end{bmatrix}
    \begin{bmatrix}
        x \\ y \\ z \\ 1
    \end{bmatrix}
\end{equation}
\begin{itemize}
    \item $f_x$ and $f_y$ are the focal lengths of the camera lens.
    \item $c_x$ and $c_y$ are the principal points.
    \item $R$ and $t$ represent the rotation and translation of the camera's position.
    \item $x$, $y$, and $z$ are coordinates of a point in the world.
    \item $u$ and $v$ are the resulting pixel coordinates on the camera sensor.
    \item $s$ is a scale factor.
\end{itemize}
Intrinsic parameters are usually given from the camera manufacturers and extrinsic parameters can be obtained from hand-eye calibration \cite{Daniilidis1999HandEyeCU}, assuming that the world origin is the same as the robot's origin. The scale factor is crucial but can't be directly determined from a single image without additional semantic information. However, in stereo vision systems, it can be inferred. This camera model is essential for understanding how light rays in a scene are captured by the camera as in Figure \ref{fig:real_camera_model} or projected from the camera in rendering techniques like ray casting as in Figure \ref{fig:nerf_camera_model}, which is an effective approximation.

NeRF \cite{Mildenhall2020NeRFRS} uses the ray casting technique with multiple discrete samples along the ray. Essentially, it measures the occupancy or opacity, as well as the color, of each discrete point along the ray and aggregates their color information based on occupancy to produce a single color output for each pixel. Assuming we have equidistant points on the ray sequentially, defined as $\mathbf{x}_{t_i}=\mathbf{r}_0+\mathbf{r}_d t_i$, where $t_{min}\leq t \leq t_{max}$ and $i$ is the index of points incrementally increasing, we obtain: 
\begin{equation}
\begin{split}
    & \alpha_i = \alpha(\mathbf{x}_{t_i}) \quad\text{(Opacity)}\\
    & c_i = c(\mathbf{x}_{t_i}) \quad\text{(Color)}\\
    & T_i = \Pi_{j=1}^{i-1}(1-\alpha_i) \quad\text{(Transmittance)}\\
    & \hat{C} = \sum_{i=1}^n T_i \alpha_i c_i \quad\text{(Estimated pixel color)} \\
\end{split}
\label{eq:nerf}
\end{equation}
where $n$ is number of point on the ray. This makes the rendering process differentiable, offering a way to optimize the occupancy function, $\alpha$, and the color function, $c$.

In contrast to NeRF \cite{Mildenhall2020NeRFRS}, Gaussian Splatting \cite{Kerbl20233DGS} does not employ ray casting. Instead, it populates numerous 3D Gaussian primitives and rasterizes them directly onto an image plane. This method uses the fact that the projected 3D Gaussians maintain Gaussian properties in 2D. By leveraging the probabilistic nature of the Gaussian distribution as opacity information, Gaussian Splatting achieves smooth gradients for translations and rotations of Gaussians by calculating its effect on the pixel. Additionally, it incorporates color and spherical harmonics for each Gaussian ellipsoids, resulting in a realistic camera view.

Gaussian Splatting offers an explicit representation of the scene, facilitating manual modifications or particle physics simulations, unlike NeRF, which is implicit. However, it comes with the drawback of being highly memory-intensive compared to NeRF. Additionally, its implementation requires very well-engineered GPU codes. Therefore, there is a need for another method that combines the advantages of both approaches: providing an explicit way to modify the scene while also enabling differentiable rendering in a simpler manner.

In this study, to fulfill the aforementioned requirements and provide a good representation for robot control, we employ Gaussian Process Distance Field (GPDF) \cite{DBLP:journals/corr/abs-2010-11487, Gentil2023AccurateGP}, originally designed for mapping tasks using point cloud sensor readings. However, we discovered that it offers more than just distance information; it can also estimate other properties such as color. Furthermore, although the original GPDF utilizes the interpolation capability of Gaussian Process, by leveraging the learning capability of Gaussian Process, we found that differentiable rendering similar to NERF can be achieved. This mean we can optimize explicit point cloud, which is an input for the GPDF. This will be further shown in Chapter \ref{sec:shape_representation} and Chapter \ref{sec:exploration}.



\subsection{Visual uncertainty and exploration}
Built on the surface reconstruction methods using differentiable rendering, researchers have developed various approaches around these techniques to represent uncertainty about the environment. Lee et al. \cite{lee2022uncertainty} employed the weight distribution of color samples along the ray to model volumetric or occupancy uncertainty, while Pan et al. \cite{Pan2022ActiveNeRFLW} integrated Bayesian neural networks with the original NERF to capture uncertainty. Jin et al. \cite{Jin2023NeUNBVNB} uses photometric loss function using color and reference images to sample next-best-view. Zhan et al. \cite{Zhan2022ActiveRMAPRF} adopted a volumetric approach, leveraging the entropy of termination probability to discern whether a point is occupied or not. Wen et al. \cite{Jiang2023FisherRFAV} utilized Fisher information with approximations to efficiently quantify surface points in both NERF and Gaussian splatting. He et al. \cite{He2023ActivePU} employed predictive information of color, occupancy, and depth to enable a drone to explore the environment using an ensemble of NERFs. 
While various surface uncertainty quantification measures are valuable, our approach adopts the method from He et al. \cite{He2023ActivePU} for three key reasons. Firstly, their framework offers a thorough formulation of information gain across different modalities, encompassing not only occupancy but also color and depth. Secondly, their method is adaptable across different models, not limited to neural networks. This versatility enhances its applicability to a broader range of scenarios and methodologies. Lastly, they demonstrated the effectiveness of their method with a drone, which shows potential to be applied for other robots.

Visual explorations beyond shape reconstruction, such as semantic identification, is very different from shape reconstruction. For example, researchers have explored techniques like object classification \cite{DBLP:journals/corr/abs-2108-00737} using pose ambiguity to find a good pose for classification and object segmentation in cluttered environments \cite{Hausman2012SegmentationOC} using feature tracking and heuristic rigidity evaluation. However, the emergence of language-image models like CLIP \cite{Radford2021LearningTV} has enabled simultaneous semantic identification and reconstruction. This allows embedding semantic information such as classification and segmentation into geometry, which can then be extended to semantic exploration using existing reconstruction techniques. While we do not directly demonstrate the effectiveness of exploring semantic information, we do show that our shape representation, GPDF, has the capability to store and optimize semantic information. Then, its information gain can be measured with He et al. \cite{He2023ActivePU}'s method, enabling various visual exploration as same as the shape reconstruction.
This demonstrates not only the GPDF's capability of serving as a unified spatial representation to store multimodal data but also illustrates how to incorporate visual structure beyond geometry, addressing one of the questions in Section \ref{sec:motivation} 


\subsection{Tactile exploration}
Tactile sensors provide information about the local surface of an object. Depending on the sensor type, they offer a wealth of information including contact points, forces, slips, and other parameters crucial for manipulation \cite{Shirai2023TactileTM, Hogan2020TactileDM, She2019CableMW}. Additionally, tactile sensors can measure parameters such as roughness and hardness through vibration analysis \cite{Ding2017TactilePO}, either via precise calibration methods or employing learning-based approaches \cite{Yuan2016EstimatingOH}. Moreover, tactile sensors can evaluate friction coefficients and stiffness by analyzing known object properties and determine fabric material from its texture \cite{Cao2023MultimodalZL}. They can also estimate the physical characteristics of sponges in latent space through exploratory identification methods \cite{Aoyama2023FewShotLO}. This comprehensive information greatly enhances a robot's decision-making and interaction capabilities.

Various tactile sensors utilize different underlying technologies, such as pressure sensors, capacitance or resistance sensors, and cameras \cite{Kappasov2015TactileSI, Roberts2021SoftTS}. Regardless of their technology, the primary objective remains consistent: measuring the deformation of the soft component of the sensor and its local extent.

On way to distinguish these sensors is the density of measurements. Camera-based sensors\cite{Yuan2017GelSightHR, Taylor2021GelSlim3H, Lepora2021SoftBO, Yin2022MultimodalPA} use cameras to generate high-resolution observations, resulting in thousands of pixel measurements  but with constraints regarding distance from the nominal surface. The other tactile sensors yield sparse measurements (1-20) from individual signal readers within the sensor array such as uSkin sensor from XELA Robotics \cite{XELA_Robotics_2018} or sensors from Contactile \cite{Contactile_2019}.  While camera-based sensors excels in detailed depth observation, intricate properties beyond depth such as forces or other surface properties may be better suited for a corse measurement as they often need to be aggregated from information across the entire surface or may simply be unable to be measured across the entire surface of the tactile sensor.

\begin{figure}[tb]
    \begin{minipage}[c]{0.48\linewidth}
        \centering
        \includegraphics[width=0.5\linewidth]{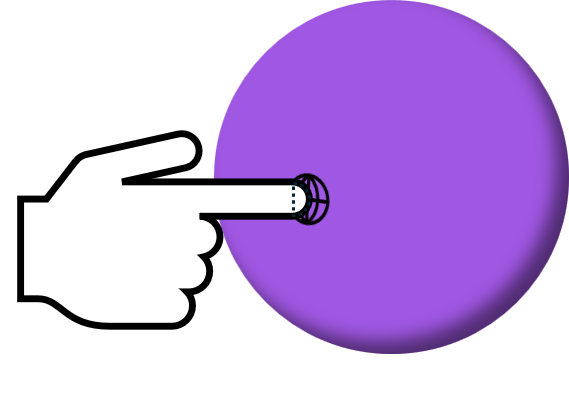}
        \caption{Illustration of a real tactile model where deformation happens}
        \label{fig:real_tactile_model}
    \end{minipage}
    \hfill
    \begin{minipage}[c]{0.48\linewidth}
        \centering
        \includegraphics[width=0.5\linewidth]{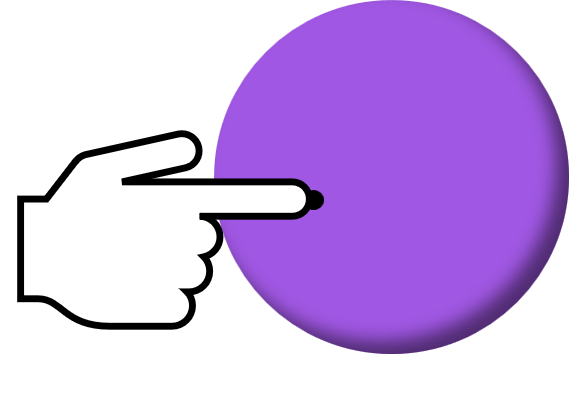}
        \caption{Illustration of a point-tactile model where everything is rigid}
        \label{fig:point_tactile_model}
    \end{minipage}%
\end{figure}

Another consideration for tactile model is rigidity of the object not just the sensor. When the tactile sensor and object make contact, both of their shapes deform, resulting in a complex surface contact as in Figure \ref{fig:real_tactile_model}. Efforts have been made to accurately model this non-rigidity, such as the hydroelastic contact model \cite{Masterjohn2021VelocityLA}. However, for simplicity, we assume that the object is rigid so that each measurement caused by deformation of the sensor can be regarded as point-tactile model depicted in Figure \ref{fig:point_tactile_model}.


For shape or geometry exploration, the inherently localized nature of tactile sensors makes it challenging to extract comprehensive information with just a few touches. To overcome this limitation, Driess et al. \cite{8793773} and Khadivar et al. \cite{KHADIVAR2023104461} utilized Gaussian Process Implicit Surface (GPIS). Their approach not only aggregated data from various touch trials but also facilitated robot control to maintain continuous contact with the object surface. This strategy enabled successful object shape reconstruction with minimal exploration, assuming an initial contact between the robot and the object. However, achieving this initial contact poses challenges, leading some researchers \cite{DBLP:journals/corr/abs-2103-00655, DBLP:journals/corr/abs-2109-09884} to leverage a point cloud of the object obtained from a fixed RGBD camera for better grasp pose or contact pose estimation. Suresh et al. \cite{DBLP:journals/corr/abs-2011-07044} addressed the fusion of 2D object reconstruction and tracking using GPIS and SLAM, while Sudharshan et al. \cite{Suresh2023NeuralFW} extended this framework to 3D using a combination of vision, camera-based tactile sensor and neural rendering without GPIS. Swann et al. \cite{Swann2024TouchGSVS} utilized GPIS with tactile exploration to improve Gaussian splatting from vision in parts of object that are noisy due to reflection or refraction. We adopt a similar approach to methods utilizing GPIS but leverage GPDF for more compact and improved surface uncertainty estimation. Additionally, as described in visual exploration, we demonstrate how differentiable rendering can be achieved, following a methodology similar to the work of Sudharshan et al. \cite{Suresh2023NeuralFW}. Finally, we show how to aggregate object property information, similar to Caccamo et al. \cite{Caccamo2016ActivePA}, who applied a standard Gaussian Process to actively perceive and identify deformable surfaces.


\section{Contributions and thesis organization}
Our approach adopts the Gaussian Process Distance Field (GPDF) \cite{Gentil2023AccurateGP} as the primary shape representation and shows its potential in the active perception domain. GPDF exhibits several beneficial characteristics compared to other shape representations such as Neural Implicit Surfaces (NeuS) \cite{Wang2021NeuSLN}, which provides distance using a neural network but does not have all following characteristics as GPDF:
\begin{enumerate}
    \item \textbf{Distance Field Generation:} It can output distance to a surface and its gradient and Hessian, aiding in collision avoidance and other control scenarios.
    \item \textbf{Multi-modal Data Integration:} It can seamlessly combine various types of multi-modal data.
    \item \textbf{Shape Representation Conversion:} It takes modifiable point cloud representation as input to generate an distance field with controllable interpolation.
    \item \textbf{Surface Uncertianty Quatification:} It provides a measure of uncertainty on the data surface, which is useful for guiding exploration.
    \item \textbf{Incremental Updates:} It supports incremental updates, allowing you to both add new data or optimize existing models.
\end{enumerate}

\begin{figure}[tb]
    \centering
    \includegraphics[width=\textwidth]{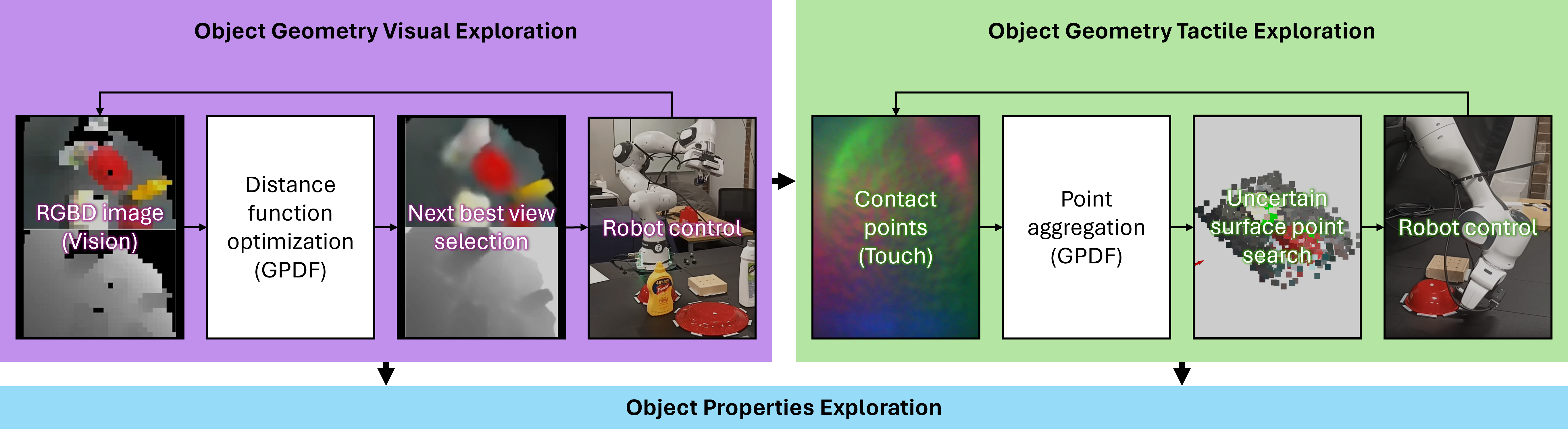}
    \caption{  
    The exploration begins with a rough estimate through vision, followed by finer detail analysis using tactile sensing. Using differentiable rendering, the system refines the point cloud positions and colors. The robot prioritizes information gain by seeking optimal views and reducing uncertainty at contact points. Once the geometry is recovered, the system can explore other object properties using the refined distance function, alone or in combination with vision and touch inputs.}
    \label{fig:pipeline}
\end{figure}
In subsequent chapters, we demonstrate how these characteristics facilitate the integration of observations from two modalities (refer to Figure \ref{fig:pipeline}): \textbf{vision}, which provides dense measurements with squared distance uncertainty \cite{8768489}, and \textbf{touch}, which offers less dense but low-uncertainty localized measurements.
In Chapter \ref{sec:hardware}, we detail the basic setup for the experiment, with a focus on the tactile sensor and its image processing, which is crucial for perceiving the environment. To convert raw images from tactile sensors into useful data like depth images or forces, we implemented a custom pipeline and calibrated the reflectance function and stiffness matrix. Our chosen shape representation, the Gaussian Process Distance Field (GPDF), is introduced in Chapter \ref{sec:shape_representation}. We begin by discussing why we selected it as our representation and its fundamental formulation, including methods for handling noisy observation points. Additionally, we delve into incremental update, which is important component in Active Perception to incorporate past observation. Moreover, we discuss methods to reduce the Gaussian Process's cubic complexity ($O(n^3)$) using heuristic point cloud sampling and kernel approximation. These techniques, like fixing the number of data points with inducing points or approximating the kernel function, help achieve linear computational complexity, enabling Gaussian Processes to handle thousands of data points efficiently. Chapter \ref{sec:exploration} explores how the GPDF can be optimized using multiple views from the RGBD camera and contacts from the tactile sensor, alongside methods for measuring their uncertainty to guide exploration strategies. The resulting camera pose or point will serve as reference poses to guide the robot controller also introduced in Chapter \ref{sec:exploration}. In the context of exploring object properties beyond shape, we retain the same assumptions used in shape exploration, with the given knowledge of $\Omega$ and $\partial\Omega$. We conduct real-world robot experiments in Chapter \ref{ch:property_exploration_touch} to validate our approach to integrating object properties such as material into GPDF.




\chapter{Experimental Setup and Perception Pipeline}
\label{sec:hardware}

In this chapter, we introduce the hardware setup for the experiment and outline the pipeline for processing raw image data from the tactile sensor's camera to generate depth images and estimate force measurements from them, as detailed in Section \ref{sec:tactile_calibration}. These components serve as fundamental elements largely forming the perception part of the active perception system for touch.
\begin{figure}[tb]
    \centering
    \includegraphics[width=1.0\textwidth]{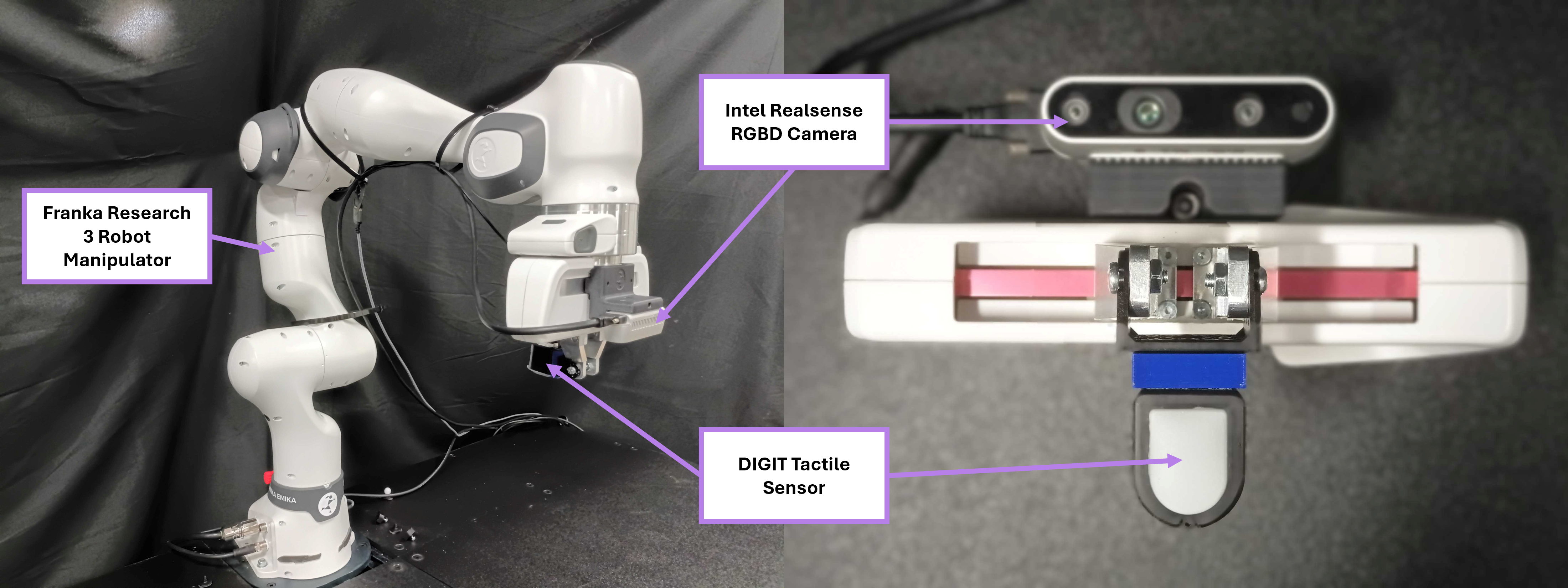}
    \caption{Hardware setup for the experiments. The left image provides an overview of the full robot arm setup, while the right image offers a close-up view of the sensors.}
    \label{fig:hardware}
\end{figure}

\section{Experimental setup}
Figure \ref{fig:hardware} shows the hardware configuration for this work. We have a robot arm that is going to do visual and tactile exploration using a depth camera and touch sensors. The hardware we are using is listed below.
\begin{itemize}
    \item Franka Research 3 robot manipulator
    \item Customized DIGIT tactile sensor \cite{Lambeta2020DIGITAN}
    \item Intel Realsense D435 RGBD camera
\end{itemize}
Since the sensors are attached to the end-effector of the robot, their pose can be determined in the robot's base frame through hand-eye calibration \cite{Tsai1988ANT}.


\section{Customized tactile sensor}\label{sec:tactile_calibration}
\begin{figure}[tb]
    \centering
    \includegraphics[width=0.8\textwidth]{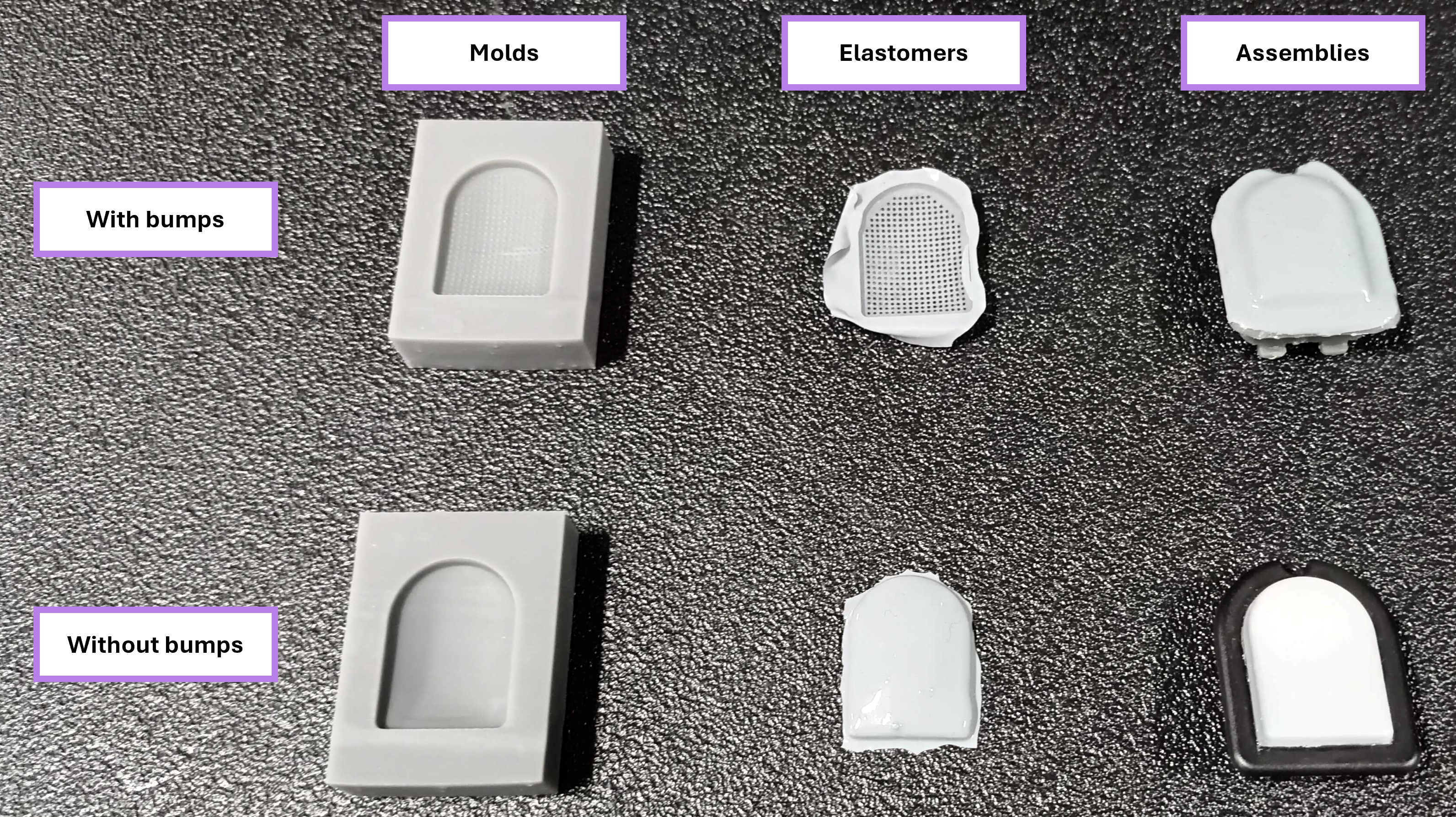}
    \caption{Manufacturing process of custom silicone elastomer with and without bumps.}
    \label{fig:custom_tactile_manufacture}
\end{figure}

All camera and gel-based tactile sensors \cite{Yuan2017GelSightHR, Taylor2021GelSlim3H, Yin2022MultimodalPA}, including the DIGIT tactile sensor \cite{Lambeta2020DIGITAN} utilized in our study, employ either light reflectance gradients or LEDs for measuring depth (z-axis) deformation, and marker tracking for measuring xy-axis deformation.  Specifically, the camera captures the reflection of LEDs from the reflective coating on the surface of the gel or elastomer, which varies with the deformation of the material, as depicted in Figure \ref{fig:digit_exploded}. However, the off-the-shelf DIGIT sensor does not include an option for including dot markers on the outer surface of the gel. Therefore, we customized the mold for the gel to feature hundreds of dot bumps, as depicted in the top left of Figure \ref{fig:custom_tactile_manufacture}, printed with Stereolithography (SLA) printers. These bumps allow us to fill them with black silicone inks, creating markers for tracking. A total of 359 0.5mm holes with a 1mm distance grid were created, with typically 204 points visible from the camera. We use Smooth-On products \cite{Smooth-On} for silicone gels and related materials, which are recommended by Azulay et al. \cite{Azulay2023AllSightAL}. The manufacturing process is as below:
\begin{figure}[tb]
    \centering
    \includegraphics[width=0.7\textwidth]{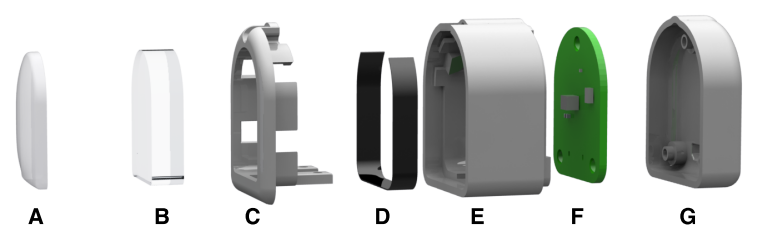}
    \caption{Image from the original DIGIT paper \cite{Lambeta2020DIGITAN}. Exploded view of a single DIGIT sensor. A) elastomer, B) acrylic window, C) snap-fit holder, D) lighting PCB, E) plastic housing, F) camera PCB, G) back housing.}
    \label{fig:digit_exploded}
\end{figure}
\begin{enumerate}
    \item Remove air bubbles from the silicone using air compressors before molding it into the custom mold. Once air bubbles are eliminated, pour the silicone into the mold and let it set for a minimum of 24 hours.
    \item Gently remove the silicone from the mold using alcohol. Apply a mixture of black pigment and silicone to cover the outer part of the elastomer. After covering, clean the surface thoroughly using alcohol and cotton swabs to ensure a smooth finish. Let is dry for few hours.
    \item Apply a mixture of reflective gray pigment, silicone, and silicone solvent to coat the outer part of the elastomer, as in the middle elastomers shown in Figure \ref{fig:custom_tactile_manufacture}. This coating dries within minutes.
    \item Attach the inner part of the elastomer to clear acrylic using silicone glue. Then, join it with the outer cover. For durability, reapply the reflective material on the surface as needed.
\end{enumerate}
In subsequent sections, we elaborate on the image processing techniques employed to extract deformation and force information, as well as the novel methodologies we have applied.


\section{Image processing pipeline \& force calibration}
\label{sec:image_processing_pipeline}
\begin{figure}[tb]
    \centering
    \includegraphics[width=1.0\textwidth]{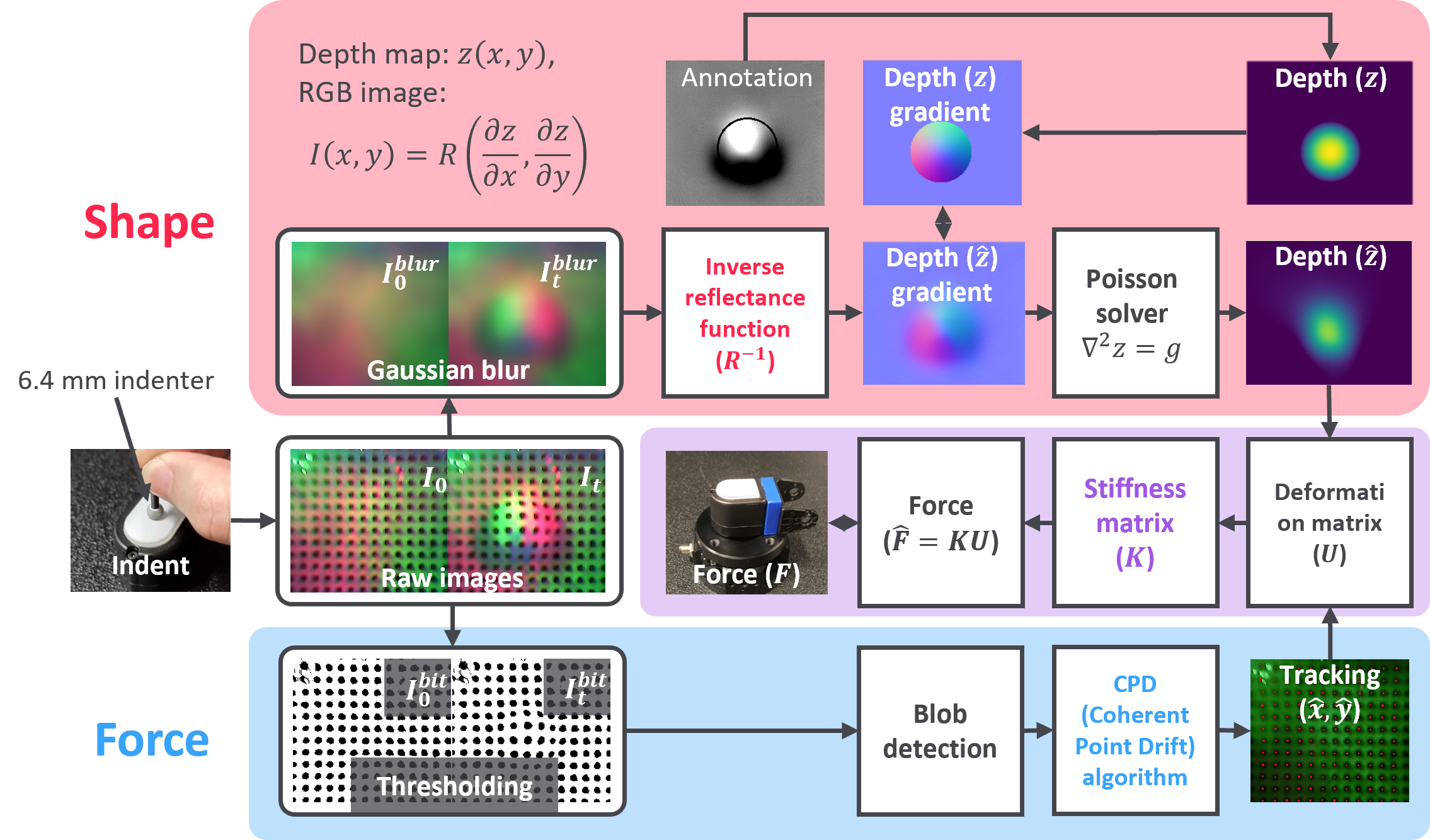}
    \caption{Image processing pipeline for the DIGIT tactile sensor that works in 30-40 Hz 
    }
    \label{fig:tactile_processing_pipeline}
\end{figure}

Our image processing methodology closely follows the architecture of GelSight \cite{Yuan2017GelSightHR}, with changes in each component, as illustrated in Figure \ref{fig:tactile_processing_pipeline}.


To estimate depth, we initially compare the current image with the initial or nominal image obtained without any contact. However, due to the presence of dot markers, which can interfere with proper color gradient computation, we first apply Gaussian blur to the extent that removes all tracker dots. Then, from the resulting color gradient, we infer the depth gradient from calibrated inverse reflectance function, which can subsequently be utilized to estimate the actual depth using a Poisson solver as in the top part of Figure \ref{fig:tactile_processing_pipeline}. We elaborate on the inverse reflectance functions in Section \ref{sec:irf}

To estimate xy deformation, we utilize blob tracking on the adaptive threshold image. Adaptive thresholding is crucial because the color gradient in the image can hinder accurate blob tracking. From the xy positions of the markers in the initial and current images, we can calculate the xy deformation. Therefore, tracking the correspondence between the initial marker positions and the current marker positions becomes a critical task, which we address using the Coherent Point Drift (CPD) algorithm as in the bottom part of Figure \ref{fig:tactile_processing_pipeline} and detailed in Section \ref{sec:cpd}.

As a final step, we use the full xyz deformation of markers to calculate the force. We obtain the z values for the markers from the depth map, which directly represent the z deformation value, thus providing us with the complete xyz deformation of the markers. Then, we utilize a calibrated stiffness matrix to convert deformation values into forces in Newton, as detailed in Section \ref{sec:stiff_calib}


\subsection{Inverse reflectance function calibration}
\label{sec:irf}
As described in the GelSight paper \cite{Yuan2017GelSightHR}, the height or depth map, $z(x, y)$, is related to the RGB color image, $I(x,y)$, through the following equation:
\begin{equation}
    I(x,y) = R\left(\frac{\partial z}{\partial x}, \frac{\partial z}{\partial y}\right)
\end{equation}
where $R$ is the reflectance function, and $x$ and $y$ represent pixel positions or their actual lateral positions. Having an inverse function of the reflectance function implies that we can obtain the gradient of $z$ solely from color pixels. Consequently, we can solve the Poisson equation $\nabla^2z=g$, where the boundary condition for $g$ is typically zero. For a flat-surfaced gel with well-positioned LEDs, the reflectance function is a known nonlinear function. However, this is not the case for DIGIT due to its curved surface and uneven lighting. Therefore, calibration with a known indenter is necessary, meaning that we must obtain the actual depth map and corresponding color images. We utilized a 6.4mm diameter indenter to indent the gel at various positions and automatically located them using a blob detector with the color gradient image, followed by manual adjustment to ensure correct labeling to create ground truth depth and its gradient as in the upper part of Figure \ref{fig:tactile_processing_pipeline}.

For better performance, we use the pixel value at $x$ and $y$ as well as its differences with neighboring pixels, and pixel coordinates as inputs to a inverse reflectance function, $R^{-1}$, which is modelled as a single $2\times 11$ matrix. Given blurred color image of a initial frame, $I_0^{blur}$, and a current frame $t$, $I_t^{blur}$, the equations are as follows, assuming that $x$ and $y$ are pixel coordinates.
\begin{equation}
    \begin{split}
        & I_t^{diff} = I_t^{blur} - I_0^{blur}\\
        &
        \begin{bmatrix}
            \frac{\partial z}{\partial x}(x,y) \\ \frac{\partial z}{\partial y}(x,y)
        \end{bmatrix} = 
        R^{-1}
        \begin{bmatrix}
            I_t^{diff}(x, y) \\ I_t^{diff}(x-1, y) \\ I_t^{diff}(x+1, y) \\ I_t^{diff}(x, y-1) \\ I_t^{diff}(x, y+1) \\ I_t^{diff}(x-1, y-1) \\ I_t^{diff}(x-1, y+1) \\ I_t^{diff}(x+1, y-1) \\ I_t^{diff}(x+1, y+1) \\ x \\ y
        \end{bmatrix}
    \end{split}
\end{equation}
This can be solved with a linear solver, but considering computer's memory constraints, especially given the size of the image and the dataset, we use incremental fitting. Once the matrix is fitted, we can utilize it to obtain the gradient of any shape, not limited to a sphere, for the Poisson solver, which converts it into a depth map, as illustrated in Figure \ref{fig:tactile_processing_pipeline}.

\subsection{Coherent Point Drift (CPD) algorithm}
\label{sec:cpd}
\begin{figure}[tb]
    \centering
    \includegraphics[width=0.9\textwidth]{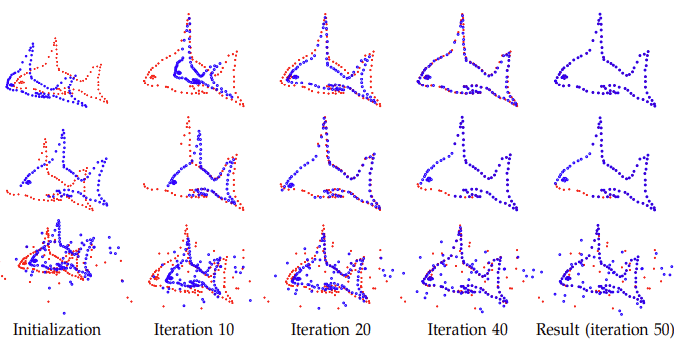}
    \caption{Image from the original CPD paper \cite{Myronenko2009PointSR} illustrating non-rigid point cloud registration}
    \label{fig:cpg_result}
\end{figure}
Previous works mainly used the Kalman filter and Hungarian algorithm \cite{Kuhn1955TheHM} to establish the correspondence between prediction and observation with a constant or zero velocity assumption. However, this method does not work well when we have many points that are close to each other and sometimes get occluded, breaking the one-to-one matching assumption in the Hungarian algorithm. Therefore, many heuristics are used to overcome these limitations. Instead, we use the Coherent Point Drift (CPD) algorithm \cite{Myronenko2009PointSR} that is usually used for non-rigid point cloud registration problems. This work considers point cloud registration as a probabilistic problem that tries to optimize for a cost function that includes regularization for velocity field smoothness from motion coherence theory \cite{Yuille1989AMA}. With this method, we can not only match points that are observed but also estimate missing points' positions as in Figure \ref{fig:cpg_result}. Since we have many points to track, we use low-rank approximation CPD for speed.

\subsection{Stiffness matrix calibration}
\label{sec:stiff_calib}
\begin{figure}[tb]
    \centering
    \includegraphics[width=1.0\textwidth]{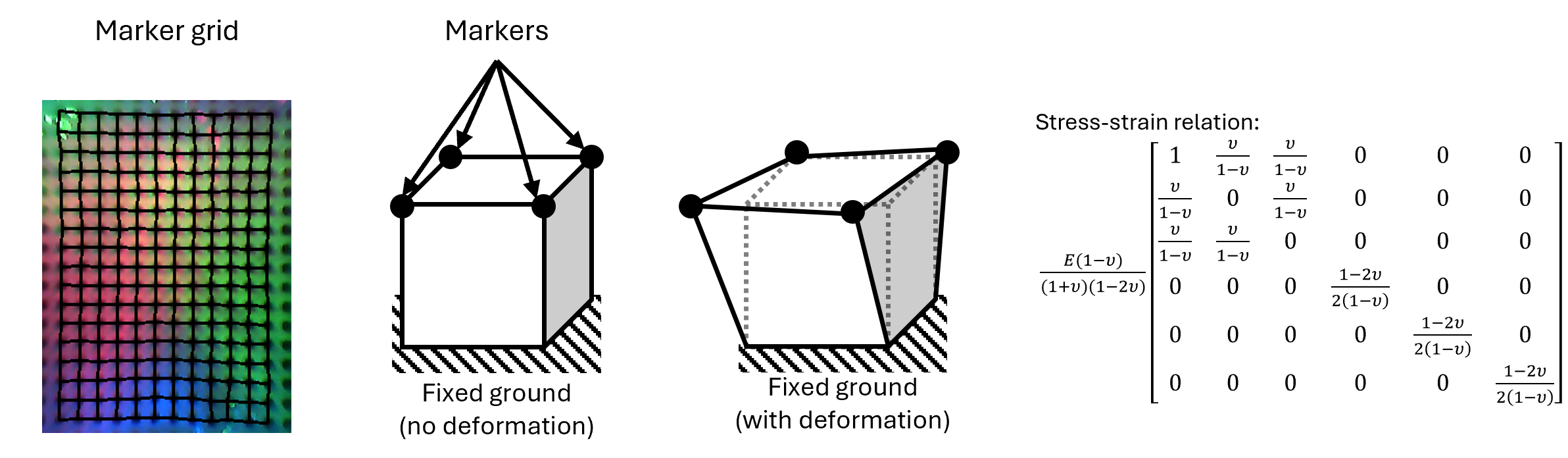}
    \caption{Markers and its neighbors forming hexahedron elements}
    \label{fig:marker_grid}
\end{figure}
For a flat gel-based tactile sensor, we can directly compute the forces acting on each marker using the Finite Element Method (FEM) theory. By treating markers as 8-node hexahedron elements with Dirichlet boundary conditions and knowing the Young’s Modulus, $E$, and Poisson’s ratio, $\nu$, of the gel, we can derive a stiffness matrix, $K$. From deformation measurements in each axis ($\delta_x$, $\delta_y$, $\delta_z$) for each markers, we get displacement matrix $U$, and the force is calculated as $F=KU$.

\begin{figure}[tb]
    \centering
    \includegraphics[width=0.8\linewidth]
    {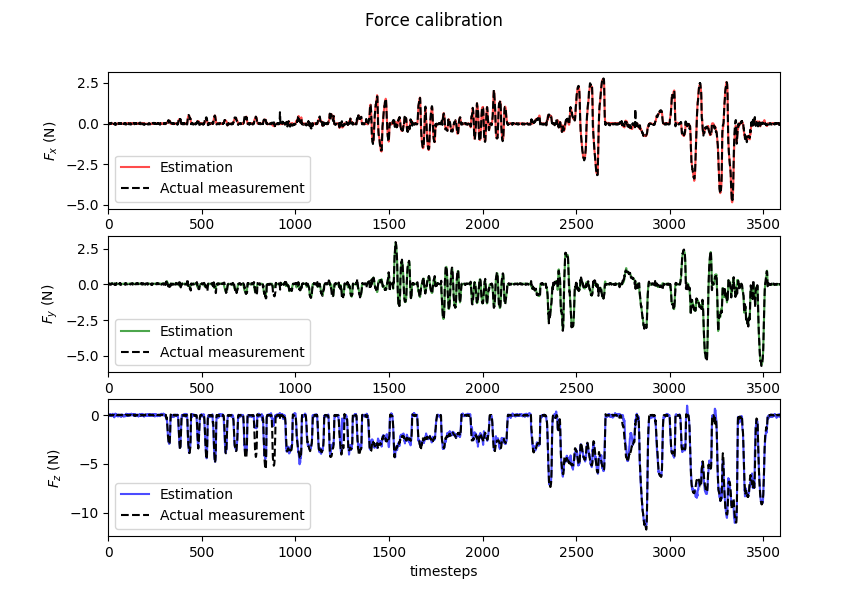}
    \caption{Force calibration result}
    \label{fig:force_calibration}
\end{figure}
However, the surface of DIGIT is curved, making it challenging to assume identical-length hexahedron elements, and the properties of the gel may vary based on the molding process. Instead of assuming a stiffness matrix $K$, we fit it using force measurements from a Bota System's force-torque sensor \cite{Kaslin2018TowardsAP} while firmly placing DIGIT on top of the sensor.

We fit the stiffness matrix $K$ such that the sum of forces in the elements closely matches the actual measurements from force sensor using gradient descent. We can use a compact stiffness matrix that outputs three values representing the total force sum directly. For this, by utilizing a linear solver, we can obtain an optimal value; however, this approach may lead to overfitting due to a lack of awareness of the structural relationships between markers. While this issue can be mitigated with a large amount of data, we opt for a different approach. We utilize gradient descent with constraining stiffness matrix values that do not align with neighboring markers to be zero. The calibration result is in Figure \ref{fig:force_calibration}.


\chapter{Representation for Spatial Data}\label{sec:shape_representation}
In this chapter, we provide comprehensive explanation and methodology about Gaussian Process Distance Function (GPDF). In Section \ref{sec:shape_representation_criteria}, we show reasoning behind our decision to use GPDF. Additionally, we offer an explanation of the details of GPDF's formulation in Section \ref{sec:GPDF}. We describe in Section \ref{sec:incremental_update} how GPDF can be used within incremental update paradigm that is able to incorporate incoming data streams. Furthermore, we explore how GPDF's computational complexity can be alleviated with different approximation techniques as shown in Section \ref{sec:heristic_approximation} for a heuristic approximation using point cloud downsampling and also explore and provide a comprehensive comparison of state-of-the-art kernel approximation techniques and their effect on the GPDF distance and gradients accuracy in Section \ref{sec:kernel_approximation}.
\section{Shape representations}
\label{sec:shape_representation_criteria}
\begin{figure}[!h]
    \centering
    \includegraphics[width=\textwidth]{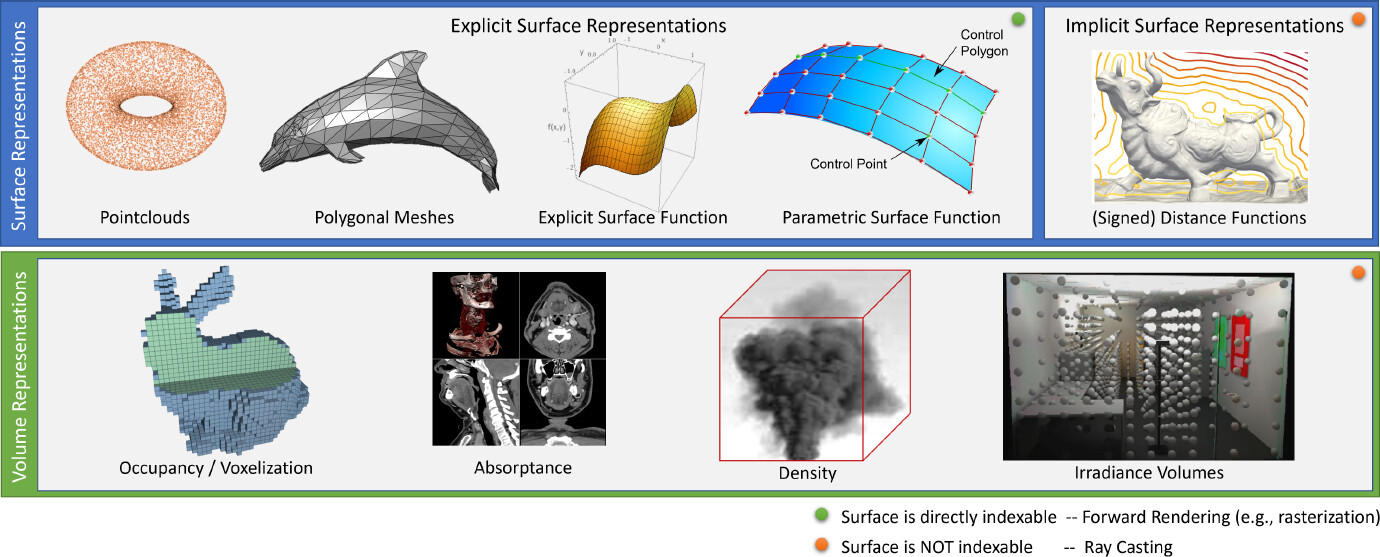}
    \caption{Different representations of object shape. Image from \cite{Tewari2021AdvancesIN}}
    \label{fig:shape_representations}
\end{figure}
In the field of computer graphics, various methods have been developed to represent object shapes, as illustrated in Figure \ref{fig:shape_representations}. These methods are typically classified based on two main factors: whether they describe the surface or the volume of an object, and whether they use an implicit or explicit approach. While techniques like ray casting enable exploration in volumetric space, tactile sensing primarily provides surface information. Therefore, we opt for surface-based object shape representation to combine both visual and tactile data. 


Signed Distance Function (SDF), which is implicit surface representation, is widely favored as the preferred shape representation due to its differentiability, continuity, and its ability to provide information throughout Euclidean space rather than just on the surface unlike explicit surface representations. In robotics, the SDF has found applications in Simultaneous Localization and Mapping (SLAM), contact simulation, manipulation, trajectory optimization, and more \cite{Ortiz2022iSDFRN, Tracy2022DifferentiableCD, Narang2022FactoryFC, Li2023RepresentingRG, Koptev2023NeuralJS, Zhang2023ContinuousIS}. 
These examples demonstrate the versatility of SDF and their flexible usage compared to explicit surface representations. This versatility motivates our exploration of using SDF as a shape representation.

\begin{figure}[tb]
    \begin{minipage}[c]{0.49\linewidth}
        \centering
        \includegraphics[width=1.0\linewidth]{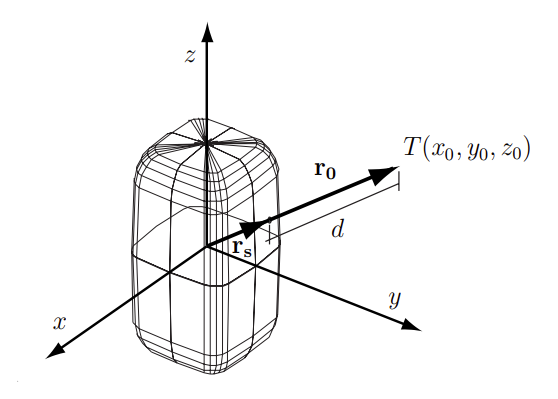}
        \caption{Radial Euclidean distance of superquadrics. Image from \cite{Jakli2000SuperquadricsAT}.}
        \label{fig:radial_euclidean_distance}
    \end{minipage}%
    \hfill
    \begin{minipage}[c]{0.49\linewidth}
        \centering
        \includegraphics[width=0.7\linewidth]
        {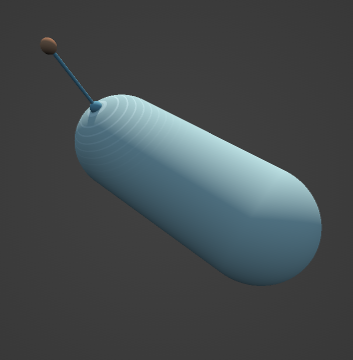}
        \caption{Signed distance of a capsule. Image from \cite{iq_2021}.}
        \label{fig:signed_distance_capsule}
    \end{minipage}
\end{figure}

However, integrating observations from tactile sensors, which inherently produce point or discrete data, into an SDF poses challenges. Additionally, because the SDF is an implicit representation, controlling the fitted or learned function becomes difficult. One approach to leverage both advantages of explicit and implicit surface representations is to find a function that converts explicit representations such as point clouds from sensors into implicit representations. Previous efforts have explored similar ideas, such as using superquadrics—a parametric surface that can fit to a point cloud \cite{Bajcsy1987ThreeDO, Liu2021RobustAA}. Superquadrics does not give SDF directly as in Figure \ref{fig:signed_distance_capsule}, but give analytical radial distance function as in Figure \ref{fig:radial_euclidean_distance} or can obtain SDF from iterations. While superquadrics provide abstract geometry and can describe scenes through composition, they are time-consuming to fit and cannot handle complex shapes without composition. 

To address these challenges, recent approaches utilize neural networks to convert point clouds into distance functions \cite{Merwe2019LearningC3, Chou2022GenSDFTL, Chibane2020NeuralUD}. Trained on extensive 3D object datasets, these networks can even handle unseen objects and optimize shape representations based on observations. However, they are typically limited to individual objects and do not consider object color or texture, which could also be useful for inferring geometry. To overcome these limitations, we propose an approach centered around using an explicit point cloud with features as the foundational shape representation. Point cloud can be transformed into a continuous form using Gaussian Processes' interpolation and learning capability. This means that any surface represented with a point cloud can be used to create signed distance function without any offline learning on large amounts of data.  This method allows us to seamlessly incorporate multi-modal observations unlike neural network models that require several models for different modalities, enhancing the robustness and flexibility of our representation in capturing tactile information and other object properties.


\section{Gaussian Process Distance Field}\label{sec:GPDF}
In this section, we begin by briefly introducting the concept of Gaussian Process Regression (GPR) in Secetion \ref{subsec:gpr}. Following that, we delve into the background and formulation of Gaussian Process Distance Function (GPDF) in Section \ref{subsec:gpdf}. Here, we also discuss methods for iteratively enhancing distance prediction accuracy and obtaining proxy uncertainty for distance estimation. Finally, in Section \ref{subsec:nigp}, we explore how to handle noisy inputs, particularly in cases where inputs have varying levels of noise variance.

\subsection{Gaussian Process Regression}\label{subsec:gpr}
Gaussian Process Regression (GPR) \cite{Seeger2004GaussianPF} is a probabilistic model used for interpolation. It has random variables indexed by time or space, with finite subsets following multivariate Gaussian distributions. When given input $\mathbf{x}\in\mathbf{R}^n$, a function $f$, and noisy observations $y=f(\mathbf{x}) + \epsilon$ of the function, where $\epsilon\sim\mathcal{N}(0, \sigma_y^2)$, we can say that GP models function $f$ as $\mathcal{GP}(0,k(\mathbf{x}, \mathbf{X}))$, where $(\mathbf{X},\mathbf{y})$ represents the observed data points and $k$ denotes the kernel function. Then, the predictive mean and variance of $y$ is
\begin{equation}
    \begin{split}
        & \hat{f}(\mathbf{x})=k(\mathbf{x},\mathbf{X})(K(\mathbf{X}, \mathbf{X})+\sigma_y^2 I)^{-1}\mathbf{y}\\
        & \mathbf{var}(f(\mathbf{x})) = k(\mathbf{x}, \mathbf{x}) - k(\mathbf{x},\mathbf{X})(K(\mathbf{X}, \mathbf{X})+\sigma_y^2 I)^{-1}k(\mathbf{X}, \mathbf{x})
    \end{split}
\end{equation}

\begin{figure}[!tbp]
  \centering
  \includegraphics[width=\linewidth]{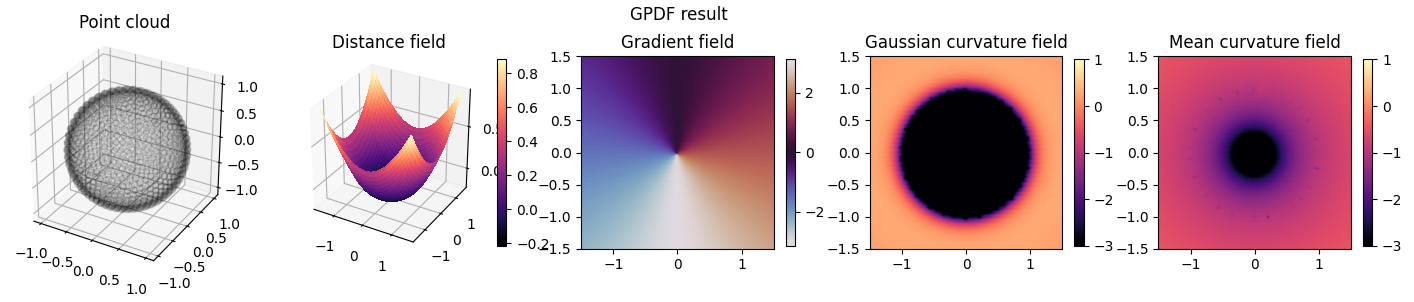}
  \caption{An illustration of GPDF using a 3D point cloud. From a raw point cloud, GPDF gets distance, gradient, and curvature fields from query points, which lie on the z=0 plane. The values for curvatures have been clipped between -3 and 1.}
  \label{fig:gpdf_demonstration}
\vspace{-10pt}
\end{figure}

\subsection{Gaussian Process Distance Field}\label{subsec:gpdf}
The concept of utilizing Gaussian Processes (GPs) to represent object shape was initially introduced through Gaussian Process Implicit Surface \cite{williams2007gaussian}. This work leveraged pseudo-occupancy information as input and showcased impressive interpolation capabilities, providing highly detailed reconstructions of object shapes. However, its effectiveness was predominantly limited to regions near the surface, and obtaining accurate occupancy information solely from RGBD images proved challenging. Subsequent advancements in this line of research led to the development of Gaussian Process Distance Field \cite{Gentil2023AccurateGP}. This approach only relies on surface points as inputs and generates a continuous distance field through the use of a technique known as reverting function as in Figure \ref{fig:gpdf_demonstration}.

\begin{figure}[!tbp]
  \centering
  \includegraphics[width=\linewidth]{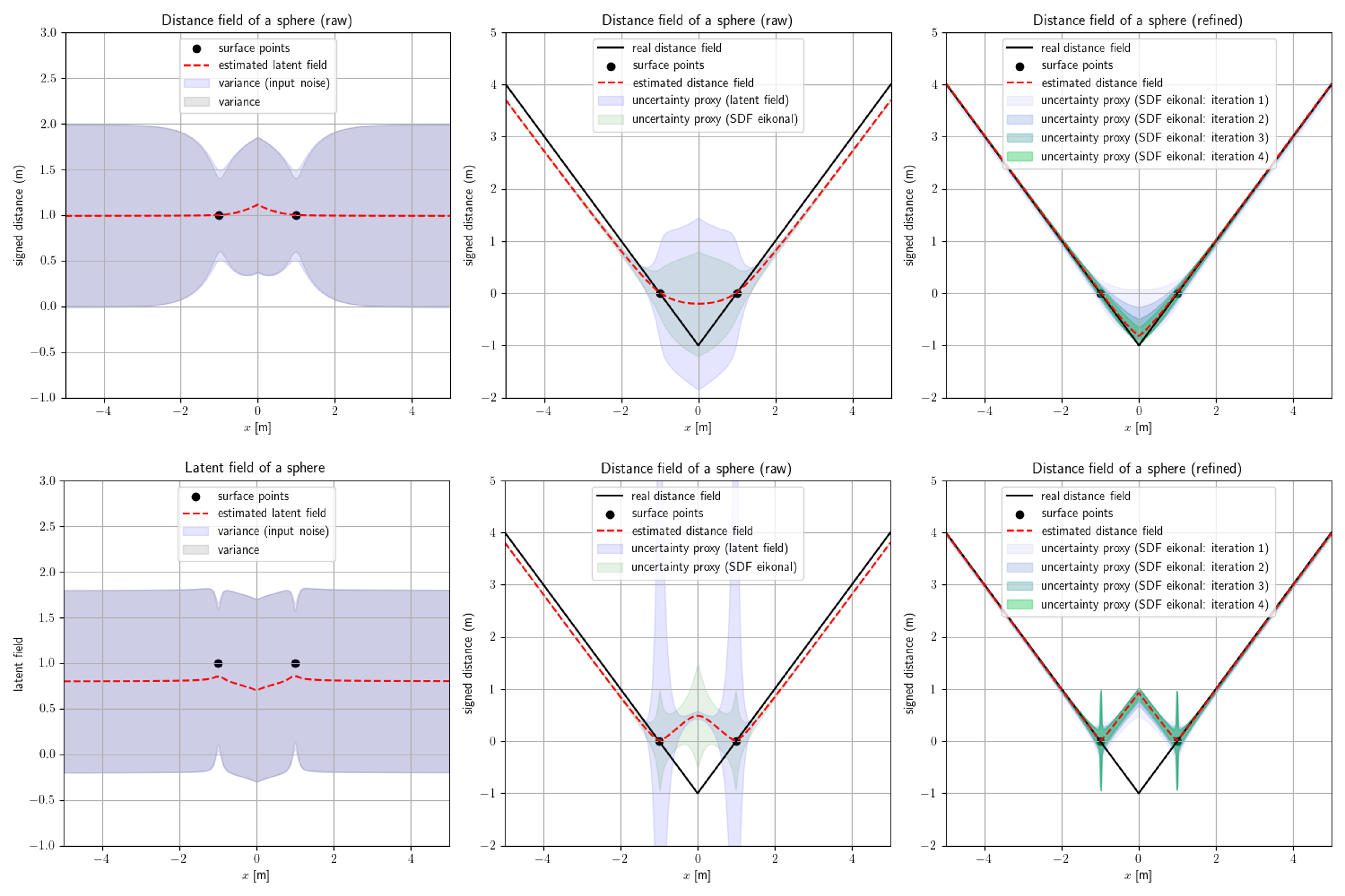}
  \caption{A 1D slice of a distance field (top: $l=1$, bottom: $l=0.1$) of a sphere point cloud. The image in the middle illustrates a comparison between the ground truth SDF and the initial estimation derived from the GPDF. On the right, the image illustrates an improved estimation achieved through five iterations of gradient descent, employing the ray marching concept \cite{Choi2023TowardsFD}.}
  \label{fig:gpdf_iteration}
\vspace{-10pt}
\end{figure}

To further elaborate on the methodology, let's assume that $\mathbf{x}\in\partial\Omega$ is points on an object's surface in $\mathbb{R}^3$. Gentil et al. \cite{Gentil2023AccurateGP} models occupancy field $o(\mathbf{x})\sim\mathcal{GP}(0,k(\mathbf{x},\mathbf{x}'))$ using GP and use the reverting function $r(o(\mathbf{x}))$ to output the distance field $d(\mathbf{x})$, which satisfies $|\nabla d(\mathbf{x})|=|\frac{\partial r}{\partial o}\nabla o(\mathbf{x})|=1$. Reverting function $r$ is an inverse of the kernel $k$ where $r(k(\mathbf{x}, \mathbf{x}'))=||\mathbf{x}-\mathbf{x}'||$. If we set $\mathbf{x}_*$ as a query point, the predictive posterior mean and variance of the occupancy field of the query point are
\begin{equation}
    \begin{split}
        & \hat{o}(\mathbf{x_*})=k(\mathbf{x_*},\mathbf{X})(K(\mathbf{X},\mathbf{X})+\sigma_y^2 I)^{-1}\mathbf{y}\\
        & \mathbf{var}(o(\mathbf{x_*})) = k(\mathbf{x_*},\mathbf{x_*}) - k(\mathbf{x_*},\mathbf{X})(K(\mathbf{X},\mathbf{X})+\sigma_y^2 I)^{-1}k(\mathbf{X},\mathbf{x_*})
    \end{split}
\end{equation}
\noindent
where $\mathbf{X}=[\mathbf{x}_1,\cdots,\mathbf{x}_n]$, $\mathbf{y}=[y_1,\cdots,y_n]=\mathbf{1}$, $\mathbf{x}_i\in\mathbb{R}^D$,and $y_i=o(\mathbf{x}_i)+\epsilon_{y,i}$ for $\epsilon_{y,i}\sim\mathcal{N}(0, \sigma_y^2)$ and $i=1,\cdots,n$. The initial approach involves estimating the unsigned Euclidean distance function, utilizing either the Rational Quadratic kernel, $\left(1+\frac{d^2}{2 \alpha l^2}\right)^{-\alpha}$,  or Squared Exponential kernel, $\exp\left(-\frac{d^2}{2l^2}\right)$.

However, we have found that by using the bijectiveness of the Mat\'ern kernel with $\nu=1/2$, $\exp\left(-\frac{d}{l}\right)$, we can also estimate the signed distance function \cite{Choi2023TowardsFD}. To verify the bijectiveness of the reverting function, please refer to Appendix \ref{covariance_kernels}. Nevertheless, this enhancement comes with a minor trade-off in accuracy, attributed to the kernel's limitations in handling high frequencies. To address this drawback, we have incorporated ray marching techniques, iteratively refining the accuracy in our approach as in Figure \ref{fig:gpdf_iteration}. The equation for iteration for query point, $\mathbf{x_*}$, is as below.
\begin{equation}
    \mathbf{x_*}\leftarrow \mathbf{x_*} - d(\mathbf{x_*}) \frac{\nabla d(\mathbf{x_*})}{||\nabla d(\mathbf{x_*})||_2}
\end{equation}
For the Mat\'ern kernel with $\nu=1/2$, the way of getting the surface normal and curvature is described in Appendix \ref{normal_curvature}. The challenge posed by GPDF arises from its dependence on surface points as inputs, confining well-defined uncertainty exclusively to these surfaces. In Figure \ref{fig:gpdf_iteration}, the discrepancy is evident: minimal variance is observed in the vicinity of the surface points, while it intensifies elsewhere. To overcome this limitation, the original paper proposes a remedy by evaluating the Mahalanobis distance between the derivative of the latent field, $\frac{\partial|k(d)|}{\partial d}|_{\hat{d}}$, and the gradient of the inferred field, $\|\nabla \hat{o}(\mathbf{x})\|$, using covariance, $\Sigma_{\hat{o}}(\mathbf{x})$. The full equation is 
\begin{equation}
    \sqrt{(\frac{\partial|k(d)|}{\partial d}|_{\hat{d}}-\|\nabla \hat{o}(\mathbf{x})\|)\Sigma_{\hat{o}}(\mathbf{x})^{-1}(\frac{\partial|k(d)|}{\partial d}|_{\hat{d}}-\|\nabla \hat{o}(\mathbf{x})\|)}
\end{equation}
This serves as a surrogate for distance uncertainty, as can be seen in the middle image of Figure \ref{fig:gpdf_iteration} \cite{Gentil2023AccurateGP}.

An alternative method for gauging uncertainty entails harnessing the inherent characteristics of the eikonal equation, where $|\nabla d(\mathbf{x})|=1$ must be satisfied. Consequently, a significant deviation from one in the inferred gradient of the distance function, $\sigma=|1-\|\nabla \hat{d}(\mathbf{x})\||$, indicates an uncertain point. In the context of the iterative ray marching process, uncertainty measurements are combined, as depicted in the right image of Figure \ref{fig:gpdf_iteration}, using the equation below.
\begin{equation}
    \sigma^2=\left(\sum_{i=1}^n (1/\sigma_i^2)\right)^{-1}
\end{equation}
where $n$ represent the number of iterations. 

Both methodologies successfully capture uncertainty by assigning high uncertainty values to points on and within the surface in high interpolation scenarios. However, in low interpolation cases, as illustrated in bottom images of Figure \ref{fig:gpdf_iteration}, where an unsigned distance field is formed, the latent field uncertainty proxy fails to provide accurate uncertainty estimates for the center of the sphere, particularly when using the Mat\'ern $1/2$ kernel. In Chapter \ref{sec:exploration}, alternative approaches for evaluating uncertainty concerning the object surface in relation to the camera view will be investigated. To examine the influence of the interpolation parameter $l$, please refer to Appendix \ref{interpolation_parameter}. In essence, a value for $l$ that is at least twice the nominal distance between observation points tends to yield satisfactory results.

\subsection{Noisy input Gaussian Process}\label{subsec:nigp}
Earlier, our assumption was based on the presence of uniform noise solely in the observation. However, this assumption may not hold true in most cases. Specifically, for GPDF, assuming or modelling white noise on the observation is hard as we fix the observation to be 1. Instead, we know that the noise originates from the input $\mathbf{X}$. Therefore, we now consider the scenario of noisy input, employing the method elucidated by Mchutchon et al. \cite{NIPS2011_a8e864d0} without observation noise.
\begin{equation}
    \begin{split}
        & y_i = o(\mathbf{x}_i + \epsilon_{x,i}) \approx o(\mathbf{x}_i) + \epsilon_{x,i}^T \partial_{\Bar{f}} \\
        & \hat{o}(\mathbf{x_*})=k(\mathbf{x_*},\mathbf{X})(K(\mathbf{X},\mathbf{X}) + diag(\Delta_{\Bar{f}}\mathbf{\Sigma}_x\Delta_{\Bar{f}}^T))^{-1}\mathbf{y}\\
        & \mathbf{var}(o(\mathbf{x_*})) = k(\mathbf{x_*},\mathbf{x_*}) - k(\mathbf{x_*},\mathbf{X})(K(\mathbf{X},\mathbf{X})+diag(\Delta_{\Bar{f}}\mathbf{\Sigma}_x\Delta_{\Bar{f}}^T))^{-1}k(\mathbf{X},\mathbf{x_*})
    \end{split}
\end{equation}
\noindent
where $\epsilon_{x,i}\sim \mathcal{N}(\mathbf{0}, \mathbf{\Sigma}_x)$ with diagonal matrix $\mathbf{\Sigma}_x$, $\partial_{\Bar{f}}$, which is a gradient of one mean function value, and $\Delta_{\Bar{f}}$, which is a gradient of $n$ mean function values. This allow each data point to have a distinct variance unlike previous assumption where all input data points were considered to follow the same Gaussian distribution.

The original GPDF papers caution against using the aforementioned method, asserting that the derivative of the latent field's mean is null. Nevertheless, in our testing, even a subtle correction term induced preferable changes, giving very curved areas higher uncertainty and flat areas lower uncertainty, as illustrated in the left image of Figure \ref{fig:gpdf_iteration}.
This discrepancy appears to stem from confusion between the derivative and gradient. Notably, we can fix $\mathbf{\Sigma}_x$ based on sensor specifications, eliminating the need for hyperparameter learning through marginal likelihood. However, as there is no closed analytical solution for Gaussian Processes (GP) due to its reliance on its own gradient with respect to observation points, we can seek an incremental iterative approach for updates as proposed in Section \ref{sec:incremental_update}.

\section{Incremental update}\label{sec:incremental_update}
An important aspect to consider for shape representation is how to incorporate past observations into the representation or update it each time a new observation is received. This ensures a well-functioning closed loop in active perception. Therefore, we explore two methods for incorporating past observations in GPs.


\subsection{Addition and deletion of data points}
Due to the occlusion or locality, observation occurs incrementally in an online fashion, especially for tactile sensors. However, training the GP with the entire set of old and new observations from scratch proves to be highly inefficient due to the costly matrix inverse process. As a remedy, we integrate the old observations $\mathbf{X}_1\in\mathbb{R}^{n_1\times3}$ and the new observations $\mathbf{X}_2\in\mathbb{R}^{n_2\times3}$ into a combined matrix $\mathbf{X}_3\in\mathbb{R}^{(n_1+n_2)\times3}$ by applying additive incremental updates to the inverse matrix using block matrix inverse. In this way, unlike neural network, GP can incorporate new data points without fitting the whole function again.
\begin{equation}
    \begin{split}
        & (K_{33}+D_3)^{-1}=
        \begin{bmatrix}
            K_{11} +D_1 & K_{12} \\
            K_{12}^T &  K_{22}+D_2
        \end{bmatrix}^{-1} =
        \begin{bmatrix}
            J_{11} & J_{12} \\
            J_{12}^T & J_{22}
        \end{bmatrix} \\        
        & = \tiny\begin{bmatrix}
            (K_{11} +D_1)^{-1}+ (K_{11} +D_1)^{-1}K_{12}(K_{22}+D_2)^{-1}K_{12}^{T}(K_{11} +D_1)^{-1} & -(K_{11} +D_1)^{-1}K_{12}(K_{22}+D_2)^{-1} \\
            -(K_{22}+D_2)^{-1}K_{12}^{T}(K_{11} +D_1)^{-1} & (K_{22}+D_2)^{-1}
        \end{bmatrix}\\
        & K_{ij} = K(\mathbf{X}_i, \mathbf{X}_j) \text{ for $i,j\in\{1, 2, 3\}$} \\
        & D_i = diag(\Delta_{\Bar{f}_i}\mathbf{\Sigma}_x\Delta_{\Bar{f}_i}^T) \text{\quad or \quad} \sigma_y^2 I \text{\quad for $i\in\{1, 2, 3\}$} 
    \end{split}
\end{equation}
Therefore, if we already have the inverse matrix $(K_{11} + D_1)^{-1}$, obtaining $(K_{33} + D_2)^{-1}$ involves only taking the actual inverse of the matrix $(K_{22} + D_2)^{-1}$, with the remaining steps being matrix multiplication. The deletion process mirrors the additive process. Suppose we have the original inverse matrix $(K_{33} + D_2)^{-1}$ and want to remove observations associated with $\mathbf{X}_2$. In that case, we shift the rows and columns of the original inverse matrix to create a block matrix. This separates the observations we wish to retain, $\mathbf{X}_1$, from those we want to delete, $\mathbf{X}_2$, forming the mentioned block matrix. The resulting expression is then given by:
\begin{equation}
    (K_{11} + D_1)^{-1} = J_{11} - J_{12}(J_{22})^{-1}J_{12}^T
\end{equation}
This incremental update proves to be highly efficient, especially when the size of the initial set, represented by $n_1$, is significantly larger than the new observations, denoted as $n_2$, with a preferable condition being $n_1\gg n_2$ where $n_1$ is at least 3-5 times greater than $n_2$. The efficiency is further pronounced in the case of GP approximations, as the incremental update essentially simplifies to real addition and subtraction operations, along with taking the inverse of the result. This process exhibits a fixed computational complexity of $O(m^3)$, where $m$ signifies the approximation parameter.

\subsection{Optimizing inducing points}
Typically, Gaussian Processes (GPs) aggregate observation points continuously. However, by employing the inducing point approach, where a fixed number of points $\mathbf{X}_m \in \mathbb{R}^{m\times 3}$ is utilized, updates are performed solely on these inducing points. This is achieved by minimizing the Kullback-Leibler (KL) divergence between the GP generated from all observations and the GP constructed solely with the inducing points. Maximizing the equation below facilitates this minimization:
\begin{equation}
\begin{split}
    & \mathcal{L}(\mathbf{X}_m) = \log \mathcal{N}(\mathbf{y}|\mathbf{0}, \sigma_y^2I + Q) -\frac{1}{2\sigma_y^2}\rm{Tr}(K(\mathbf{X}, \mathbf{X}) - Q) \\
    & Q = K(\mathbf{X}, \mathbf{X}_m)K(\mathbf{X}_m, \mathbf{X}_m)^{-1}K(\mathbf{X}_m, \mathbf{X})\\
\end{split}
\end{equation}
In practice, more numerically stable and factorized equation is used \cite{Seeger2004GaussianPF}. The computational complexity of this approach is $O(nm^2)$.
\begin{figure}[!tbp]
  \centering
  \includegraphics[width=0.5\linewidth]{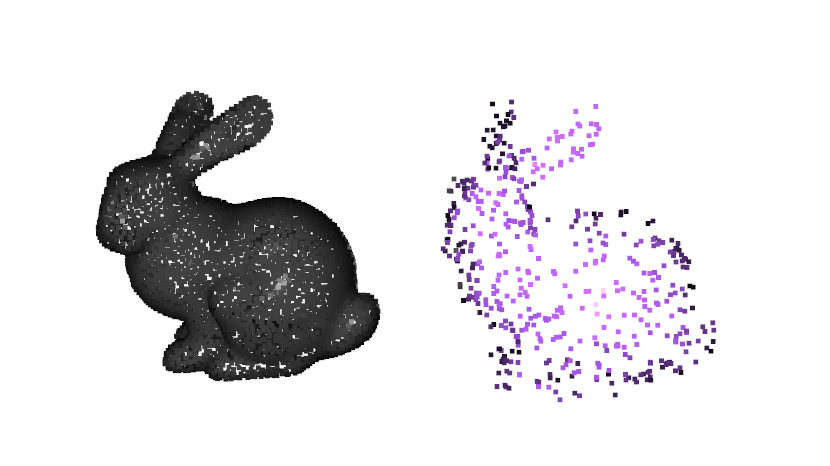}
  \caption{Original point cloud of Stanford Bunny (left) and downsampled point cloud using the inducing point method (right)}
  \label{fig:downsample_inducing_points}
\vspace{-10pt}
\end{figure}
The inducing point method can serve for both downsampling and upsampling purposes, as depicted in Figure \ref{fig:downsample_inducing_points}. In Chapter \ref{sec:exploration}, we illustrate how we can extend the inducing point method using differentiable rendering of SDF for calculating information gain of camera poses.

\section{Heuristic approximation: point cloud downsampling} \label{sec:heristic_approximation}

\begin{figure}[!tbp]
  \centering
  \includegraphics[width=\linewidth]{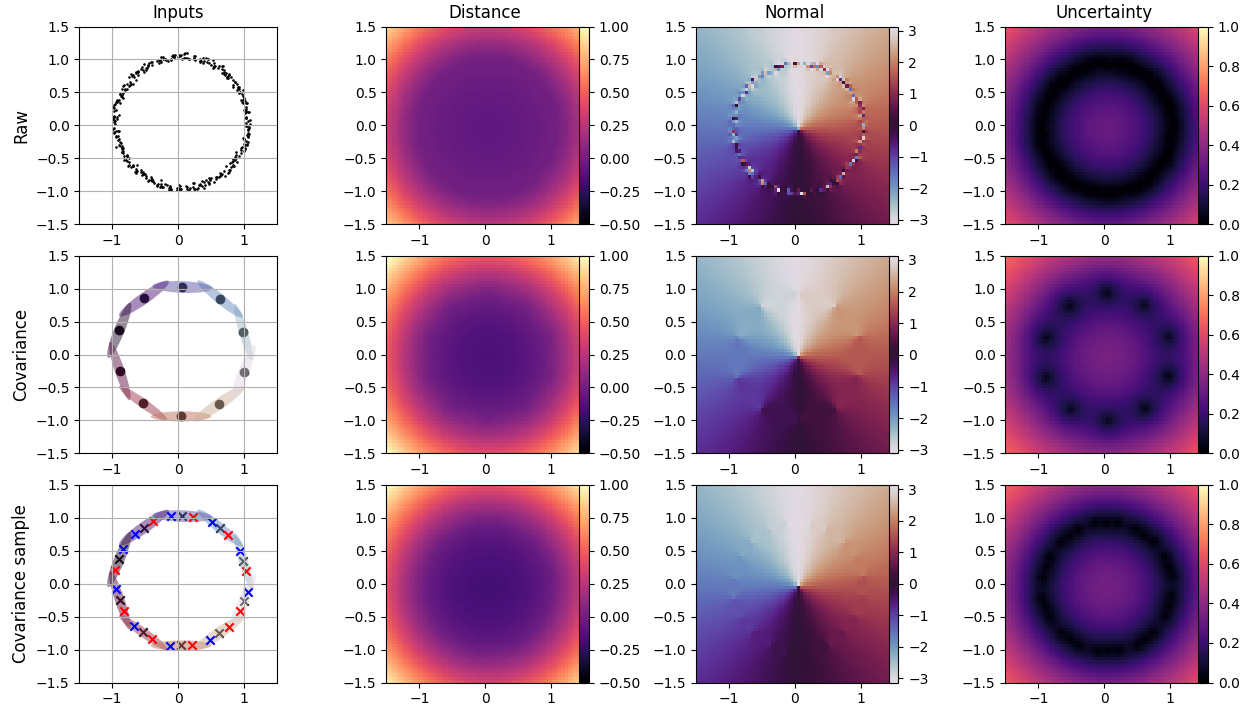}
  \caption{Results of GPDF for raw input and down-sampled inputs.}
  \label{fig:downsampling_demo}
\vspace{-10pt}
\end{figure}
An abundance of observations often leads to repetitive data, resulting in marginal improvement in reconstruction quality. Moreover, inadequate tuning of observation noise hyperparameters can detrimentally affect the estimation of surface normals, as depicted in Figure \ref{fig:downsampling_demo}. To overcome this challenge, we can employ a simple yet effective heuristic approach: voxel down-sampling of the point cloud used for the GP. However, this introduces a challenge in managing uncertainty due to the reduced number of observations. While there isn't a universal solution for simultaneously addressing noise and ensuring accurate uncertainty quantification, we tackle these issues by implementing an incremental updating voxel down-sampling technique. This approach maintains the distribution and covariance of sample points within voxels, providing a practical solution to alleviate these challenges. Before discussing the details of updating voxel samples, let's explore two methods of utilizing the covariance matrix.

The first method entails utilizing the covariance matrix as input noise and utilizing a noisy input Gaussian Process (GP). This approach yields outcomes akin to those seen in the second row of Figure \ref{fig:downsampling_demo}. The effect on uncertainty is evident in the increased uncertainty of data points, aiding in mitigating the discontinuity induced by down-sampling. However, the uncertainty still deviates somewhat from that observed when using raw inputs.

Another approach leverages the fact that the covariance matrix can be regarded as the primitive shape within the voxel. It is evident that the covariance is nearly orthogonal to the surface normal, implying that its eigenvector with the minimum eigenvalue aligns with the surface normal and eigenvalue itself representing the surface uncertainty. The remaining eigenvectors serve as shape primitives, forming a line in 2D or a plane in 3D as in Figure \ref{fig:downsampling_demo}.

By building on this insight, we can enhance down-sampling by extracting additional samples from eigenvectors other than the one corresponding to the minimum eigenvalue. Combining these additional samples with the mean of the samples enables us to obtain down-sampled points with uncertainty comparable to the original dataset, as depicted in the last row of Figure \ref{fig:downsampling_demo}. This eliminates the need for using noisy input data since the noise is already accounted for during covariance calculation. Alternatively, we can use the minimum eigenvalue as a proxy for the uncertainty of the observation. One drawback of this method is its increased demand for inputs—three times more in the case of 2D and five times more in 3D compared to raw down-sampling. While it offers an effective approach to preserve uncertainty during down-sampling, the higher input requirement represents a trade-off that should be carefully considered within the broader context of the application.

\begin{figure}[!tbp]
  \centering
  \includegraphics[width=\linewidth]{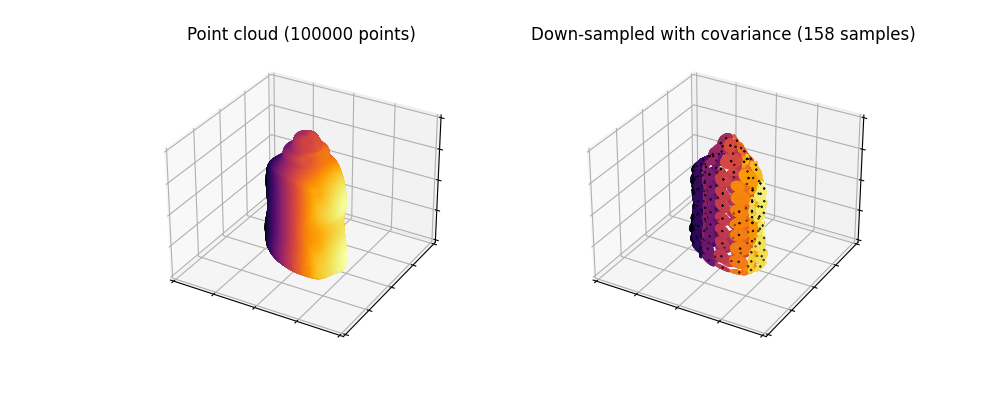}
  \caption{An illustration of voxel down-sampling. From 100,000 points on the object using a 2 cm unit voxel, we can get 158 samples, which will result in 640 points of input to GPDF that closely resemble the surface.}
  \label{fig:voxel_downsampling}
\vspace{-10pt}
\end{figure}
For an incremental voxel sampling update, we just have to keep three parameters, 13 values for 3D, for each voxel: the number of samples within the voxel, $n_{\text{sample}}$, the sum of the samples, $\sum^{n_{\text{sample}}}_i \mathbf{x}_{\text{sample},i}$, and the sum of the multiplication of the sample and its transpose, $\sum^{n_{\text{sample}}}_i \mathbf{x}_{\text{sample},i}\mathbf{x}_{\text{sample},i}^T$. Then the mean and covariance can be calculated as below:
\begin{equation}
    \begin{split}
        & \mu_{\text{sample}} = \frac{1}{n_{\text{sample}}}\sum^{n_{\text{sample}}}_i \mathbf{x}_{\text{sample}, i}\\
        & E(\mathbf{X_{\text{sample}}X_{\text{sample}}^T}) = \frac{1}{n_{\text{sample}}}\sum^{n_{\text{sample}}} \mathbf{x}_{\text{sample},i}\mathbf{x}_{\text{sample},i}^T \\
        & \Sigma_{\text{sample}} = E(\mathbf{X_{\text{sample}}X_{\text{sample}}^T}) - \mu_{\text{sample}}\mu_{\text{sample}}^T
    \end{split}
\end{equation}
An example of down-sampling can be seen in Figure \ref{fig:voxel_downsampling}. However, we still have the problem of not being able to deal with the noise of the different sensors. Therefore, we utilize the weighted version where we keep four parameters: the sum of the weights, $\sum^{n_{\text{sample}}}_i w_{i}$, the weighted sum of the samples, $\sum^{n_{\text{sample}}}_i w_i\mathbf{x}_{\text{sample},i}$, and the weighted sum of the multiplication of the sample and its transpose, $\sum^{n_{\text{sample}}}_i  w_i \mathbf{x}_{\text{sample}_i}\mathbf{x}_{\text{sample}_i}^T$. Then the weighted mean and covariance can be calculated as below:
\begin{equation}
    \begin{split}
        & \mu_{\text{sample}} = \frac{1}{\sum^{n_{\text{sample}}}_i w_{i}}\sum^{n_{\text{sample}}}_i \mathbf{x}_{\text{sample}, i}\\
        & E(\mathbf{W}\mathbf{X_{\text{sample}}X_{\text{sample}}^T}) = \frac{1}{\sum^{n_{\text{sample}}}_i w_{i}}\sum^{n_{\text{sample}}} w_i\mathbf{x}_{\text{sample},i}\mathbf{x}_{\text{sample},i}^T \\
        & \Sigma_{\text{sample}} = E(\mathbf{W}\mathbf{X_{\text{sample}}X_{\text{sample}}^T}) - \mu_{\text{sample}}\mu_{\text{sample}}^T
    \end{split}
\end{equation}
This incremental process can be done on a GPU using sparse data structures, which enables us to calculate things quickly and efficiently.


\section{Kernel approximations}\label{sec:kernel_approximation}
Gaussian Processes (GP) suffer from a notable drawback due to their limited scalability. This limitation arises from a computational complexity of $O(n^3)$ for computation and $O(n^2)$ for memory during training, where $n$ represents the number of observations. This scalability challenge becomes especially pronounced in visual observations, where handling thousands of pixel inputs is common. In contrast, tactile observations, characterized by sparse inputs of fewer than a thousand points, do not face the same level of scalability issues. To address this challenge, several approaches have been explored to approximate the training and inference processes of Gaussian processes.


All state-of-the-art approaches, which will be introduced throughout this section, that address scalability issues in Gaussian Processes involve approximating the kernel function within some range. These approximations follow the basic equation below:
\begin{equation}
\begin{split}
    & K(\mathbf{X}_1, \mathbf{X}_2) \approx A_{\mathbf{X}_1}BA_{\mathbf{X}_2}^T \\
\end{split}
\end{equation}
where $\mathbf{X}_1 \in \mathbb{R}^{n_1 \times 3}$, $\mathbf{X}_2 \in \mathbb{R}^{n_2 \times 3}$, $A_{\mathbf{X}_1} \in \mathbb{R}^{n_1 \times m}$, $A_{\mathbf{X}_2} \in \mathbb{R}^{n_2 \times m}$, and $B \in \mathbb{R}^{m \times m}$. Utilizing this approximation and Woodbury’s matrix identity, we obtain the approximate inverse matrix equation:
\begin{equation}
    \begin{split}
        & (K(\mathbf{X},\mathbf{X}) + D)^{-1} \approx (A_{\mathbf{X}}BA_{\mathbf{X}}^T + D)^{-1}\\
        & = D^{-1} -D^{-1} A_{\mathbf{X}}(B^{-1}+A_{\mathbf{X}}^T D^{-1} A_{\mathbf{X}})^{-1}A_{\mathbf{X}}^T D^{-1}
    \end{split}
\end{equation}
This approximation enables simplification of the inference and variance calculation of GP as follows:
\begin{equation}
    \begin{split}
        & \hat{o}(\mathbf{x_*})=k(\mathbf{x}_*,\mathbf{X})(K(\mathbf{X},\mathbf{X}) + D)^{-1}\mathbf{y}\\
        & \approx A_{\mathbf{x}_*}BA_{\mathbf{X}}^T(W_{\mathbf{X}}B A_{\mathbf{X}}^T + D)^{-1}\mathbf{y}\\
        & = A_{\mathbf{x}_*}(B^{-1}+A_{\mathbf{X}}^T D^{-1} A_{\mathbf{X}})^{-1}A_{\mathbf{X}}^T D^{-1}\mathbf{y}\\
        & \mathbf{var}(o(\mathbf{x_*})) = k(\mathbf{x_*},\mathbf{x_*}) - k(\mathbf{x_*},\mathbf{X})(K(\mathbf{X},\mathbf{X})+D)^{-1}k(\mathbf{X},\mathbf{x_*})\\
        & \approx A_{\mathbf{x}_*}(B-B A_{\mathbf{X}}^T(A_{\mathbf{X}}B A_{\mathbf{X}}^T + D))^{-1}A_{\mathbf{X}}B)A_{\mathbf{x}_*}^T\\
        & = A_{\mathbf{x}_*}(B^{-1}+A_{\mathbf{X}}^T D^{-1} A_{\mathbf{X}})^{-1}A_{\mathbf{X}}^T
    \end{split}
\end{equation}
This approximation reduces the complexity to $O(n)$ for both training and inference given $m\ll n$. In the subsequent subsections, we will elaborate on each kernel approximation method in detail. Towards the end of the section, we will compare these methods based on the trade-off between reconstruction accuracy and computational efficiency.


\subsection{Structured Kernel Interpolation (SKI)}
Structured Kernel Interpolation (SKI), also known as KISS-GP \cite{Stanton2021KernelIF}, employs a grid-based approach to approximate $K(\mathbf{X},\mathbf{X})\approx W_{\mathbf{X}}K(\mathbf{U},\mathbf{U})W_{\mathbf{X}}^T$. Here, $\mathbf{U}\in\mathbb{R}^{m\times D}$ represents the grid points, and $W_{\mathbf{X}}\in\mathbb{R}^{n\times m}$ serves as the weighting matrix for $\mathbf{X}$ with respect to $\mathbf{U}$. Typically, this weighting matrix is sparsely constructed using linear interpolation. One drawback of this approach is its reliance on a predefined task space where fixed grid points are placed. However, this method can be highly efficient since computing the weighting matrix is just interpolation with 8 neighboring points, which is computationally easy and sparse.

\subsection{Hilbert-space approximation}
Hilbert-space approximation takes a distinctive approach such that it does no rely on some reference points, it simplifies the kernel itself using its spectral density and $\Omega_{HS}$ which is the subset of input space $\mathbb{R}^D$ \cite{RiutortMayol2020PracticalHS}. For a uni-dimensional input
space $\Omega_{HS}\in{-L, L}\in\mathbb{R}$ and input values $x, x' \in\Omega_{HS}$, the kernel function can be written as
\begin{equation}
    \begin{split}
        & k(x,x')=\sum_{i=1}^{\infty} S(\sqrt{\lambda_i})\phi_i(x)\phi_i(x')\approx \sum_{i=1}^{m} S(\sqrt{\lambda_i})\phi_i(x)\phi_i(x')=\bm{\phi}(x)\bm{\Delta}\bm{\phi}(x')^T 
    \end{split}
\end{equation}
where $S$ is spectral density function listed in Appendix \ref{covariance_kernels}, $\lambda_i=\left(\frac{i\pi}{2L}\right)^2$, $\phi_i(x)=\sqrt{1/L}\sin(\sqrt{\lambda_i}(x+L))$, $\bm{\phi}(x)=[\phi_1(x),\cdots,\phi_m(x)]$, and $\bm{\Delta}=diag(S(\sqrt{\lambda_1}),\cdots,S(\sqrt{\lambda_m}))$. Therefore, we get $K(\mathbf{X},\mathbf{X})\approx\bm{\Phi}\bm{\Delta}\bm{\Phi}^T$ where $\bm{\Phi}=[\bm{\phi}(\mathbf{x}_1),\cdots,\bm{\phi}(\mathbf{x}_n)]^T$. The multi-dimensional input case is also very similar as described in \cite{RiutortMayol2020PracticalHS}. Similar to SKI, the Hilbert-space approximation yields a kernel matrix with a comparable form and shares the limitation related to the boundary. One crucial consideration is the relationship between the hyperparameter $m$ and $L$. Typically, the value of $L$ needs to be 1.2 times longer per dimension than the input values for accurate results. Moreover, as the length of $L$ increases, a higher value of $m$ becomes necessary. This connection is comprehensively explained in \cite{RiutortMayol2020PracticalHS}.

\subsection{Nystr\"om approximation}
Nystr\"om approximation \cite{Ounpraseuth2008GaussianPF} uses reference points, $\mathbf{\hat{X}}$, which do not have to be on the grid. Instead, it uses eigen-decomposition of kernel matrix of reference points, $K(\mathbf{\hat{X}},\mathbf{\hat{X}})$, to perform approximation. For input values $x,x'\in\mathbb{R}^D$, the kernel function can be approximated as
\begin{equation}
    \begin{split}
        & k(x,x')=\sum_{i=1}^{\infty} \mu_i\psi_i(x)^T\psi_i(x')\approx \sum_{i=1}^{l} \mu_i\psi_i(x)^T\psi_i(x') \\
        & \psi_i(x)=\sum_{j=1}^m k(x, \hat{x}_j) \bm{\phi}_{i,j}\quad,\quad\mu_i=1/\lambda_i \\
        & K(\mathbf{X},\mathbf{X})\approx \mathbf{\Psi}(\mathbf{X})\bm{M}\mathbf{\Psi}(\mathbf{X})^T =K(\mathbf{X},\mathbf{\hat{X}})\mathbf{\Phi}\bm{M}\mathbf{\Phi}^TK(\mathbf{\hat{X}},\mathbf{X}) \\
        & \mathbf{\Phi}=[\phi_1, \cdots, \phi_l] \quad,\quad \bm{M}=diag(\bm{\mu}) \quad,\quad \bm{\mu}=[\mu_1, \cdots, \mu_l]
    \end{split}
\end{equation}
where $\{(\lambda_i, \bm{\phi}_i)\}_{i\in\{1,\cdots,l\}}$ are the eigen-pairs with rank $l$ of $K(\mathbf{\hat{X}},\mathbf{\hat{X}})$ \cite{Yuan2023UniFusionUC}. In selecting reference points, it's essential for them to be representative of the data points. Options include choosing from a grid, uniform sampling within the boundary, or selecting some from observations. While this method is also susceptible to the curse of dimensionality, a noteworthy point is that the parameter $l$ does not necessarily have to be equal to the number of reference points. This flexibility provides additional avenues to enhance computational efficiency.

\subsection{Comparison}
\begin{figure}[!tbp]
  \centering
  \includegraphics[width=1.0\linewidth]{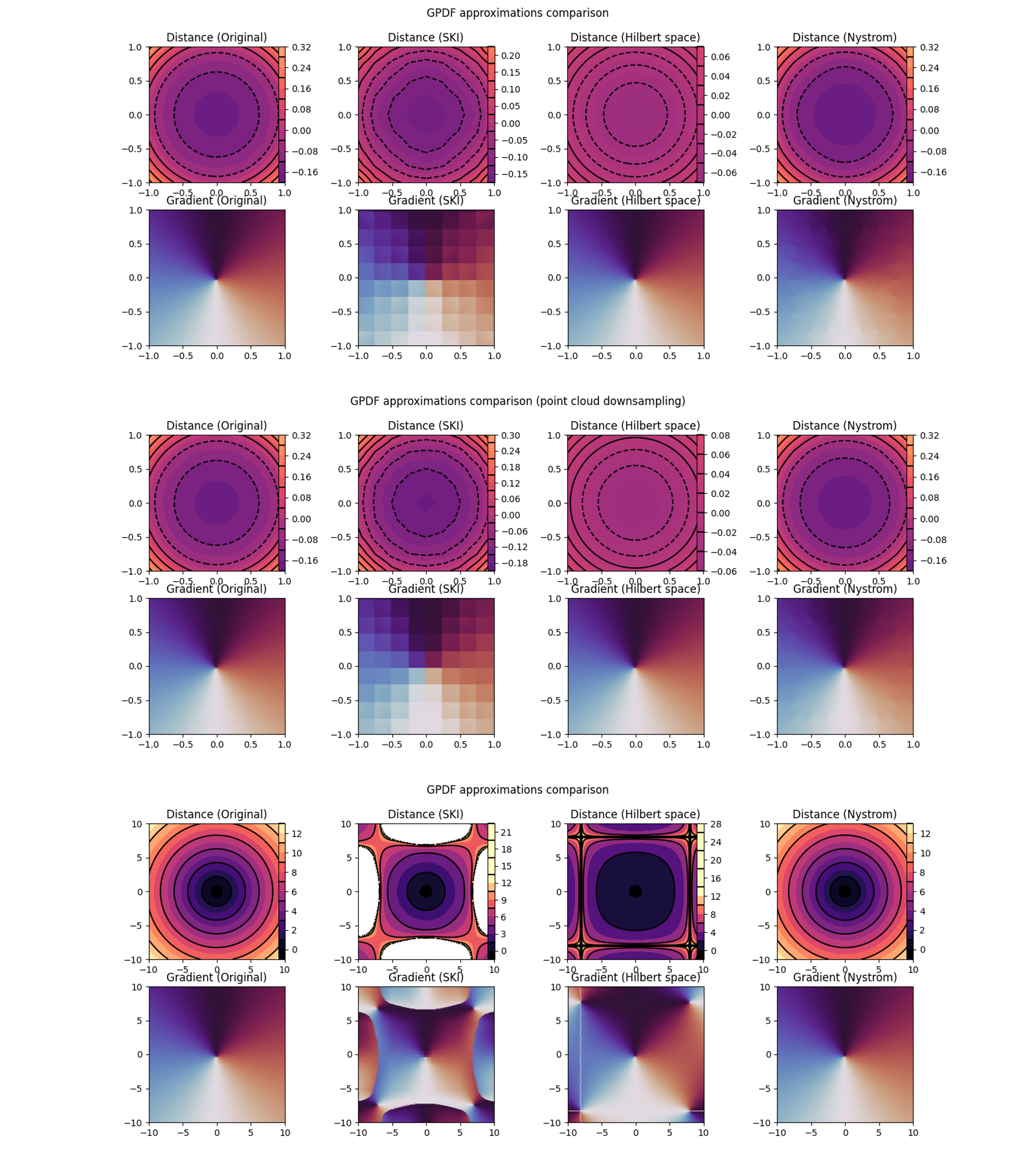}
  \caption{GPDF results (1-2 row: within the approximation bounding box, 3-4 row: with point cloud approximation, 5-6 row: outside the approximation bounding box)}
  \label{fig:comparison_results_combined}
\vspace{-10pt}
\end{figure}
The visualization of each kernel approximation for Gaussian Process Distance Function (GPDF) for a sphere are demonstrated in Figure \ref{fig:comparison_results_combined}. For all approximation methods, the bounding box was set to [-5, 5] for both axes, and the grid size was 41 for each axis, resulting in a grid size of 0.25x0.25. Within the bounding box, most of the results were fairly satisfactory, except for the SKI method, which exhibited a very non-continuous gradient, and the Hilbert space method, which produced a less accurate approximation of distance function and its quality heavily dependent on the training points showing inconsistency.
For inference outside the bounding box, only the Nystr\"om method was able to provide a good distance function and gradient, as expected from the formulation. For point cloud approximation, we used 0.1 size of grid, which converted original 3000 training points into 204 points. When used with the SKI and Hilbert space methods, we observed minor discrepancies in the queried distance, but overall, they closely resembled the results obtained without point cloud approximation. This shows that point cloud approximation can effectively used with all kernel approximation methods.

\begin{figure}[tb]
    \begin{minipage}[c]{0.48\linewidth}
        \centering
        \includegraphics[width=1.\linewidth]
        {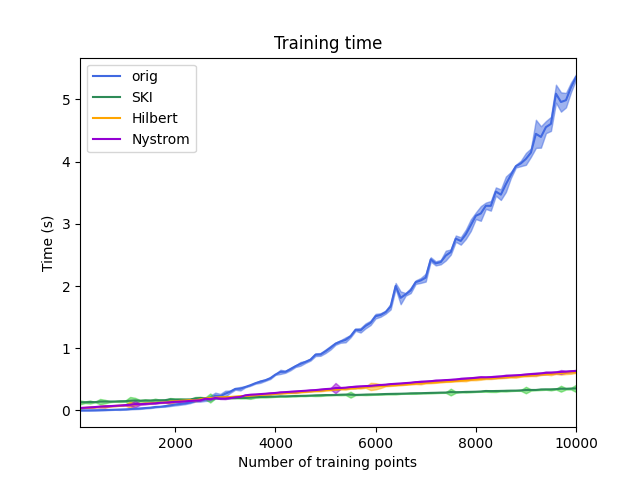}
        \caption{Training time comparison between kernel approximation methods}
        \label{fig:comparison_training_time}
    \end{minipage}
    \hfill
    \begin{minipage}[c]{0.48\linewidth}
        \centering
        \includegraphics[width=1.0\linewidth]
        {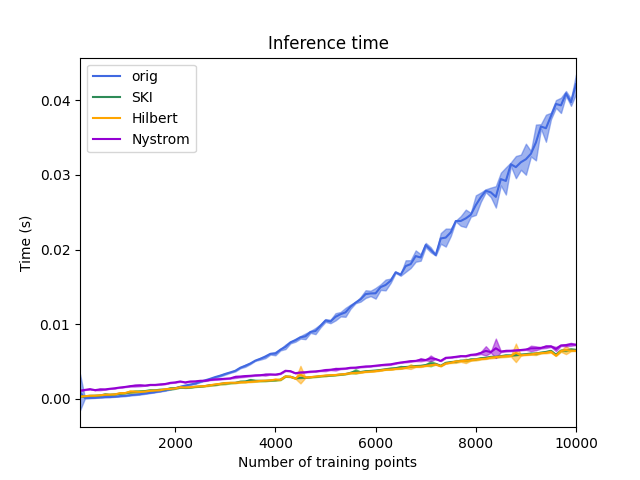}
        \caption{Inference time comparison between kernel approximation methods}
        \label{fig:comparison_inference_time}
    \end{minipage}%
\end{figure}
To test computation speed, we conducted experiments using all methods with a single core and observation or training points generated from the surface of a sphere with a varying number. Inference time is measured for single query point. We tested both training and inference time and results are show in Figure \ref{fig:comparison_training_time} and Figure \ref{fig:comparison_inference_time}.
In terms of computational speed, SKI is the fastest method for both training and inference, achieving less than 0.1 seconds for training and 2-3 milliseconds for inference. This is followed by the Hilbert space method and Nystr\"om method, respectively, which are usually 3-4 times slower than SKI. However, for scenarios involving numerous inference points simultaneously, the Nystr\"om method outperformed the Hilbert space method, primarily due to the latter's reliance on complex spectral decomposition processes.

In conclusion, for datasets containing more than 2000 data points, if a smooth distance field is not necessary and results within the boundary are acceptable, SKI is a good option. However, if smoothness is a crucial factor and distance outside the approximation region is required, the Nystr\"om method becomes a preferable choice. In our case, we use original GP for small number of data points or with point cloud approximation. For a large number of data points, we utilize the Nystr\"om method since it closely resembles the original GP and ensures the smoothness of gradients, which is crucial for tactile exploration.

\section{Advantages of GPDF over Neural Network based techniques}
This section outlines the benefits of Gaussian Process Distance Field (GPDF) compared to neural network-based approaches in active perception for robotics.
\begin{enumerate}
    \item \textbf{Distance Measurement:} GPDF inherently provides accurate distance measurements, even at extended ranges from objects. Neural networks require additional regularization, such as Eikonal loss or smoothness regularization, to approximate a distance function and tend to be less accurate with increasing distance from objects. GPDF also supports controllable interpolation, allowing it to model occlusion effects as in Section \ref{sec:exploration}, which many implicit representation methods cannot do.
    \item \textbf{Multi-modal Data Integration:} As shown in Chapter \ref{sec:exploration}, GPDF can effortlessly integrate different types of multi-modal data, unlike neural networks, which typically need separate models for each data type. This adaptability makes GPDF a good fit for applications where neural networks face challenges, such as incorporating material properties, while being equally capable in tasks like differentiable rendering and information gain quantification.
    \item \textbf{Conversion of Point Clouds to Distance Fields:} As explained in Section \ref{sec:GPDF}, GPDF can convert point cloud data into distance fields. Unlike neural networks, which must choose between using fixed conversion functions or flexible weights for optimization, GPDF can do both at the same time. This adaptability enables accurate modeling from point cloud data and can easily accommodate dynamic changes, like deformable objects or structural breakage.
    \item \textbf{Surface Uncertainty Quantification:} GPDF inherently quantifies uncertainty on the data surface without the need for additional training or complex techniques. This built-in uncertainty measure is invaluable for guiding robotic exploration and assessing risk, as explained in Section \ref{sec:GPDF}.
    \item \textbf{Incremental Updates:} When adding new data, neural networks generally need a full re-training or re-optimization of the entire model. GPDF, on the other hand, allows for incremental updates by optimizing and aggregating new data points, providing a more efficient approach. This feature, discussed in Section \ref{sec:incremental_update}, is particularly useful when integrating tactile data, where each observation is valuable and re-training a full model is often impractical.
\end{enumerate}
Sections \ref{sec:heristic_approximation} and \ref{sec:kernel_approximation} address GPDF's main drawback—computational complexity—with approximation techniques, like voxel-based methods for smaller data sets and the Nystr\"om method for larger ones. We use these techniques in Chapter \ref{sec:exploration} to maintain high performance while keeping computational demands.

Overall, the GPDF framework offers a simpler, unified approach that integrates key functions like distance measurement, multi-type data integration, shape conversion, uncertainty quantification, and incremental updates. This leads to more consistent results, lower computational costs, and greater adaptability in robotics and active perception.

\chapter{Geometry Exploration from Vision and Touch}\label{sec:exploration}


In this chapter, we describe different ways to measure uncertainty of surface or object geometry and propose exploration strategies to guide robot motions to maximize information gain using the GPDF surface representation described in detail in Chapter \ref{sec:shape_representation}. In Section \ref{sec:shape_exploration_vision}, we introduce a method to measure information gain of Signed Distance Functions (SDF) and outline surface exploration strategies for vision. Subsequently, in Section \ref{sec:surface_exploration_touch}, we integrate tactile information with vision into the exploration pipeline. Finally, Section \ref{sec:integrating_vision_touch} discusses the benefits of closely combining visual and tactile exploration methods.
\section{Shape exploration from vision} \label{sec:shape_exploration_vision}
We investigate the reconstruction of geometry using vision. Initially, we explore how differentiable rendering methods can optimize inducing points for GPDF, even in cases where there is no direct point cloud observation or where parts of the object are not well captured by the depth camera, such as reflective or translucent materials. This approach extends the inducing point method discussed in Chapter \ref{sec:shape_representation}. Since the original KL divergence needs the set of a point and its observation, this new method provides an additional avenue for enhancement. Depending on the situation, we can use differentiable rendering to optimize point clouds, or use the previous inducing point method with point cloud approximation, or even combine both methods using differentiable rendering and KL divergence. However, since we use differentiable rendering to quantify information gain, we mainly focus on this approach throughout this section. In the later part of the section, we explore how to incorporate high-dimensional modalities and controllers for implementation, including simulation and real robot experiments.

\subsection{Rendering Signed Distance Function}\label{subsec:rendering}
\begin{figure}[tb]
    \begin{minipage}[c]{0.48\linewidth}
        \centering
        \includegraphics[width=1.\linewidth]
        {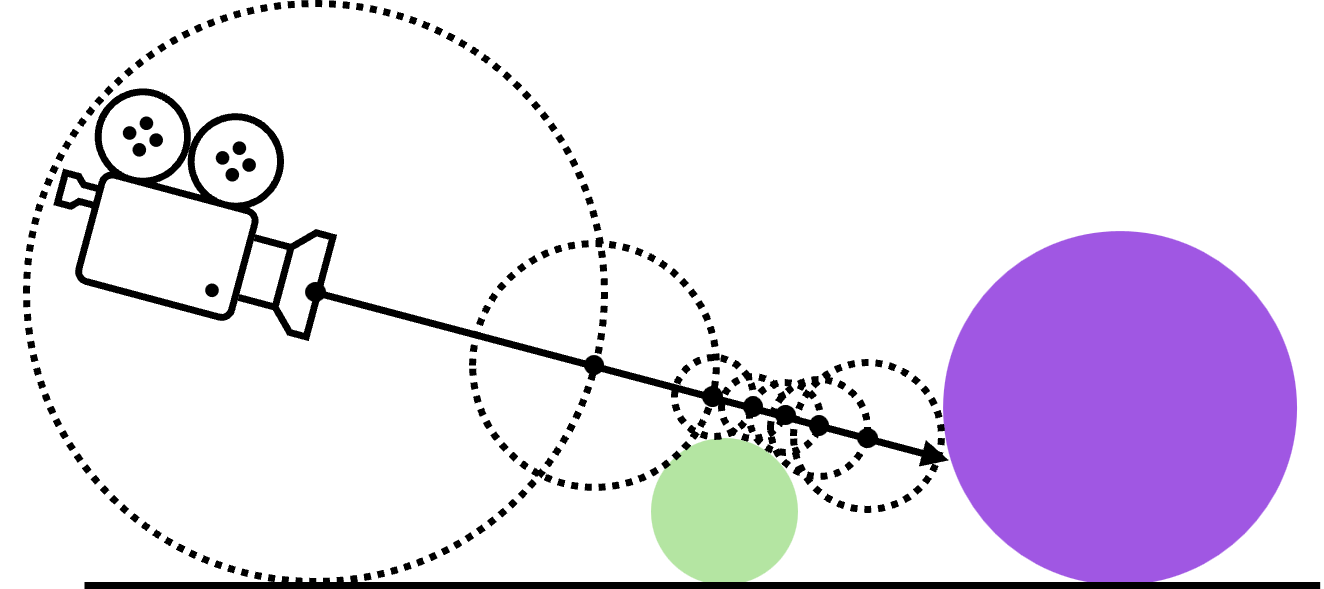}
        \caption{Illustration of sphere ray marching}
        \label{fig:ray_marching_sphere}
    \end{minipage}
    \hfill
    \begin{minipage}[c]{0.48\linewidth}
        \centering
        \includegraphics[width=1.0\linewidth]
        {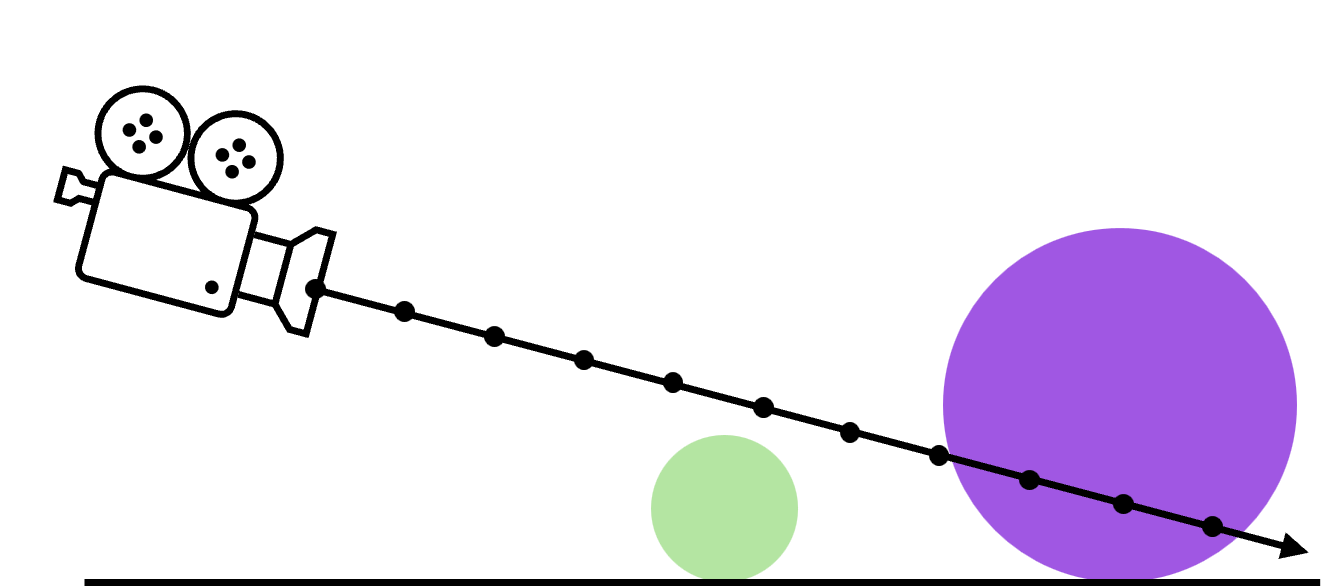}
        \caption{Illustration of volumetric ray marching}
        \label{fig:ray_marching_volumetric}
    \end{minipage}%
\end{figure}
To render Signed Distance Functions (SDF), we employ ray marching techniques to determine where rays intersect with surfaces. A common method is distance-aided ray marching, such as sphere tracing (see Figure \ref{fig:ray_marching_sphere}). This technique facilitates rapid ray propagation by utilizing Lipschitz bounds \cite{Galin2020SegmentTU, Aydinlilar2023ForwardIF}, which involves querying the distance at the ray's endpoint. When a ray intersects with a surface, i.e., it reaches a distance within a certain threshold, we can query surface properties like normal or color at the intersection point to generate the image.
Let's denote the ray casting equation as $\mathbf{x}_r=\mathbf{r}_0+\mathbf{r}_d D$, where:
\begin{itemize}
    \item $\mathbf{x}_r$ is the surface point of the object first intersected by the ray,
    \item $\mathbf{r}_0$ is the starting point of the ray,
    \item $\mathbf{r}_d$ is the ray's direction, and
    \item $D$ is the distance to the intersection point on the object's surface.
\end{itemize}
The values of $\mathbf{r}_0$ and $\mathbf{r}_d$ depend on the camera's pose, while $D$ is determined iteratively through sphere tracing. This iterative process poses challenges for differentiation due to the complexity of the distance metric $D$. Even after just two iterations, $D$ takes a complex form.
\begin{equation}
    \begin{split}
        & D=d(\mathbf{r}_0)+d(\mathbf{r}_1)+d(\mathbf{r}_2) \\
        & \mathbf{r}_1 = \mathbf{r}_0 + \mathbf{r}_d d(\mathbf{r}_0) \\
        & \mathbf{r}_2 = \mathbf{r}_0 + \mathbf{r}_d (d(\mathbf{r}_0)+d(\mathbf{r}_1))\\
        & d(\cdot ) \quad\text{(SDF)}
    \end{split}
\end{equation}
For $n-1$ steps it becomes,
\begin{equation}
    \begin{split}
        & d_n=\sum_{i=0}^n f(\mathbf{r}_i) \quad,\quad \mathbf{r}_i=\mathbf{r}_0+\mathbf{r}_d\sum_{j=0}^{i-1}d_j\
    \end{split}
\end{equation}
The iterative process continues until the ray approaches the surface at a distance below a specified threshold. Consequently, each ray point undergoes a variable number of iterations. Computing the gradient of this expression becomes notably less manageable or efficient, particularly with multiple iterations. Therefore, evaluating the rendered image through sphere marching is more akin to a sampling approach rather than optimization through gradient computation.

Another approach is to utilize volumetric rendering, as illustrated in Figure \ref{fig:ray_marching_volumetric}. This method relies not only on the Signed Distance Function (SDF) of the object but also on colors or other properties defined in Euclidean space. Wang et al. \cite{Wang2021NeuSLN} introduced a volumetric rendering method with SDF very similar to NeRF \cite{Mildenhall2020NeRFRS} as explained in Chapter \ref{sec:introduction}. With GPDF without the reverting function, we can have an approximate occupancy function, but here we demonstrate using SDF directly. Assuming we have equidistant points on the ray sequentially just like NeRF \cite{Mildenhall2020NeRFRS} as in Equation \ref{eq:nerf}, defined as $\mathbf{x}_{t_i}=\mathbf{r}_0+\mathbf{r}_d t_i$, where $t_{min}\leq t \leq t_{max}$ and $i$ is the index of points incrementally increasing, we obtain: 
\begin{equation}
\begin{split}
    & \Phi_s(\cdot) = (1+e^{-s (\cdot)})^{-1} \quad\text{(Sigmoid function)}\\
    & \alpha_i = \max\left(\frac{\Phi_s(d(\mathbf{x}_{t_i})) - \Phi_s(d(\mathbf{x}_{t_{i+1}}))}{\Phi_s(d(\mathbf{x}_{t_i}))}, 0\right) \quad\text{(Opacity)}\\
    & T_i = \Pi_{j=1}^{i-1}(1-\alpha_j) \quad\text{(Transmittance)}\\
    & \hat{C} = \sum_{i=1}^n T_i \alpha_i c_i \quad\text{(Estimated pixel color)} \\
    & \hat{D} = \sum_{i=1}^n T_i \alpha_i d(\mathbf{x}_{t_i}) \quad\text{(Estimated pixel depth)}
\end{split}
\end{equation}
Here, $n$ represents the number of points on the ray, and $s$ is a hyperparameter controlling the smoothness of the sigmoid function, typically set as a high positive constant. This formulation allows for differentiable rendering of color, depth, and other incorporated properties. A drawback of this method is its requirement for many points on the ray to achieve accurate estimation. While an adaptive ray point sampling strategy could ease this drawback, in our approach, we utilize the boundary of the task space to reduce the total length of the ray. This results in denser sampling, thereby improving reconstruction accuracy.

\begin{figure}[tb]
        \centering
        \includegraphics[trim={3.5cm 0 3.5cm 0},clip, width=1.0\linewidth]
        {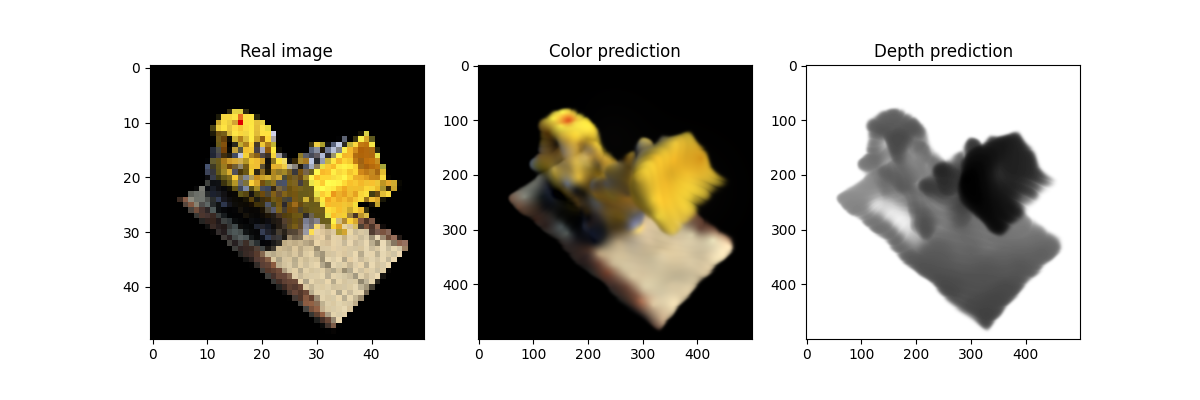}
        \caption{Optimized GPDF using multi-view color images}
        \label{fig:gpdf_fit}
\end{figure}
\begin{figure}[tb]
        \centering
        \includegraphics[width=1.0\linewidth]
        {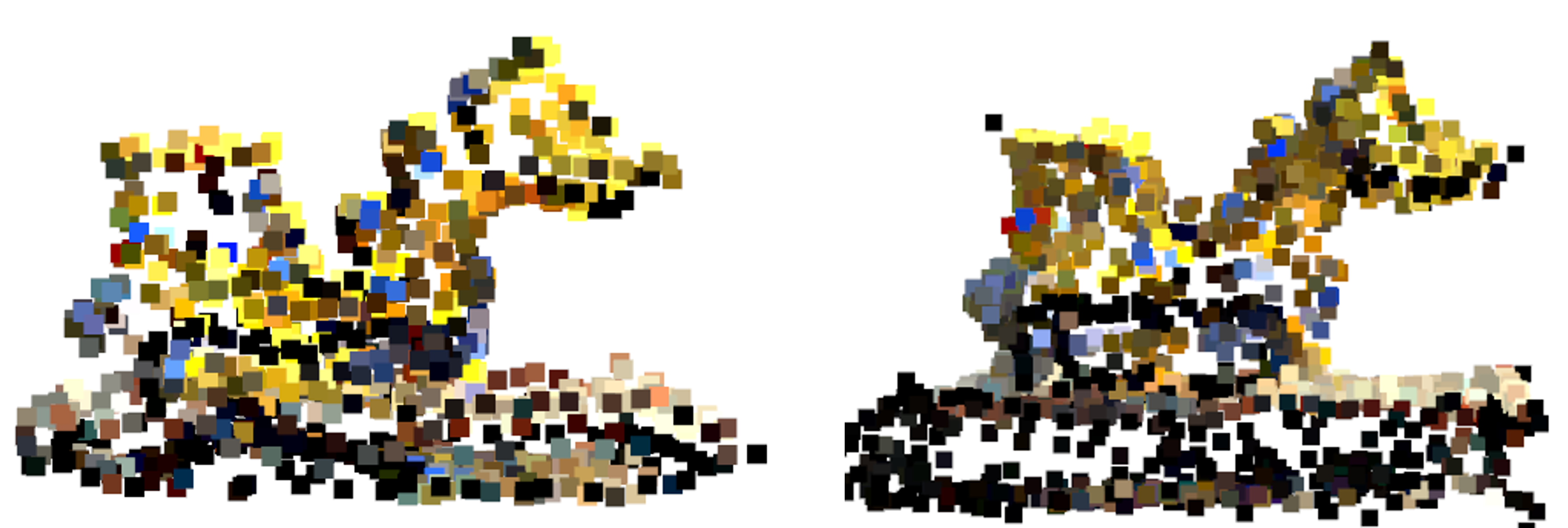}
        \caption{Point cloud obtained from optimized GPDF (left: initialized with 1000 points, right: initialized with 2000 points)
        }
        \label{fig:gpdf_fit_pc}
\end{figure}


Using the volumetric differentiable rendering approach and GPDF, we can optimize the inducing points of GPDF to align with camera views, as depicted in Figure \ref{fig:gpdf_fit}. Only color images with known camera poses are necessary to generate a point cloud of an object, as illustrated in Figure \ref{fig:gpdf_fit_pc}. 
As more inducing points are used, not only are more intricate details acquired, but also the accuracy of geometry estimation improves. This can be observed in Figure \ref{fig:gpdf_fit_pc} by noting the flatness of the ground upon which the LEGO bulldozer is positioned. In the toy LEGO example shown in the figures, we omitted the use of sphere harmonics to model the direction of the rays. Consequently, the rendering produces a somewhat monocolor blurred image. Nevertheless, it successfully reconstructs a point cloud, enabling us to apply traditional refinement techniques to eliminate spurious points or make modifications as needed. GPDF can also be optimized using depth images or both color and depth images.

\subsection{Information gain of camera poses}
Differentiable rendering enables the preservation of past observations through optimization and the generation of various camera views. We can quantify the information gain of camera poses, as defined by He et al. \cite{He2023ActivePU}. Given camera pose $x$ and observation $y$, the information gain is expressed as:
\begin{equation}
    \begin{split}
        & IG(y_{\rm future}, y_{\rm past}) = H(y_{\rm future}) - \int dp(y_{\rm past}) H(y_{\rm future}|y_{\rm past}) \\
        & = H(y_{\rm future}) - \int dp(\xi) H(y_{\rm future}|\xi)
    \end{split}
\end{equation}
where $\xi$ denotes the true scene of the environment, and $H$ represents Shannon entropy. $p(\xi)$ and Shannon entropy are calculated using an ensemble of generative models like NERF, as follows:
\begin{equation}
    \begin{split}
        & p(\xi) = \frac{1}{m} \sum_{k=1}^m \delta_{\xi_k}(\xi) \\
        & H(y_{\rm future}|\xi) \approx -p(y_{\rm future}|\xi) \log p(y_{\rm future}|\xi)
    \end{split}
\end{equation}
where $\delta$ is the Dirac-delta function, and $m$ is the number of ensembles. He et al. modeled color and depth observations as Gaussian distributions, each with a variance defined as:
\begin{equation}
    \begin{split}
        & \mathbf{var}(\hat{C}) = \sum_{i=1}^n T_i \alpha_i (c_i - \hat{C})^2 \\
        & \mathbf{var}(\hat{D}) = \sum_{i=1}^n T_i \alpha_i (d_i - \hat{D})^2 \\
    \end{split}
\end{equation}
\begin{figure}[!tbp]
  \centering
  \includegraphics[trim={2cm 0 2cm 0},clip, width=1.0\linewidth]{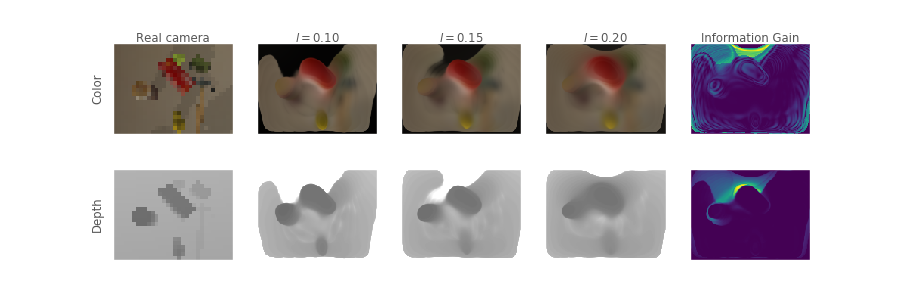}
  \caption{Information gain from GPDF ensemble with varying interpolations in simulation, trained with frontal view (column 1: real camera data, column 2-4: GPDFs with different interpolations, column 5: depth information gain)}
  \label{fig:ensemble_interpolation}
\vspace{-10pt}
\end{figure}
We adopt the same ensemble approach for GPDF since GPDF can act as a generative model for camera views. However, the issue arises from the inability of SDF to effectively model occupancy, which is crucial for dealing with occlusions. To address this challenge, we utilize multiple GPDFs with varying interpolation values as in Figure \ref{fig:ensemble_interpolation}. This strategy enables us to identify areas most impacted by changes in interpolation values, thus indicating a higher likelihood of occluded regions. Alternatively, instead of training multiple GPDFs, we can train a single GPDF incorporating both inducing points and interpolation values. This allows us to represent them as Gaussian distributions, facilitating the rendering of different images by adding noises to the inducing points and interpolation values.

\subsection{Multi-modality}
\begin{figure}[!tbp]
  \centering
  \includegraphics[width=1.0\linewidth]{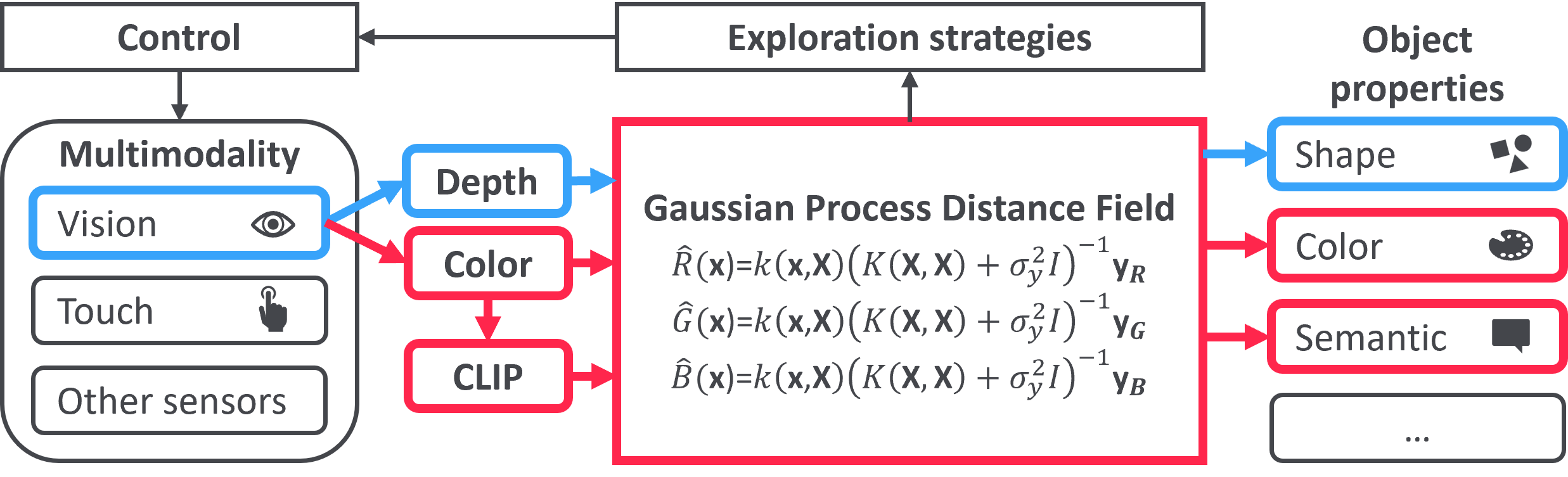}
  \caption{Multimodal incorporation in Gaussian Process Distance Field}
  \label{fig:multimodal_visual_exploration}
\vspace{-10pt}
\end{figure}

To compute the occupancy field in GPDF, we set $\mathbf{y} = \mathbf{1}$. The estimated occupancy at a point $\mathbf{x_*}$ in a 3D Euclidean space is then given by:
\begin{equation}
    \hat{o}(\mathbf{x_*}) = k(\mathbf{x_*}, \mathbf{X})  (K(\mathbf{X}, \mathbf{X}) + \sigma_y^2 I)^{-1} \mathbf{y}
\end{equation}
where $\mathbf{X}$ is a set of training points, and $\mathbf{y}$ is their corresponding labels. The same formulation can be used to infer other characteristics by modifying $\mathbf{y}$. For instance, if we use training data points with RGB color values—$y_R$, $y_G$, $y_B$—normalized between 0 and 1, we can derive RGB color fields. This flexibility allows us to compute various feature fields within the same GPDF framework as in Figure \ref{fig:multimodal_visual_exploration}. The following formulas represent the Gaussian Process-based estimates for RGB color values at a given point $\mathbf{x_*}$ in a 3D space:
\begin{equation}
    \begin{split}
        & \hat{R}(\mathbf{x_*})=k(\mathbf{x_*},\mathbf{X})(K(\mathbf{X},\mathbf{X})+\sigma_y^2 I)^{-1}\mathbf{y}_R\\
        & \hat{G}(\mathbf{x_*})=k(\mathbf{x_*},\mathbf{X})(K(\mathbf{X},\mathbf{X})+\sigma_y^2 I)^{-1}\mathbf{y}_G\\
        & \hat{B}(\mathbf{x_*})=k(\mathbf{x_*},\mathbf{X})(K(\mathbf{X},\mathbf{X})+\sigma_y^2 I)^{-1}\mathbf{y}_B\\
    \end{split}
\end{equation}
The vectors $\mathbf{y}_R$, $\mathbf{y}_G$, and $\mathbf{y}_B$ contain the corresponding RGB values for these training points. With this formulation, you can derive RGB estimates from existing training points. Alternatively, you can optimize the RGB values of inducing points based on given observations, allowing for improved color interpolation results, as shown in Appendix \ref{interpolation_parameter}.

\begin{figure}[!tbp]
  \centering
  \includegraphics[width=0.5\linewidth]{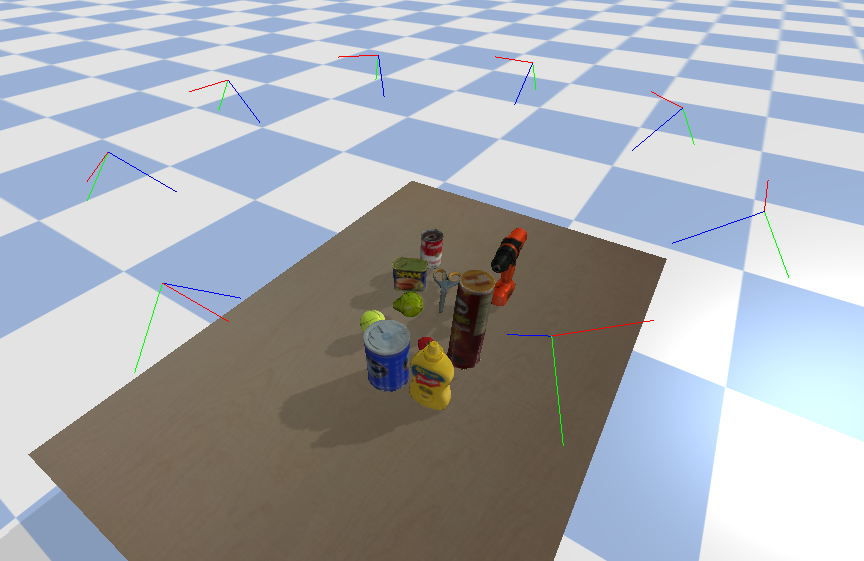}
  \caption{Simulation setup of YCB datasets on the table with 8 different camera poses}
  \label{fig:ycb_table}
\vspace{-10pt}
\end{figure}

\begin{figure}[!tbp]
  \centering
  \includegraphics[width=1.\linewidth]{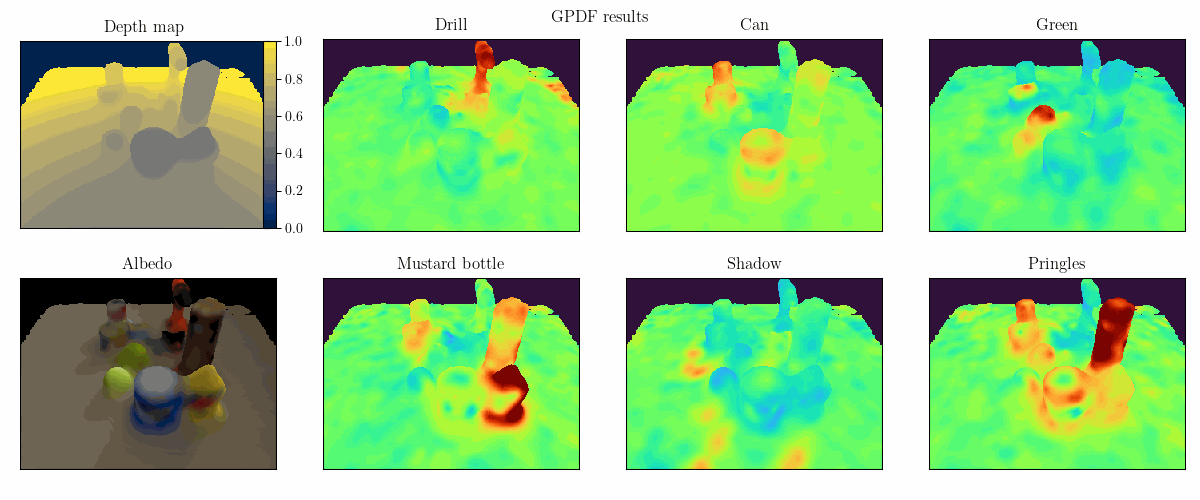}
  \caption{Rendered image of GPDF and CLIP features compared with the language embedding}
  \label{fig:clip_image}
\vspace{-10pt}
\end{figure}

This methodology is not limited to color; it can also be applied to high-dimensional features such as CLIP \cite{Radford2021LearningTV} embeddings with 768 long feature vector, which encompass both visual and language modalities. Recent research has demonstrated that CLIP embeddings can be transformed into dense features, facilitating tasks such as language-guided robot manipulation \cite{Shen2023DistilledFF}. To estimate high-dimensional features like CLIP embeddings using Gaussian Processes, we use training dataset with points $\mathbf{X}$ in a 3D space, and a corresponding matrix $\mathbf{Y}_{\rm CLIP}\in\mathrm{R}^{n\times768}$, where each row in $\mathbf{Y}_{\rm CLIP}$ represents a high-dimensional feature vector for each training point. For a given test point $\mathbf{x_*}$, the GP-based estimate for its high-dimensional feature vector, $\hat{\mathbf{y}}(\mathbf{x_*})$, can be obtained by extending the earlier formulation:
\begin{equation}
    \hat{\mathbf{y}}_{\rm CLIP}(\mathbf{x_*}) = k(\mathbf{x_*}, \mathbf{X}) (K(\mathbf{X}, \mathbf{X}) + \sigma_y^2 I)^{-1} \mathbf{Y}_{\rm CLIP}.
\end{equation}
To evaluate this capability, we created a Pybullet simulation environment using the YCB datasets \cite{alli2017YaleCMUBerkeleyDF}, as depicted in Figure \ref{fig:ycb_table}.
The process of creating a CLIP field using Gaussian processes involves the following steps:
\begin{enumerate}
    \item Capture RGBD Images: Obtain RGBD images from various poses of the scene.
    \item Extract Dense CLIP Embeddings: Utilize the RGB images to generate dense CLIP embeddings \cite{Shen2023DistilledFF}, scaling them back to match the original image size.
    \item Conversion to Point Cloud: Transform the RGBD+CLIP images into a point cloud representation, incorporating both RGB and CLIP features.
    \item Field Model Generation: Develop field models for each RGB color and the 768 features in the CLIP embedding using Gaussian processes.
    \item Rendering Process: To render views from any camera pose:
    \begin{enumerate}
        \item Perform sphere marching using GPDF to determine hit points. Check Chapter \ref{subsec:rendering} for the equation. 
        \item Conduct inference on the obtained hit points using the appropriate field models for both RGB and CLIP features.
    \end{enumerate}
\end{enumerate}
This methodology facilitates the generation of rendered views from any camera pose by leveraging the combined information of RGB and CLIP features through Gaussian processes. In Figure \ref{fig:clip_image}, you can observe not only the rendered color and depth map of a camera pose but also the cosine similarity between the image CLIP embedding and the language CLIP embedding. This visualization highlights the image in red when the word and the corresponding part of the image are similar, providing valuable insights into the semantic alignment between visual and language modalities.
Indeed, Gaussian processes serve as a valuable tool for embedding semantic information into geometry, establishing a bridge between semantic information and traditional robotic algorithms. By integrating multimodal semantic information, such as object surface properties like roughness or hardness, into a unified format like Gaussian processes, we can facilitate a more holistic understanding of the environment. This approach enables robotic systems to leverage both geometric and semantic knowledge, thereby enhancing their capabilities in various tasks such as perception, planning, and manipulation.


\subsection{Low-level controller}\label{sec:controller}
\begin{figure}[!tbp]
  \centering
  \includegraphics[width=1.0\linewidth]{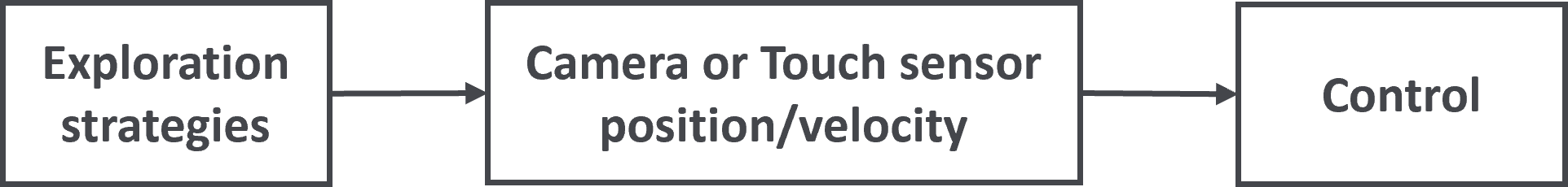}
  \caption{Control based on the Information Gain from exploration strategies}
  \label{fig:exploration_control_pipeline}
\vspace{-10pt}
\end{figure}
Both vision and tactile exploration involve targeting either a camera pose or a contact point with a normal direction in the task space of the robot as in Figure \ref{fig:exploration_control_pipeline}. To achieve this, we employ a simple quadratic programming (QP) instantaneous velocity controller for solving inverse kinematics and managing robot joint motion and self-collision similar to the QP-Ik solver proposed in \cite{Salehian2018AUF}. The formulation is as follows:
\begin{equation}
\begin{split}
\min_{\dot q} \quad&\|\dot x - \dot x_{\rm ref}\|2\\
\text{s.t.}\quad & \dot x = J(q) \dot q \\
&-\dot q \ge -\alpha (q_{\rm max} - q),\\
&\dot q \ge -\alpha (q - q_{\rm min}),\\
&\nabla_q d_{\rm ij}(q) \dot q \ge -\beta(d_{\rm ij}(q) - \epsilon)
\end{split}
\end{equation}
Here, $\dot x$ and $\dot x_{\text{ref}}$ represent the velocity and its reference vector in the task space, applicable to camera or tactile sensor velocities. Reference velocity can be an proportional value of difference with reference pose $x_{\rm ref}$ from exploration strategies such as camera pose or point of contact with surface normal information and current pose $x_{\rm cur}$. $q$ and $\dot q$ denote the robot's joint position and velocities, respectively, with $q_{\text{min}}$ and $q_{\text{max}}$ indicating the minimum and maximum joint limits. $J(q)$ represents the Jacobian matrix of forward kinematics, where $x = FK(q)$. $d_{\text{ij}}(q)$ indicates the minimum distance between two links, $i$ and $j$, where $|i-j| > 1$, within the set ${1, 2,..., n_l}$, with $n_l$ being the number of sequentially connected links. $\epsilon$ denotes a safety distance threshold. $\alpha(\cdot)$ and $\beta(\cdot)$ are strictly increasing functions that goes through the zero. If we can measure the distance a robot links to the environment, we can also add a safety constraint inside the QP formulation, similar to the self-collision constraint. However, before exploration of geometry of the environment, this is not known.
\begin{figure}[tb]
    \centering
    \includegraphics[width=0.5\textwidth]{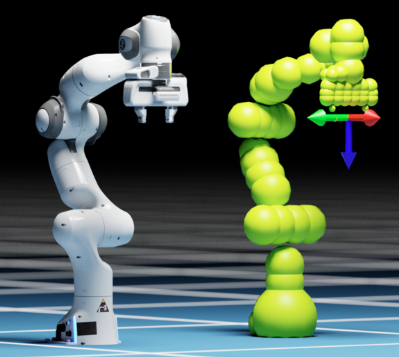}
    \caption{Sphere collision model of a robot from cuRoBo \cite{Sundaralingam2023CuRoboPC}}
    \label{fig:robot_representation}
\end{figure}
To assess the distance between links, we use the sphere collision model introduced by cuRoBo \cite{Sundaralingam2023CuRoboPC}, utilizing the closest sphere set between the links. The gradient of this distance with respect to joints can be computed either using the chain rule with the forward kinematics Jacobian of the sphere or through numerical methods.

\subsection{Implementation}
\begin{figure}[tb]
    \centering
    \includegraphics[width=1.0\textwidth]{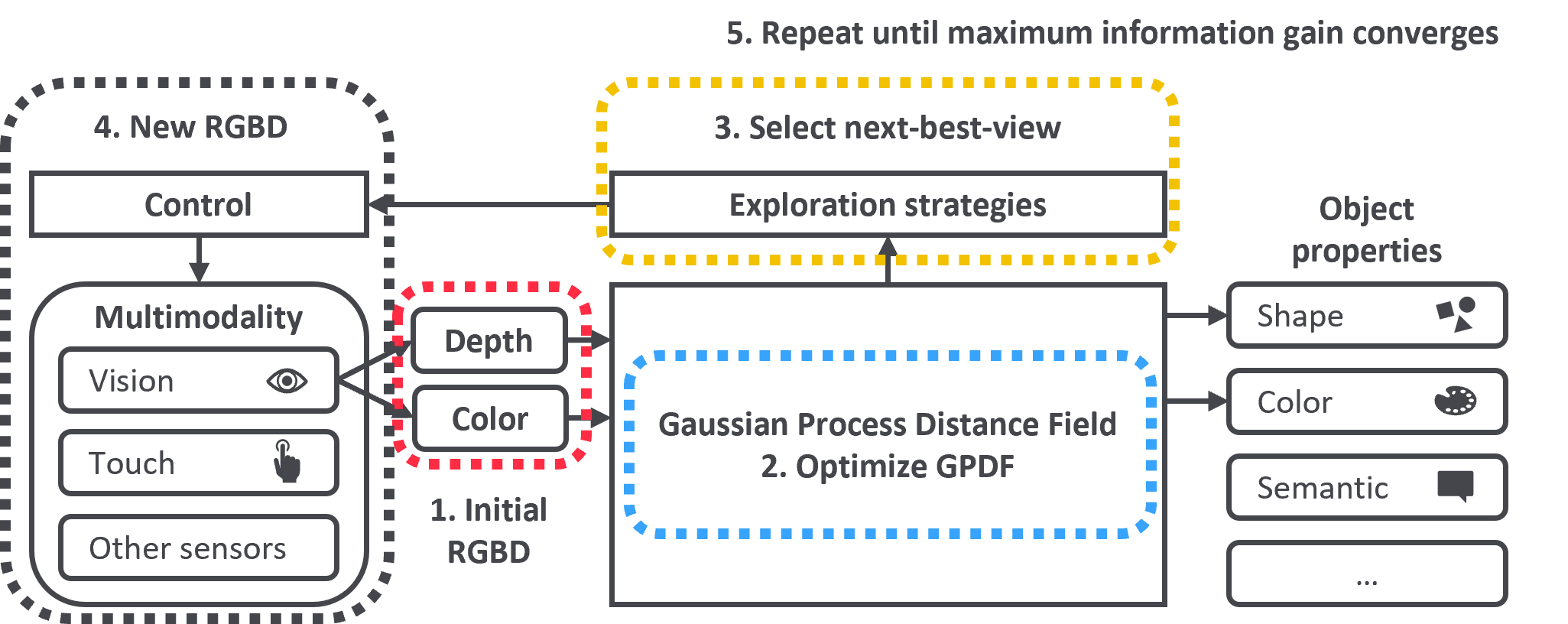}
    \caption{Visual exploration pipeline}
    \label{fig:visual_exploration_pipeline}
\end{figure}
\begin{figure}[tb]
    \begin{minipage}[c]{0.49\linewidth}
        \centering
        \includegraphics[width=0.8\linewidth]
        {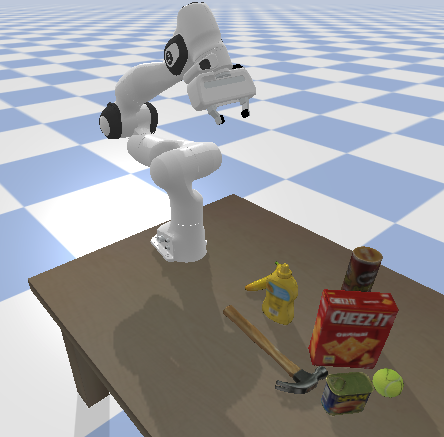}
        \caption{Simulation environment with Franka robot}
        \label{fig:simulation_env}
    \end{minipage}%
    \hfill
    \begin{minipage}[c]{0.49\linewidth}
        \centering
        \includegraphics[width=0.7\linewidth]{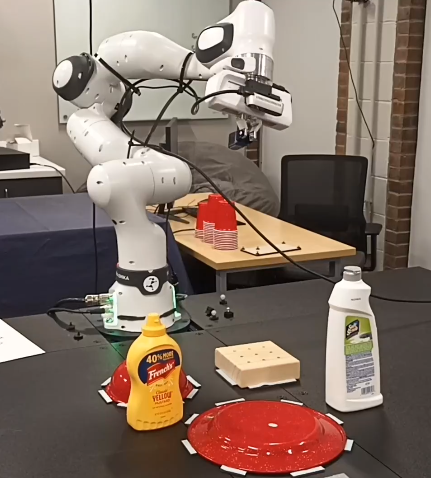}
        \caption{Real robot environment}
        \label{fig:real_robot_env}
    \end{minipage}
\end{figure}
We've implemented a simple setup for visual exploration in both simulation (Figure \ref{fig:simulation_env}) and on a real robot (Figure \ref{fig:real_robot_env}). The pipeline for visual exploration as in Figure \ref{fig:visual_exploration_pipeline} is outlined below:
\begin{enumerate}
    \item Capture an RGBD image from the initial pose and initialize GPDFs with the initial point cloud. Save the RGBD image and camera pose in the training dataset pool.
    \item Optimize the GPDFs using differentiable rendering (500 iterations).
    \item Choose the next-best-view from a set of predefined pose candidates based on the information gain and move to the sampled next-best-view and save the image for the next optimization step.
    \item Repeat steps 2-4 until a number of next-best-views is obtained or until the maximum information gain from candidate poses falls below a predefined threshold.
\end{enumerate}
The initialization of GPDFs using point cloud from initial RGBD image significantly accelerates the convergence of the GPDFs. In theory, we could continually increase the number of points in the point cloud as we add more camera views. However, for computational efficiency, we maintain a constant number of points over time. Additionally, we limit the workspace size and use small camera image sizes (32x24 pixels).

\begin{figure}[!tbp]
  \centering
  \includegraphics[width=0.6\linewidth]
    {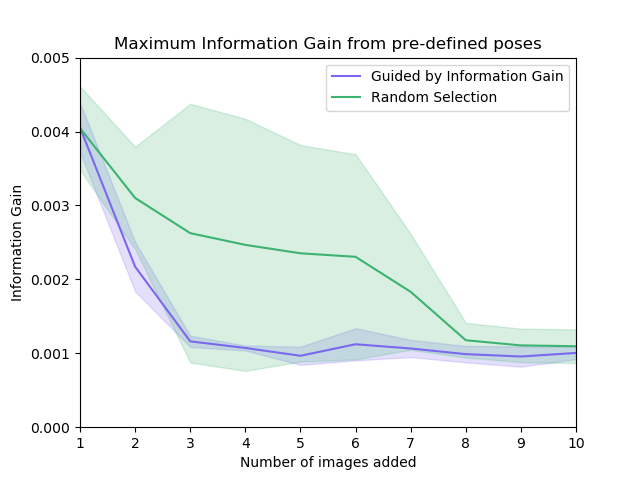}
    \caption{The maximum information gain from predefined camera poses compared to the number of sequentially collected images}
    \label{fig:ig_overtime}
\vspace{-10pt}
\end{figure}

\begin{figure}[!tbp]
  \centering
  \includegraphics[width=0.9\linewidth]{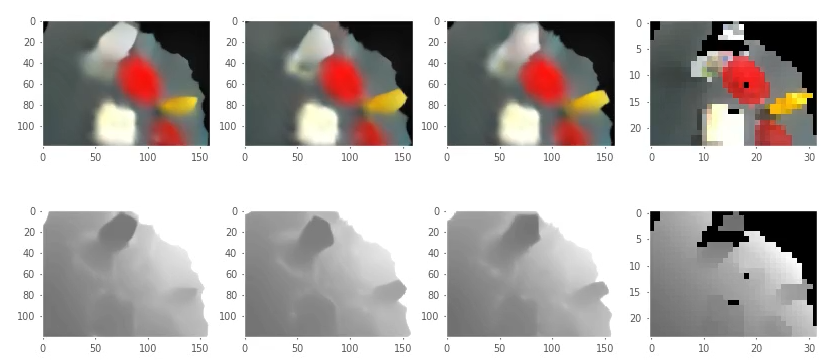}
  \caption{Rendered GPDF results with different point cloud initialization(column 1-3) and real RGBD image (column 4)
  }
  \label{fig:real_robot_visual}
\vspace{-10pt}
\end{figure}

In contrast to random view selection in simulation, employing information gain to guide the exploration accelerates the convergence of the GPDFs, as illustrated in Figure \ref{fig:ig_overtime}.
In the real robot environment, we do not observe an initially consistent decrease in information gain as seen in simulation. This inconsistency primarily stems from the noisy observations of the camera, which requires several views to start converging. Consequently, it typically requires 3-4 images initially to observe a meaningful decrease in information gain as more images are captured by the camera. The noise from the real camera can be observed in the rightmost part of Figure \ref{fig:real_robot_visual}, particularly around reflective, refractive, or sharp edge areas.

\section{Shape exploration from touch}
\label{sec:surface_exploration_touch}
Building on the previous section, we explore the use of tactile feedback to reconstruct geometry following visual exploration. We first examine how surface uncertainty from GPDF can guide tactile exploration. Next, we demonstrate this concept with a real robot in an actual implementation.

\subsection{Surface uncertainty for touch}

In contrast to other shape representation such as NeRF \cite{Mildenhall2020NeRFRS} or Gaussian Splatting \cite{Kerbl20233DGS}, GPDF inherently provides surface uncertainty suitable for tactile exploration, aiding in locating uncertain areas to touch. Previous studies have explored tactile sensing based on initial hand contact with an object \cite{KHADIVAR2023104461}. However, determining the initial points to touch through visual observation remains a less explored aspect. This task involves identifying a global maximum of uncertainty, which is inherently non-linear, rather than navigating locally and instantaneously across the surface. Furthermore, the identified points must lie on the object's surface, transforming the problem into a non-linear constrained optimization challenge. A straightforward approach to tackle this is to employ projected gradient descent, initiated from several random points in close proximity to the object. Using $\mathbf{x}$ as query input to GPDF, we get $\mathbf{var}(o(\mathbf{x}))$. And its gradient in respect to $\mathbf{x}$ is obtained using Cholesky factorization \cite{KHADIVAR2023104461} as below.
\begin{equation}
    \begin{split}
        & \frac{\partial\mathbf{var}(o(\mathbf{x}))}{\partial \mathbf{x}} = - 2 \frac{\partial k(\mathbf{x},\mathbf{X})}{\partial\mathbf{x}}(K(\mathbf{X},\mathbf{X})+ diag(\Delta_{\Bar{f}}\mathbf{\Sigma}_x\Delta_{\Bar{f}}^T))^{-1}\frac{\partial k(\mathbf{X},\mathbf{x})}{\partial\mathbf{x}}
    \end{split}
\end{equation}
\noindent
To have a gradient on the tangent plane of the surface, we project the gradient to the tangent plane using the surface normal.
\begin{equation}
    \begin{split}
        & \mathbf{n}_{\mathbf{x}}=\frac{\partial r(\hat{o}(\mathbf{x}))}{\partial \mathbf{x}} = \frac{d r(\hat{o}(\mathbf{x}))}{d \hat{o}(\mathbf{x})} \times \frac{\partial \hat{o}(\mathbf{x})}{\partial \mathbf{x}}\\
        & = \frac{d r(\hat{o}(\mathbf{x}))}{d \hat{o}(\mathbf{x})} \times \frac{\partial k(\mathbf{x},\mathbf{X})}{\partial\mathbf{x}}(K(\mathbf{X},\mathbf{X}) + diag(\Delta_{\Bar{f}}\mathbf{\Sigma}_x\Delta_{\Bar{f}}^T))^{-1}\mathbf{y} \\
        & \mathbf{s}_{\mathbf{x}} = \frac{\partial\mathbf{var}(o(\mathbf{x}))}{\partial \mathbf{x}} - \frac{\langle\frac{\partial\mathbf{var}(o(\mathbf{x}))}{\partial \mathbf{x}},\mathbf{n}_{\mathbf{x}}\rangle}{||\mathbf{n}_{\mathbf{x}}||_2}\mathbf{n}_{\mathbf{x}}
    \end{split}
\end{equation}
The projected gradient descent method and its accelerated version for $\mathbf{x}_k\in\partial\Omega$ is outlined below.
\begin{equation}
    \begin{split}
        & \text{Gradient descent: }\mathbf{x}_{k+1}=\text{Proj}_{\partial\Omega}(\mathbf{x}_{k}+\epsilon\nabla\mathbf{var}(o(\mathbf{x}_k))) \\
        & \text{Accelerated gradient descent: }
        \begin{cases}
            \mathbf{x}_{k+1}=\text{Proj}_{\partial\Omega}(\mathbf{y}_{k}+\epsilon\nabla\mathbf{var}(o(\mathbf{y}_k)))\\
            \mathbf{y}_{k+1}=\mathbf{x}_{k+1}+\frac{k-1}{k+2}(\mathbf{x}_{k+1}-\mathbf{x}_k)
        \end{cases}        
    \end{split}
\end{equation}
However, the projection process remains computationally costly, mainly due to the iterative nature of GPDF in calculating the precise distance for projecting points onto the surface. This entails multiplying the distance by the gradient of GPDF for the given query point. As a remedy, we suggest a modification to the projection step, relaxing the assumption that $\mathbf{x}_k$ must be on the object surface. Instead, we shift towards a dual-goal optimization approach, as detailed below.
\begin{equation}
    \mathbf{x}_{k+1}=\mathbf{x}_{k}-d(\mathbf{x}_k)\frac{\mathbf{n}_{\mathbf{x}_k}}{\|\mathbf{n}_{\mathbf{x}_k}\|_2}+\epsilon \left(\frac{2k+1}{k+2}\mathbf{s}_{\mathbf{x}_k}\right)\\
\end{equation}
\begin{figure}[!tbp]
  \centering
  \includegraphics[width=0.5\linewidth]{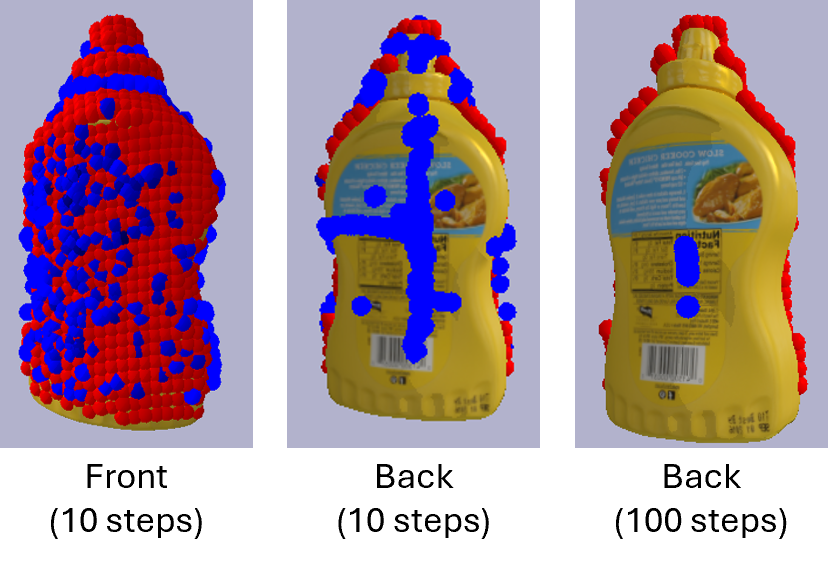}
  \caption{Gradient ascent towards higher uncertainty area (red: observed points, blue: gradient ascended points)}
  \label{fig:most_uncertain_point}
\vspace{-10pt}
\end{figure}
\begin{figure}[!tbp]
  \centering
  \includegraphics[width=0.6\linewidth]{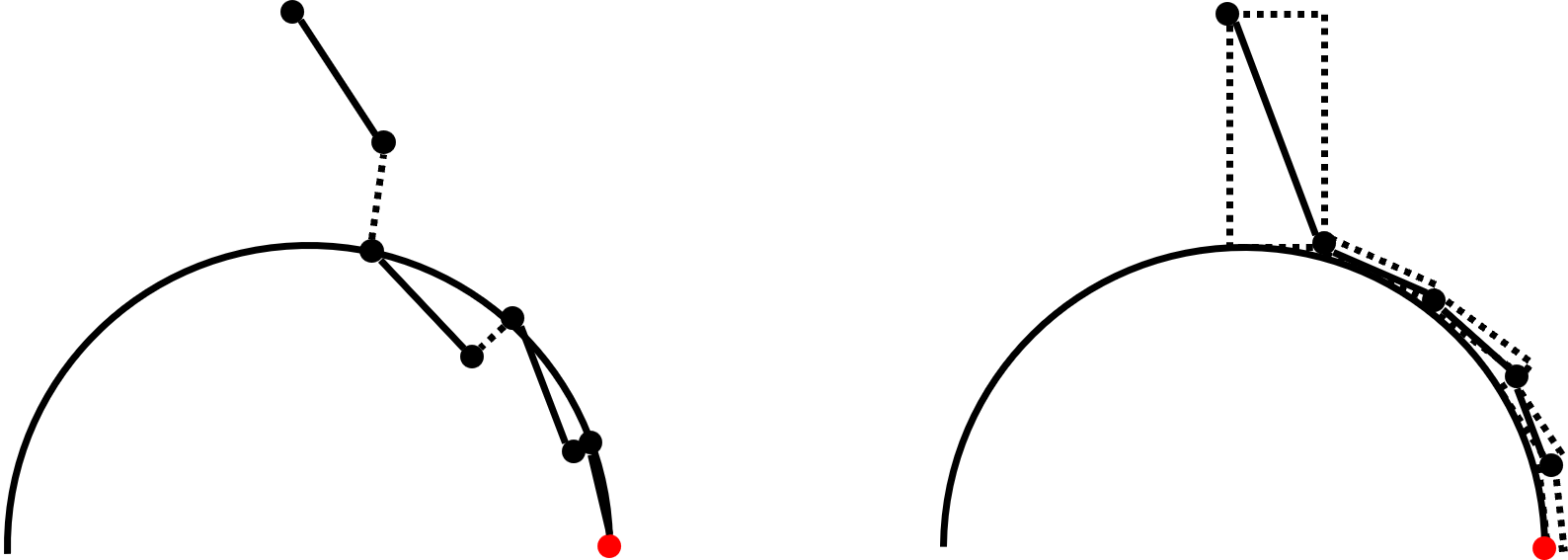}
  \caption{Optimization using accurate projection (left) and dual-goal optimization (right) to the red target point}
  \label{fig:gradient_descent}
\vspace{-10pt}
\end{figure}
In essence, for the tangential direction to the distance function, we adopt a gradient ascent approach to maximize uncertainty. Conversely, for the normal direction to the distance function, we implement a projection or distance-multiplied gradient descent strategy. The difference between the projection method and our approximation is shown in Figure \ref{fig:gradient_descent}. 


This ensures that the point converges closer to the surface where the uncertainty is well-defined as in Figure \ref{fig:most_uncertain_point}. The point with the highest uncertainty is then chosen as the estimated global minimum, where the robot should make initial contact. To control the robot's movement, we use the uncertain point $\mathbf{x}$ and its surface normal $\mathbf{n}_{\mathbf{x}}$ to determine the target pose for the tactile sensor, $x_{\rm ref}$. The reference velocity is calculated based on the current tactile sensor pose, $x$, as $\dot{x}_{\rm ref} = k(x_{\rm ref} - x)$, where $k$ is a positive constant. This is then feeded to the QP-IK controller.

\subsection{Implementation}
\begin{figure}[tb]
    \centering
    \includegraphics[width=1.0\textwidth]{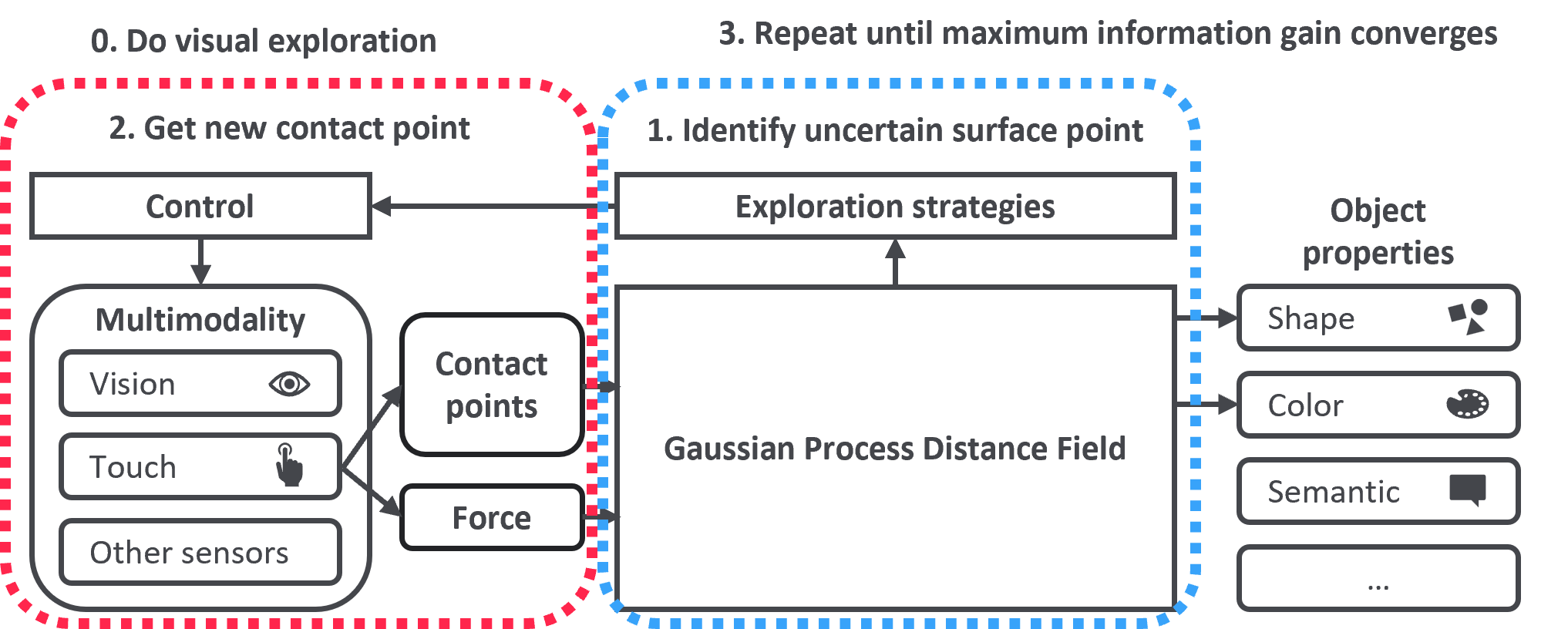}
    \caption{Tactile exploration pipeline}
    \label{fig:tactile_exploration_pipeline}
\end{figure}
We combine exploration from vision and exploration from touch using surface unceratinty as below pipeline.
\begin{enumerate}
    \setcounter{enumi}{-1}
    \item Do visual exploration and have rough estimation of scene geometry
    \item  Identify the most uncertain surface point based on the GP model.
    \item Move the robot to the identified point and add it as an observation point.
    \item Repeat steps 5-6 until the uncertainty of surface points falls below a predefined threshold.
\end{enumerate}

\begin{figure}[!tbp]
  \centering
  \includegraphics[width=0.9\linewidth]{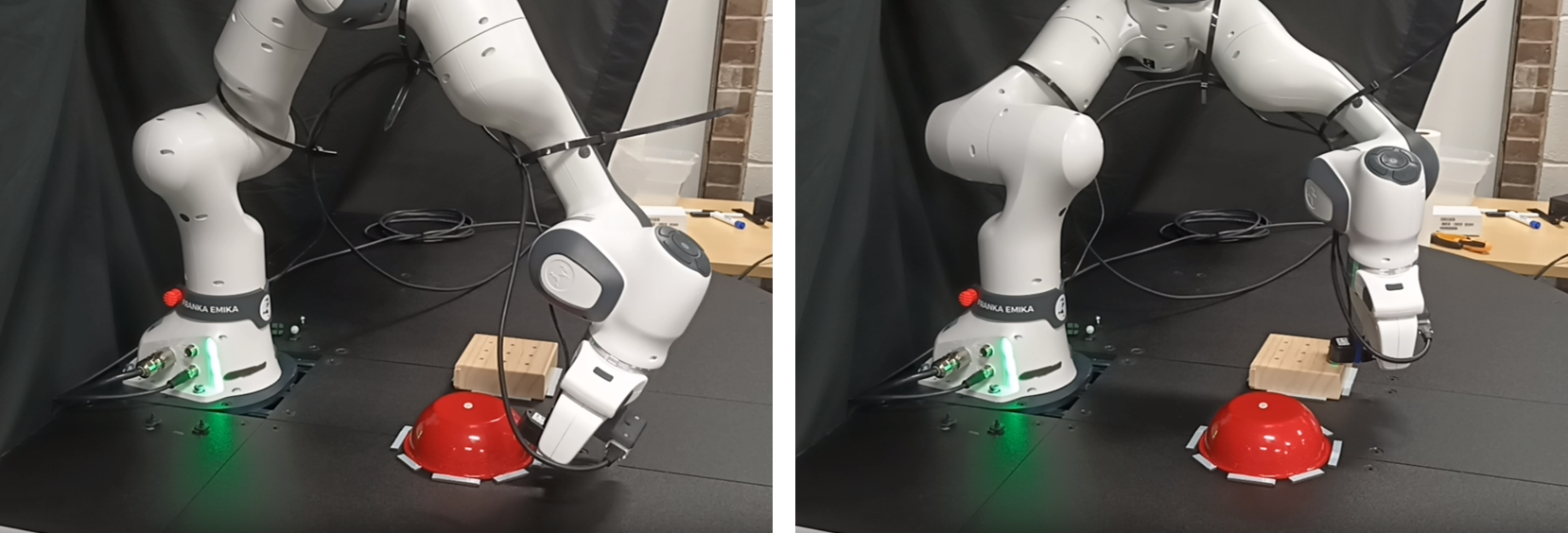}
  \caption{Tactile exploration of uncertain surface areas}
  \label{fig:real_robot_tactile}
\vspace{-10pt}
\end{figure}
\begin{figure}[tb]
    \begin{minipage}[c]{0.49\linewidth}
        \centering
        \includegraphics[width=1.0\linewidth]{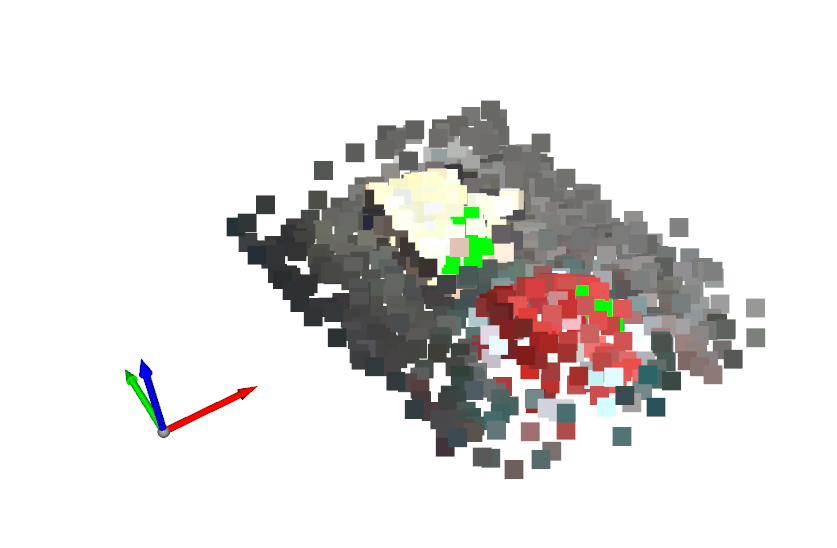}
        \caption{Point cloud of environment and explored uncertain surface point colored in green}
        \label{fig:point_cloud_tactile_exploration}
    \end{minipage}%
    \hfill
    \begin{minipage}[c]{0.49\linewidth}
        \centering
        \includegraphics[width=1.0\linewidth]{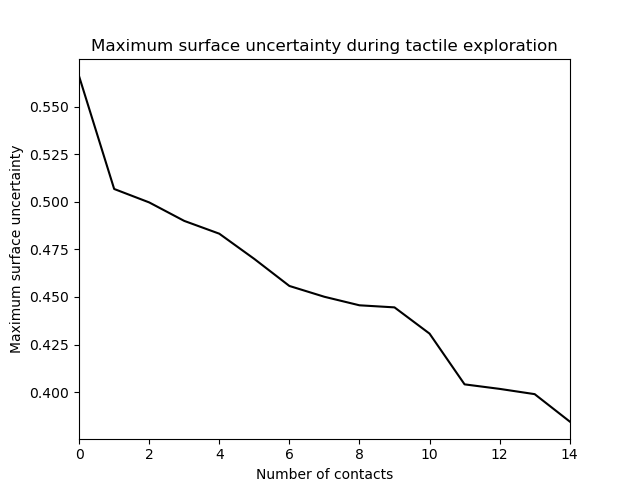}
        \caption{Maximum surface uncertainty decreasing as tactile exploration progresses}
        \label{fig:surface_uncertainty_maximum}
    \end{minipage}
\end{figure}

For tactile exploration, we conducted tests with a simpler setup as in Figure \ref{fig:real_robot_tactile}, initially explored through vision as outlined in the aforementioned pipeline. We found that the most uncertain surface areas, often coincided with regions where the depth camera had the most noise, such as reflective portions of a red ceramic bowl or the sharp edges of a rectangular wooden object, as depicted in Figure \ref{fig:point_cloud_tactile_exploration}.
From Figure \ref{fig:surface_uncertainty_maximum}, it is evident that as more points are explored using tactile sensing, the maximum surface uncertainty decreases. Additionally, the parts that were ambiguous from the vision due to noise, which cannot be resolved by simply acquiring more camera views, are successfully explored using touch.



\section{Integrating Vision and Touch}
\label{sec:integrating_vision_touch}
In earlier sections, we examined visual and tactile exploration individually to understand their roles in reconstructing an object's geometry. While each modality can be effective on its own, integrating them can yield even better results. In this section, we explore how combining vision and touch provides a deeper and more comprehensive understanding of an environment. The key idea is that each sense has unique strengths, and their integration creates a more robust exploration strategy. This combined approach offers benefits in three main areas: accuracy, efficiency, and flexibility.
\begin{enumerate}
    \item \textbf{Accuracy:} Combining visual and tactile data improves accuracy, with each sense compensating for the other's weaknesses. Vision can quickly scan a scene, identifying key features like edges, corners, and textures. Tactile exploration is useful for areas with high uncertainty, fine details, or where visual data is unreliable due to reflections or occlusions as we saw before in Section \ref{sec:surface_exploration_touch}.
    \item \textbf{Efficiency:} An integrated approach can also reduce exploration time. Vision updates the scene's geometric model using computationally intensive methods like differentiable rendering. While this is happening, tactile exploration can gather additional data. Alternatively, if tactile sensors are limited to local surfaces, visual exploration can be used to cover larger areas more quickly. This parallel approach focuses tactile efforts on critical spots identified by vision, 
    \item \textbf{Flexibility:} Combining vision and touch increases adaptability to different environments and tasks, making the system more robust and versatile. Exploration strategies can be tailored by weighing the Information Gain obtained from both senses. Alternatively, the robot's kinematics can guide the approach, with vision preferred for hard-to-reach areas and tactile exploration used for locations hidden from view.
\end{enumerate}
By integrating vision and touch, we can create a more comprehensive and effective exploration system, with each modality reinforcing the other to achieve better outcomes.

\chapter{Object Exploration from Touch} \label{ch:property_exploration_touch}
\begin{figure}[!h]
  \centering
  \includegraphics[width=0.9\linewidth]{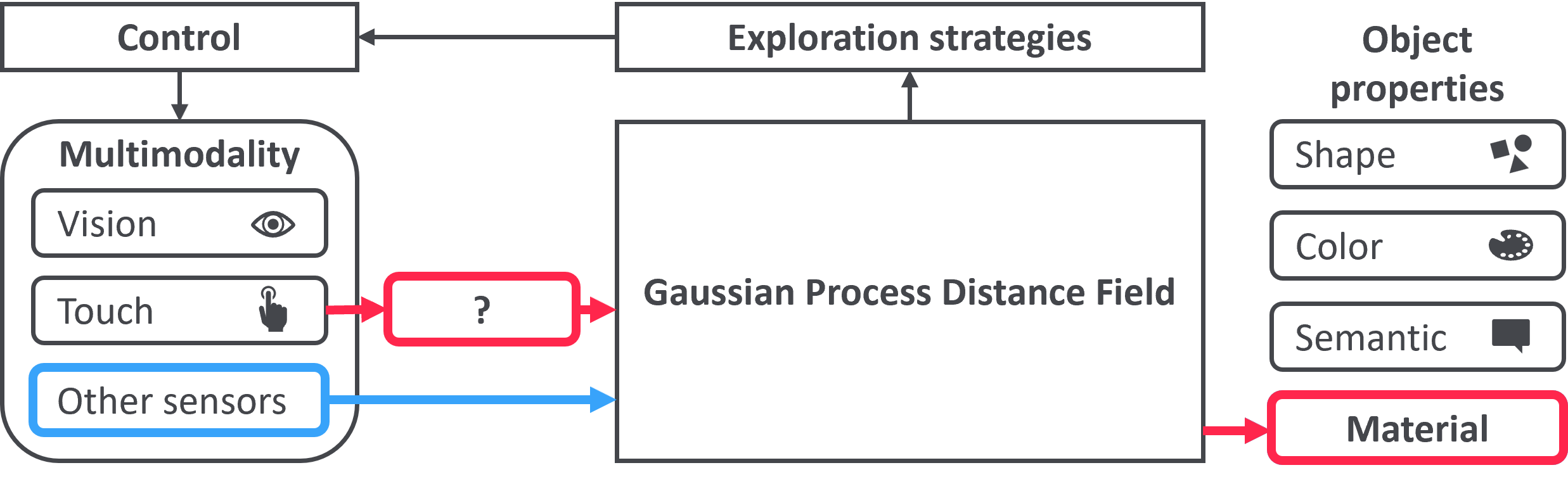}
  \caption{Exploration pipeline for material properties}
  \label{fig:object_exploration_pipeline}
\vspace{-10pt}
\end{figure}
In the previous chapter, we discussed integrating different surface modalities into the GPDF framework. Now, we explore how tactile sensors can be used to investigate object properties beyond surface geometry. As depicted in Figure \ref{fig:object_exploration_pipeline}, incorporating additional sensors can reveal more about an object's material characteristics. For example, a temperature sensor could provide a complete temperature profile across an object's surface. However, the focus of this chapter is on exploring material properties with the tactile sensors that we already have. Assuming we have full information of object or scene geometry from previous chapters, we investigate what other information these tactile sensors can provide. In Section \ref{sec:tactile_exploration_strategies}, we discuss strategies for tactile exploration. In Section \ref{sec:material_classificaiton}, we examine how tactile data can be used to classify object materials and describe how this information can be integrated into the GPDF framework.

\section{Tactile exploration strategies}
\label{sec:tactile_exploration_strategies}
\begin{figure}
    \centering
    \includegraphics[width=0.5\linewidth]{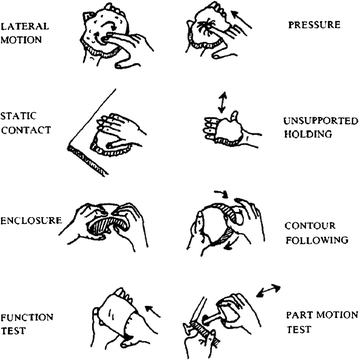}
    \caption{Hand motions from \cite{Lederman1987HandMA}. Through tactile exploration, humans can gather meaningful properties of objects.}
    \label{fig:hand_movement}
\end{figure}
Lederman and Klatzky \cite{Lederman1987HandMA} studied how humans use tactile exploration to gather information about objects, identifying three main categories of acquired knowledge: substance-related properties, focusing on the material; structure-related properties, focusing on geometry; and functional properties, focusing on articulation and affordance. They observed that specific tactile strategies are used to discern these properties. Of particular interest to us are the substance-related strategies, such as lateral motion for texture, pressure for hardness, and static contact for temperature.

\begin{figure}[tb]
    \begin{minipage}[c]{0.49\linewidth}
        \centering
        \includegraphics[width=0.8\linewidth]
        {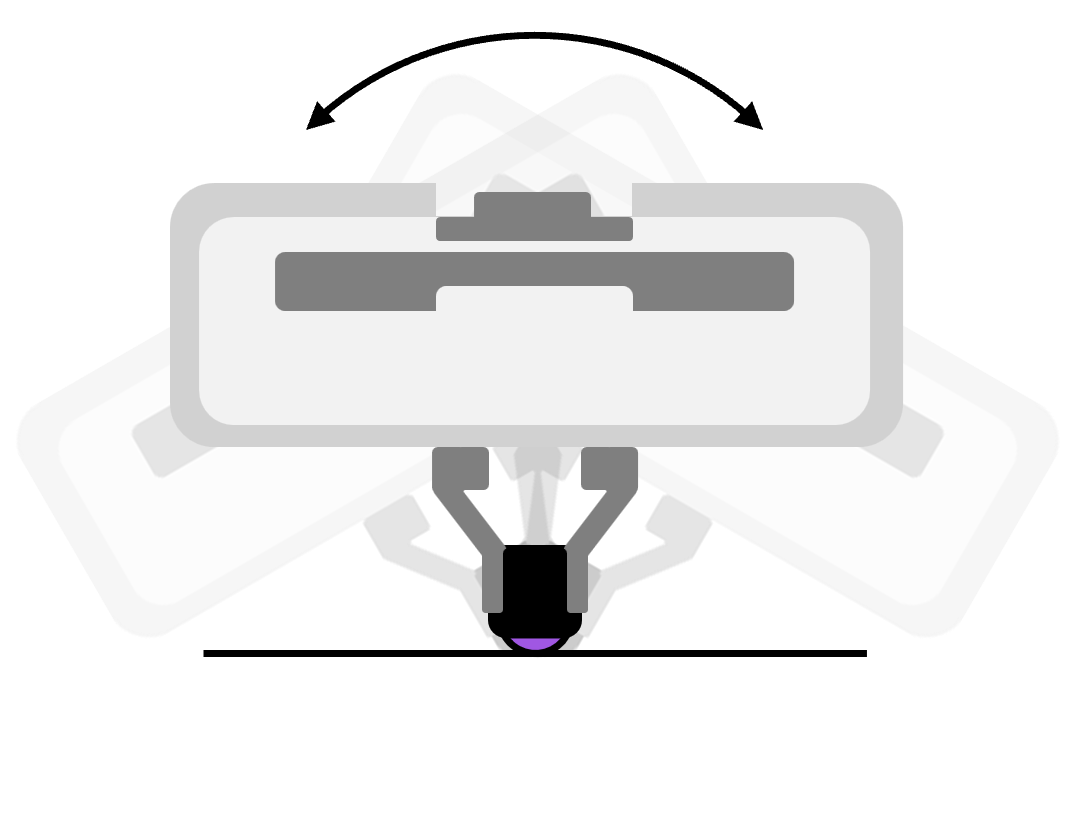}
        \caption{Object property exploration strategy using rolling}
        \label{fig:property_exploration}
    \end{minipage}%
    \hfill
    \begin{minipage}[c]{0.49\linewidth}
        \centering
        \includegraphics[width=0.4\linewidth]{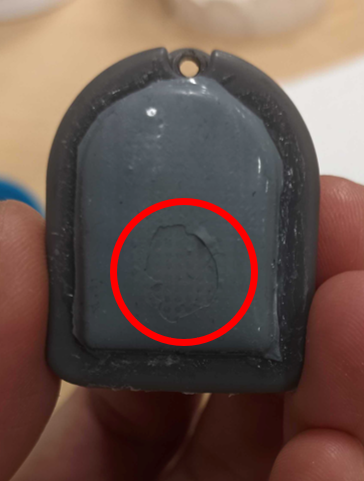}
        \caption{Sensor wearing out during sliding exploration}
        \label{fig:broken_sensor}
    \end{minipage}
\end{figure}

Sliding on the surface \cite{Ding2017TactilePO, Richardson2022LearningTF, Khojasteh2024RobustSR, Aoyama2023FewShotLO} is a widely researched lateral motion strategy for identifying materials. It can detect properties such as friction coefficients, roughness, and hardness, providing valuable insights into the nature of a material. However, sliding has its drawbacks, especially with current tactile sensors. For instance, the silicone gel used in our sensor tends to degrade over time, potentially destroying the reflective coating necessary for depth estimation, as shown in Figure \ref{fig:broken_sensor}. Many other sensors use soft materials that also wear down with use. Moreover, the sliding strategy relies on the sensor's capacity to detect high-frequency vibrations, a challenge when using cameras with frame rates typically limited to about 60 Hz.
Considering these limitations, we suggest an alternative exploration strategy that involves a rolling motion. In this approach, the robot makes contact with the object and then rolls its fingers from left to right to perceive the material, as shown in Figure \ref{fig:property_exploration}. This strategy is similar to applying pressure to assess hardness \cite{Aoyama2023FewShotLO} but offers a time-varying measurement, which can aid in further material identification. Although this method may be less effective than sliding in certain cases, it reduces the risk of sensor damage and eliminates the need to capture high-frequency vibrations by instead measuring force changes in xyz directions. This technique offers a balance between preserving sensor durability and accurately exploring material properties.

\section{Object material classification}
\label{sec:material_classificaiton}

\begin{figure}[!tbp]
  \centering
  \includegraphics[width=0.9\linewidth]{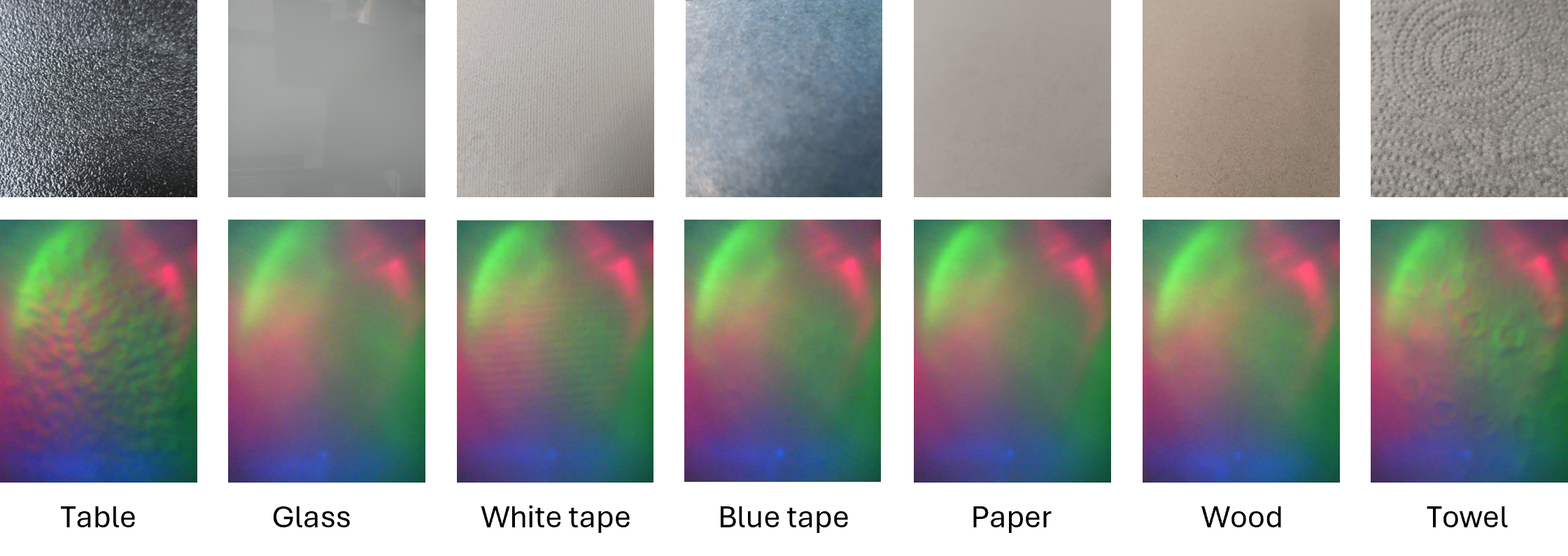}
  \caption{Different materials with its tactile sensor reading}
  \label{fig:different_matrials}
\vspace{-10pt}
\end{figure}
With the object's geometry already given, we can use tactile sensors to explore surface properties for identifying different materials. To demonstrate this, we collected tactile sensor data from various flat objects with distinct materials, including a plastic table, glass whiteboard, white fabric tape, blue-coated tape, paper, wood panel, and kitchen towel, as shown in Figure \ref{fig:different_matrials}. By examining the raw tactile sensor data, you can observe that certain materials, like the plastic table, white tape, and kitchen towel, have easily identifiable textures.

\begin{figure}[tb]
    \begin{minipage}[c]{0.49\linewidth}
        \centering
        \includegraphics[width=0.9\linewidth]
        {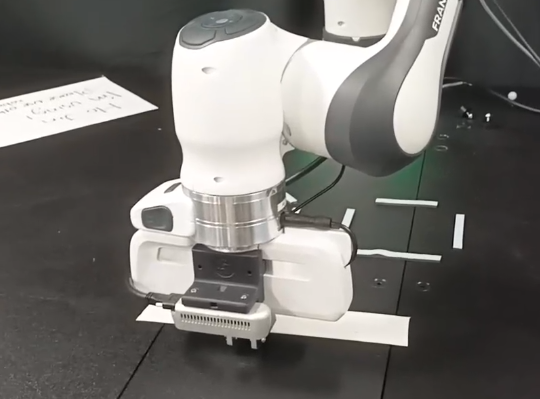}
        \caption{Real robot doing a rolling strategy}
        \label{fig:robot_rolling}
    \end{minipage}%
    \hfill
    \begin{minipage}[c]{0.49\linewidth}
        \centering
        \includegraphics[width=1.0\linewidth]{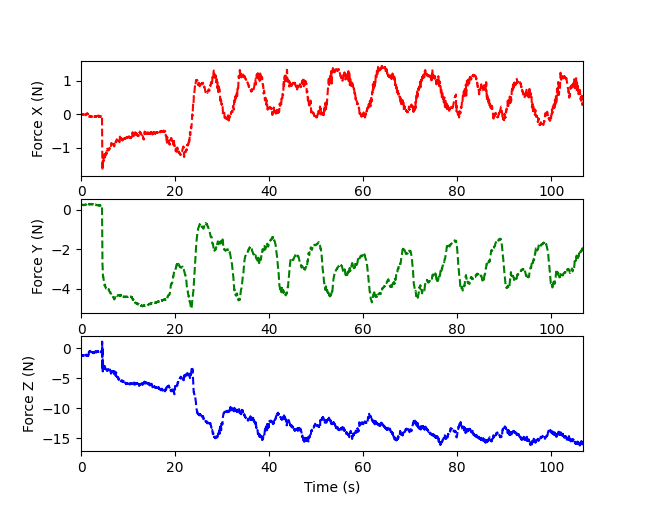}
        \caption{Force reading from the tactile sensor during rolling}
        \label{fig:forces_rolling}
    \end{minipage}
\end{figure}
To collect data for material classification, we used a rolling strategy with the robot, touching the object and then rolling its fingers left and right. As illustrated in Figure \ref{fig:robot_rolling}, we conducted three trials for each material, with each trial comprising 7-8 cycles of rolling. During the process, we used the tactile perception pipeline from \ref{sec:image_processing_pipeline} to measure force in the xyz directions to assess pressure variations. As shown in Figure \ref{fig:forces_rolling}, these force readings can provide critical information about the material's unique tactile properties.

\subsection{Classification}
We analyzed the force amplitude in each of the three directions-x, y, and z-for every rolling trial to evaluate the characteristics of different materials.
\begin{figure}[!tbp]
  \centering
  \includegraphics[width=0.6\linewidth]{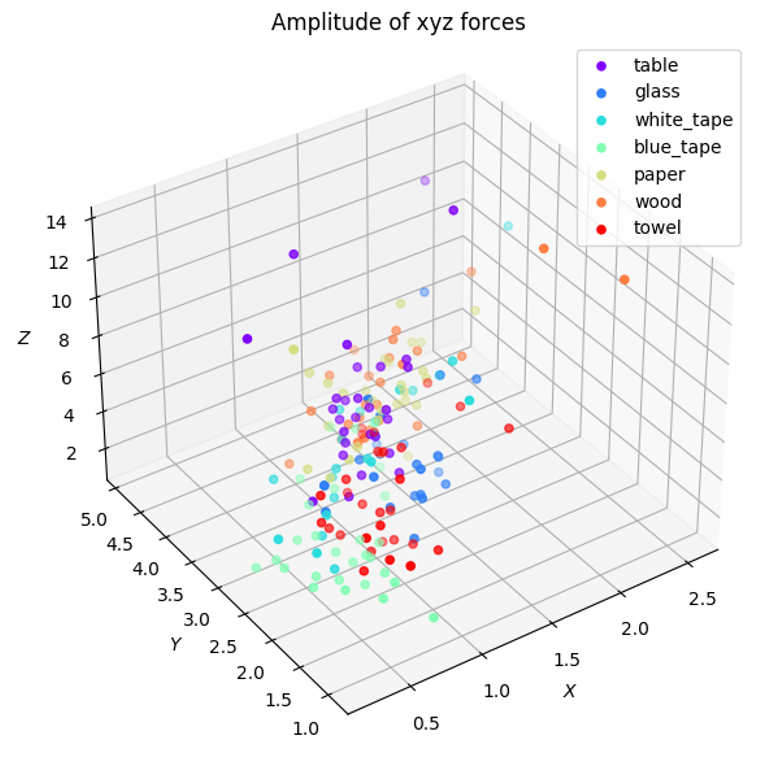}
  \caption{Rolling trials and its force amplitudes}
  \label{fig:force_comparison}
\vspace{-10pt}
\end{figure}
\begin{table}[]
\centering
\begin{tabular}{|c|c|c|}
\hline
                 & Training accuracy & Cross validation accuracy \\ \hline
SVM (RBF kernel) & 36.05\%           & 36.37\%                   \\ \hline
MLP (64x64x64)      & 82.99\%           & 33.11\%                   \\ \hline
Random Forest    & 79.59\%           & 29.87\%                   \\ \hline
\end{tabular}
\caption{Training and cross validation accuracy of different classifiers using force amplitude features}
\label{tab:material_accuracy}
\end{table}
Figure \ref{fig:force_comparison} shows that certain materials, such as blue-coated tape, exhibit a notably lower force amplitude in the z-direction (normal to the surface), suggesting that some materials produce distinct tactile patterns. However, other materials do not present clear distinctions based solely on force amplitude. We tested this force amplitude data, consisting of 187 data points, as input to different classifiers to identify patterns among the seven materials. The classifiers we tested include:
\begin{itemize}
    \item A Support Vector Machine (SVM) with a Radial Basis Function (RBF) kernel, tuned with hyperparameters $C=10^3$ and $\gamma=10^{-4}$ from a grid search on a logarithmic scale.
    \item A Multi-Layer Perceptron (MLP) with hidden layers of size 64x64x64, where grid search was used to determine optimal layer depth and width.
    \item A Random Forest model with simple feature extraction.
\end{itemize}
Table \ref{tab:material_accuracy} presents the training and cross-validation accuracies for each classifier. The cross-validation accuracy, achieved by dividing the dataset into three subsets and using each in turn as a test set, exceeded random chance (14.29\% for seven materials), indicating some ability to distinguish materials based on force amplitude. However, the cross-validation accuracy was significantly lower than the training accuracy, suggesting potential overfitting.

The limited accuracy of the classifiers indicates that force amplitude alone might not be sufficient to reliably distinguish between all seven materials. Our hypothesis is that the low-dimensional feature set is too simplistic to capture the unique tactile characteristics of each material. To improve classification accuracy, we could incorporate higher-dimensional features like:
\begin{itemize}
    \item Raw image data: This captures detailed texture information that could be useful for material classification.
    \item Deformation vectors: Data from the Coherent Point Drift (CPD) algorithm, which tracks deformations during tactile exploration, may offer additional insights into material properties.
\end{itemize}
Using these higher-dimensional features might improve classifier performance, as they provide a richer dataset to work with. Additionally, since our current rolling motion is relatively slow, adding features that reflect higher-frequency vibrations might yield more precise classification results. These improvements could also facilitate faster online processing, allowing the robot to adapt to new materials in real-time.

\subsection{Incorporating classification into multimodal sensing}
Gaussian Processes offer flexibility in incorporating classification in two distinct ways: 
\begin{enumerate}
    \item Using GPs to learn feature vectors and then passing those to a separate classifier.
    \item Using GPs as classifiers themselves.
\end{enumerate}
The first method involves using GPs to generate feature vectors, which are then fed into an external classifier. This approach is straightforward, leveraging the multimodality of GPs. For example, if we have force data at a given Euclidean point $\mathbf{x}$ in three dimensions-$y_{f_x}$, $y_{f_y}$, and $y_{f_z}$-we can fit this data using the following GP equations:
\begin{equation}
    \begin{split}
        & \hat{y}_{f_x}(\mathbf{x_*})=k(\mathbf{x_*},\mathbf{X})(K(\mathbf{X},\mathbf{X})+\sigma_y^2 I)^{-1}\mathbf{y}_{f_x}\\
        & \hat{y}_{f_y}(\mathbf{x_*})=k(\mathbf{x_*},\mathbf{X})(K(\mathbf{X},\mathbf{X})+\sigma_y^2 I)^{-1}\mathbf{y}_{f_y}\\
        & \hat{y}_{f_z}(\mathbf{x_*})=k(\mathbf{x_*},\mathbf{X})(K(\mathbf{X},\mathbf{X})+\sigma_y^2 I)^{-1}\mathbf{y}_{f_z}\\
    \end{split}
\end{equation}
Given a query point \(\mathbf{x_*}\), these outputs can be fed into a classifier to determine the material property. This does not require training the already existing classifier again.
The second method uses the GP directly for classification. For example, this method assigns a label $y_M$ to a given data point $\mathbf{x}$, where -1 represents ``not this material" and 1 represents ``this material." The GP's output is then passed through a sigmoid function to determine the likelihood that the query point corresponds to this material:
\begin{equation}
    \begin{split}
        & \hat{y}_{M}(\mathbf{x_*})=k(\mathbf{x_*},\mathbf{X})(K(\mathbf{X},\mathbf{X})+\sigma_y^2 I)^{-1}\mathbf{y}_{M}\\
        & \text{Binary classification: } \text{Sigmoid}(\hat{y}_{M})
    \end{split}
\end{equation}
If we are classifying multiple materials, we can use a softmax function for multiclass classification. This approach has the advantage of not requiring feature normalization since it relies on the GP's classification output rather than feature-based input, given a base classifier.

Both methods provide pathways to incorporate classification into multimodal sensing using GPs. The first method leverages GP-based feature extraction with a separate classifier, while the second method uses the GP for direct classification. The choice depends on the specific application requirements and feature structure.

One limitation of Gaussian Processes (GPs) in this context is that they don't inherently consider constraints imposed by a manifold. Given that our setup assumes the geometry or manifold is known, we can use Riemannian manifold Gaussian Processes \cite{Borovitskiy2020MaternGP} to limit interpolation to the surface, ensuring the GP's estimations remain consistent with the known geometry. This approach is particularly intriguing for further exploration, as it requires fewer observation points compared to conventional GPs. Another limitation of standard GPs is that they do not inherently account for relationships between different object properties. Some properties, like hardness and roughness, may have linear or nonlinear correlations, meaning that measuring one property could reduce uncertainty about the other. Exploring multi-input, multi-output Gaussian Processes \cite{Seeger2004GaussianPF} could be a valuable direction for future research, potentially enabling more efficient exploration by leveraging these property relationships.

\chapter{Conclusion}\label{sec:conclusion}
\section{Summary}
In conclusion, this work proposes an active perception framework that integrates vision and tactile sensing for accurate reconstruction and understanding of object properties. At the core of the framework is the Gaussian Process Distance Field (GPDF), which offers several desirable characteristics such as smooth analytic gradients, direct surface uncertainty measurements, and the ability to incrementally fuse multi-modal observations. To address the scalability limitations of Gaussian processes, the work explores approaches such as inducing points to reduce computation complexity.

The proposed pipeline starts with an initial shape estimate from vision using differentiable rendering of the point cloud, followed by iterative refinement through tactile exploration. By quantifying information gain on generative images and measuring surface uncertainty in the GPDF, the system can plan exploratory motions to maximize information gain or find regions of high surface uncertainty, such as potential contact points. This process enables the system to gradually build a high-fidelity model of the object's geometry. Overall, this active multi-sensor perception system holds promise for enabling robust object understanding and manipulation capabilities in unstructured environments.

\section{Limitations}
The limitations of this work revolve around scalability issues from two key aspects:
\begin{itemize}
    \item \textbf{Gaussian Process Complexity:} Gaussian Processes inherently suffer from cubic computational complexity $O(n^3)$ in the number of data points. While several effective approximation methods like inducing point were explored to limit the number of points, this still requires careful tuning to balance accuracy and computational cost trade-offs. One approach to address this challenge is to find a kernel function capable of handling complex and higher dimensional shape primitives beyond just points. This would enhance the expressivity of each primitive while requiring a relatively small number of data points. 
    \item \textbf{Differentiable Rendering:} Although differentiable rendering enables modeling complicated camera views, it poses a scalability bottleneck since the scene geometry has to be optimized using multiple generative models. This optimization process is computationally expensive and may require the robot to pause during geometry updates, deviating from human-like real-time perception. Other methods like NeRF and Gaussian splatting attempt to overcome this using large GPU memories and optimized architectures or heuristics, which could be interesting avenues to explore.
\end{itemize}

The second major limitation lies in the exploration strategies employed. Currently, explorations are conducted through discrete sampling, which diverges from the smooth, continuous, and real-time exploratory behavior observed in humans. Furthermore, humans utilize a diverse range of exploratory motions and techniques, seamlessly transitioning between them, a capability that the current system may not fully capture. One approach to address this issue is by planning out a sequence of motions that involve multiple camera views, contact poses, or even combination of those, instead of relying on a single camera view pose or single contact point. This can help generate smoother motion trajectories and improve the exploration process.

Some other potential limitations include: handling sensor noise and robust modeling of physical interactions like friction and deformations during tactile exploration. Overall, while the proposed active multi-sensor perception framework is well-motivated, its practical deployment and generalization will depend on effectively overcoming these scalability, exploration strategy, and other potential limitations.

\section{Future work}
While numerous advancements have been made in active perception technologies, many intriguing questions remain for future research:
\begin{itemize}
    \item \textbf{Exploration Policies:} How can we develop more human-like, continuous exploration policies that can intelligently combine different types of exploratory motions or smoothly transition between them, such as grasping an object, rotating it, and re-grasping? Can we learn these policies from human demonstration data?
    \item \textbf{Online Active Perception:} Can we enable the system to continuously learn and update its object representations in an online manner as new multi-modal data streams in, without requiring periodic retraining from scratch or pausing the robot's movements?
    \item \textbf{Material Property Estimation:} 
    What methods or tactile sensing abilities help distinguish between various materials and their properties? Can making a few guesses about a material's properties help predict other properties that haven't been explored yet?
    \item \textbf{Multi-modal Tactile Sensing:} Humans possess multiple tactile sensory cells that detect different properties like temperature and pain. Can we develop tactile sensors that incorporate these diverse modalities? How can we relax the point observation assumption and deal with non-rigid objects? Can the robot learn correlations between object properties and infer some properties from others to make exploration more efficient?
    \item \textbf{Task-conditioned Exploration:} Can we enable the robot to explore the environment only until it has gathered sufficient information to carry out a specific task (exploitation), without exploring beyond what is necessary? How can we quantify the sufficiency of exploration for a given task?
    \item \textbf{Safety Dependent Capability:} Can we incrementally adjust the safety constraints of a robot based on its level of certainty about the environment, allowing it to carry out tasks more efficiently or exert additional capabilities while still avoiding risky situations?
\end{itemize}
In robotics, accurate environment modeling based on visual appearance or physics is achieved through observations like differentiable rendering or system identification. Active perception will continue bridging such models with robot control and planning, enabling better environmental adaptation and task execution.

\appendix
\chapter{Covariance Kernels}\label{covariance_kernels}

Within the domain of Gaussian processes, a kernel, also referred to as a covariance function, plays a fundamental role by holding a pivotal position in establishing covariance connections among random variables within the Gaussian process framework. When integrated with the mean function, the kernel becomes a comprehensive specification for the Gaussian process, essentially serving as a means of capturing the interplay between random variables and shaping the overall behavior of the process. The reverting function for Generalized Probability Density Function (GPDF) and the spectral density for Hilbert-space approximation corresponding to each kernel function are detailed in Table \ref{tab:covariance_kernel}. From the table, we can see that only the Matern $\nu=1/2$ kernel function has a reverting function that can output negative distances.

\begin{table}[!h]
\centering
\resizebox{\textwidth}{!}{%
\begin{tabular}{|c||c|c|c|}
\hline
                                                             & \textbf{Covariance kernel} ($k(d)$) & \textbf{Reverting function} ($r(o)$) & \textbf{Spectral density} ($S(s)$) \\ \hline\hline
\begin{tabular}[c]{@{}c@{}}Rational\\ Quadratic\end{tabular} & $\left(1+\frac{d^2}{2 \alpha l^2}\right)^{-\alpha}$ & $\sqrt{2\alpha l^2 \left(o^{-\frac{1}{\alpha}}-1\right)}$ & $\frac{l s^{\alpha-D/2}\mathcal{K}_{\alpha-D/2}(s)}{2^{\alpha-1}\Gamma(\alpha)}$ \\ \hline
\begin{tabular}[c]{@{}c@{}}Square\\ Exponetial\end{tabular}  & $\exp\left(-\frac{d^2}{2l^2}\right)$ & $\sqrt{-2l^2 \log(o)}$ & $(2\pi l^2)^{D/2}\exp(-2\pi^2l^2s^2)$ \\ \hline
\begin{tabular}[c]{@{}c@{}}Matern\\ $\nu=1/2$\end{tabular}                                             & $\exp\left(-\frac{d}{l}\right)$ & $-l \log(o)$ & \multirow{2}{*}{$\frac{2^D\pi^{D/2}\Gamma(\nu+D/2)(2\nu)^{\nu}}{\Gamma(\nu)l^{2\nu}}\left(\frac{2\nu}{l^2}+4\pi^2s^2\right)^{-(\nu+D/2)}$}      \\ \cline{1-3}
\begin{tabular}[c]{@{}c@{}}Matern\\ $\nu=3/2$\end{tabular} & $\left(1+\frac{\sqrt{3}d}{l}\right)\exp\left(-\frac{\sqrt{3}d}{l}\right)$ & $\underset{d}{\mathrm{argmin}}\|o-k(d)\|^2$ &                           \\ \hline
\end{tabular}%
}
\caption{Covariance kernels, corresponding reverting function, and spectral density \cite{Gentil2023AccurateGP, Ounpraseuth2008GaussianPF}. $D$ is the dimension of input, which is three for point clouds.}
\label{tab:covariance_kernel}
\end{table}

\chapter{Surface normal and curvature}\label{normal_curvature}
The gradient of GPDF can be obtained as below:
\begin{equation*}
    \begin{split}
        & \nabla d_{\mathbf{x}_*} = \frac{d r(\hat{o}(\mathbf{x}_*))}{d \hat{o}(\mathbf{x}_*)} \cdot \frac{\partial k(\mathbf{x}_*,\mathbf{X})}{\partial\mathbf{x}_*}(K(\mathbf{X},\mathbf{X}) + diag(\Delta_{\Bar{f}}\mathbf{\Sigma}_x\Delta_{\Bar{f}}^T))^{-1}\mathbf{y}\\
        & \frac{d r(\hat{o}(\mathbf{x}_*))}{d \hat{o}(\mathbf{x}_*)} = -\frac{l}{\hat{o}}\\
        & \frac{\partial \hat{o}(\mathbf{x}_*)}{\partial \mathbf{x}_*} = \frac{\partial k(\mathbf{x}_*,\mathbf{X})}{\partial\mathbf{x}_*}(K(\mathbf{X},\mathbf{X}) + diag(\Delta_{\Bar{f}}\mathbf{\Sigma}_x\Delta_{\Bar{f}}^T))^{-1}\mathbf{y}\\
        & \frac{\partial k(\mathbf{x}_*,\mathbf{X})}{\partial\mathbf{x}_*} = -\frac{1}{l}
        \begin{bmatrix}
            \exp\left(-\frac{d_1}{l}\right)\frac{(\mathbf{x}_*-\mathbf{x}_1)^T}{d_1}& \cdots & \exp\left(-\frac{d_n}{l}\right)\frac{(\mathbf{x}_*-\mathbf{x}_n)^T}{d_n}\\
        \end{bmatrix}\\
        & d_i = \|\mathbf{x}_*-\mathbf{x}_i\|_2 \text{ for $i=1,2,...,n$} \\
    \end{split}
\end{equation*}
The surface normal is just a normalized gradient of GPDF: $\frac{\nabla d_{\mathbf{x}_*}}{\|\nabla d_{\mathbf{x}_*}\|_2}$. The surface curvature needs a hessian of GPDF to be calculated. The hessian of GPDF can be obtained as below:
\begin{equation*}
    \begin{split}
        & \nabla^2 d_{\mathbf{x}_*} = \frac{d^2 r(\hat{o}(\mathbf{x}_*))}{d \hat{o}(\mathbf{x}_*)^2} \cdot \frac{\partial \hat{o}(\mathbf{x}_*)}{\partial \mathbf{x}_*}^T \frac{\partial \hat{o}(\mathbf{x}_*)}{\partial \mathbf{x}_*}\\
        & + \frac{d r(\hat{o}(\mathbf{x}_*))}{d \hat{o}(\mathbf{x}_*)} \cdot \frac{\partial^2 k(\mathbf{x}_*,\mathbf{X})}{\partial\mathbf{x}_*\partial\mathbf{x}_*}(K(\mathbf{X},\mathbf{X}) + diag(\Delta_{\Bar{f}}\mathbf{\Sigma}_x\Delta_{\Bar{f}}^T))^{-1}\mathbf{y} \\
        & \frac{d^2 r(\hat{o}(\mathbf{x}_*))}{d \hat{o}(\mathbf{x}_*)^2} = \frac{l}{\hat{o}^2} \\
        & \frac{\partial^2 k(\mathbf{x}_*,\mathbf{X})}{\partial\mathbf{x}_*\partial\mathbf{x}_*} = 
        \begin{bmatrix}
            \frac{1}{d_1 l}\exp\left(-\frac{d_1}{l}\right)\left(\left(\frac{1}{d_1 l}+\frac{1}{d_1^2}\right)h_1-I\right) \\ \vdots \\ \frac{1}{d_n l}\exp\left(-\frac{d_n}{l}\right)\left(\left(\frac{1}{d_n l}+\frac{1}{d_n^2}\right)h_n-I\right)\\
        \end{bmatrix}^T\\
        & h_i = (\mathbf{x}_*-\mathbf{x}_i)^T(\mathbf{x}_*-\mathbf{x}_i) \text{ for $i=1,2,...,n$}\\
    \end{split}
\end{equation*}
While the computation of the Hessian involves high-dimensional calculations, optimizing efficiency becomes possible when concurrently computing distance, gradient, and Hessian. This approach enables the elimination of several repetitive calculations, resulting in overall computational savings. The curvature can be obtained as below:
\begin{equation*}
    \begin{split}
        & \text{Mean curvature: }\frac{\nabla d_{\mathbf{x}_*}\nabla^2 d_{\mathbf{x}_*}\nabla d_{\mathbf{x}_*}^T-|\nabla d_{\mathbf{x}_*}|^2 \text{Tr}(\nabla^2 d_{\mathbf{x}_*})}{2|\nabla d_{\mathbf{x}_*}|^3}\\
        & \text{Gaussian curvature: }
        \frac{\begin{vmatrix}
            \nabla^2 d_{\mathbf{x}_*} & \nabla d_{\mathbf{x}_*}^T \\
            \nabla d_{\mathbf{x}_*} & 0
        \end{vmatrix}}{|\nabla d_{\mathbf{x}_*}|^4} \\
    \end{split}
\end{equation*}

\chapter{Interpolation parameter}\label{interpolation_parameter}
We illustrate the impact of different interpolation parameters, denoted as $l$, on the reconstruction process. In Figure \ref{fig:GPDF_results_whole}, we present the outcomes obtained from the complete point cloud comprising 500 points. Conversely, Figure \ref{fig:GPDF_results_partial} showcases the results generated from a partial point cloud, obtained by vertically halving the object and removing its backside. In both figures, the upper left segment displays the uncertainty measurement of the surface, while the upper right segment presents the depth map of the image, scaled in meters. The lower left section exhibits the surface normals, whereas the lower right section displays the surface colors.

\begin{figure}
     \centering
     \begin{subfigure}[b]{0.49\textwidth}
         \centering
         \includegraphics[width=\textwidth]{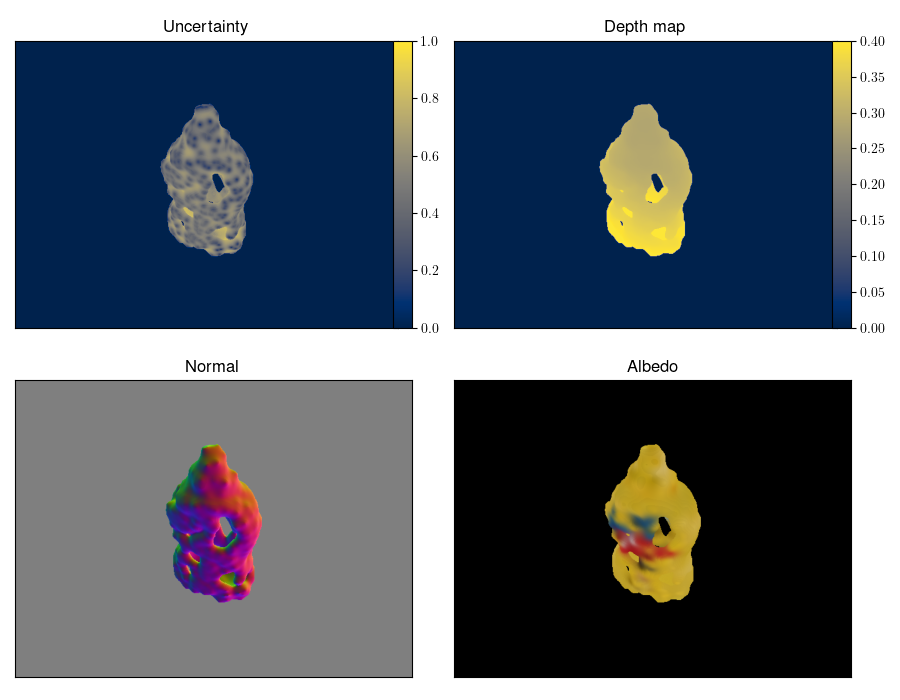}
         \caption{$l=0.01$ (front)}
     \end{subfigure}
     \begin{subfigure}[b]{0.49\textwidth}
         \centering
         \includegraphics[width=\textwidth]{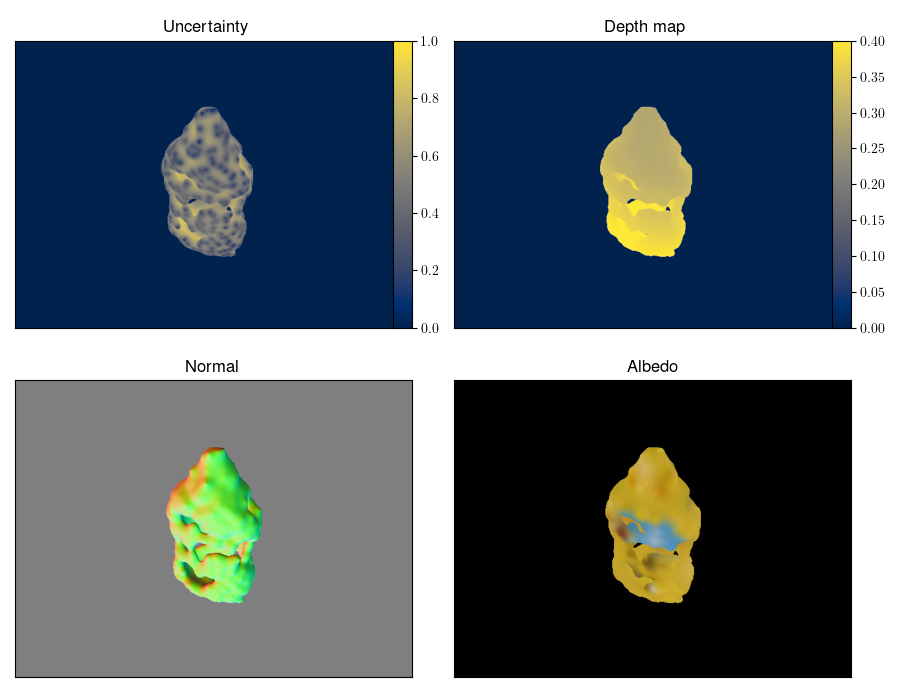}
         \caption{$l=0.01$ (back)}
     \end{subfigure}
\end{figure}
\begin{figure}\ContinuedFloat
     \centering
     \begin{subfigure}[b]{0.49\textwidth}
         \centering
         \includegraphics[width=\textwidth]{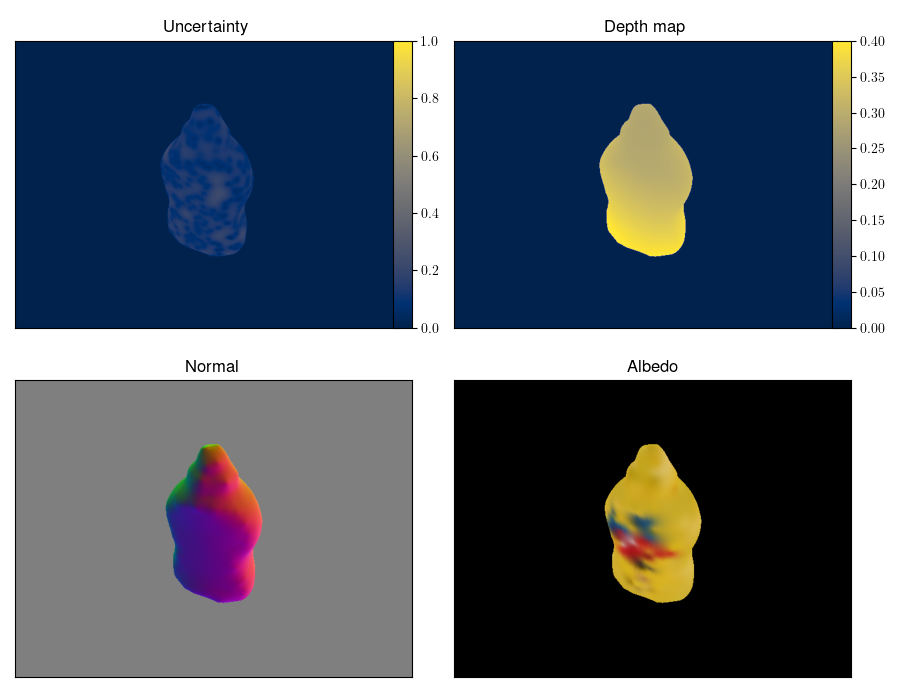}
         \caption{$l=0.05$ (front)}
     \end{subfigure}
     \begin{subfigure}[b]{0.49\textwidth}
         \centering
         \includegraphics[width=\textwidth]{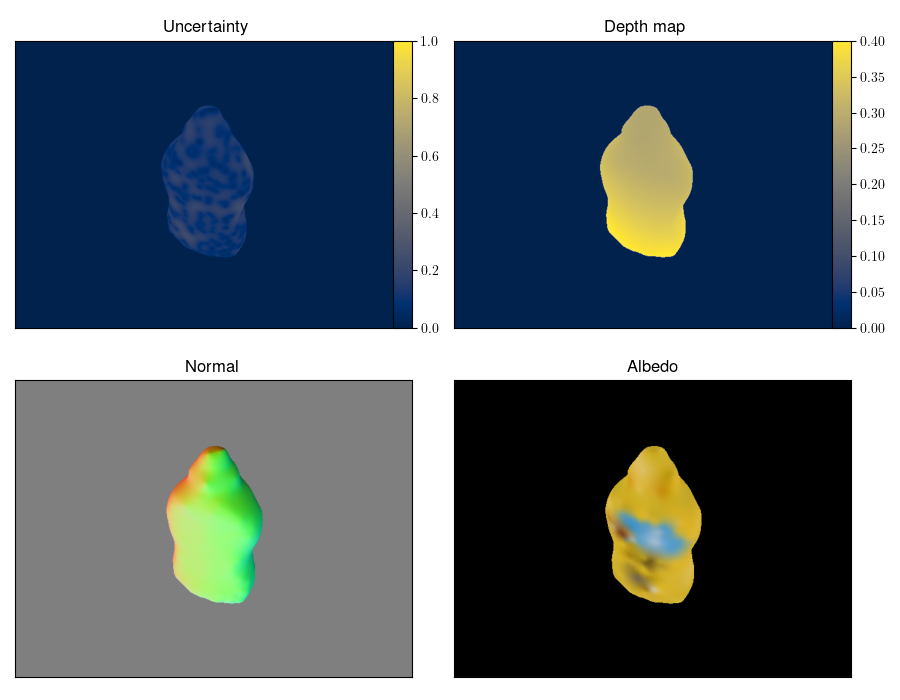}
         \caption{$l=0.05$ (back)}
     \end{subfigure}
\end{figure}
\begin{figure}\ContinuedFloat
     \centering
     \begin{subfigure}[b]{0.49\textwidth}
         \centering
         \includegraphics[width=\textwidth]{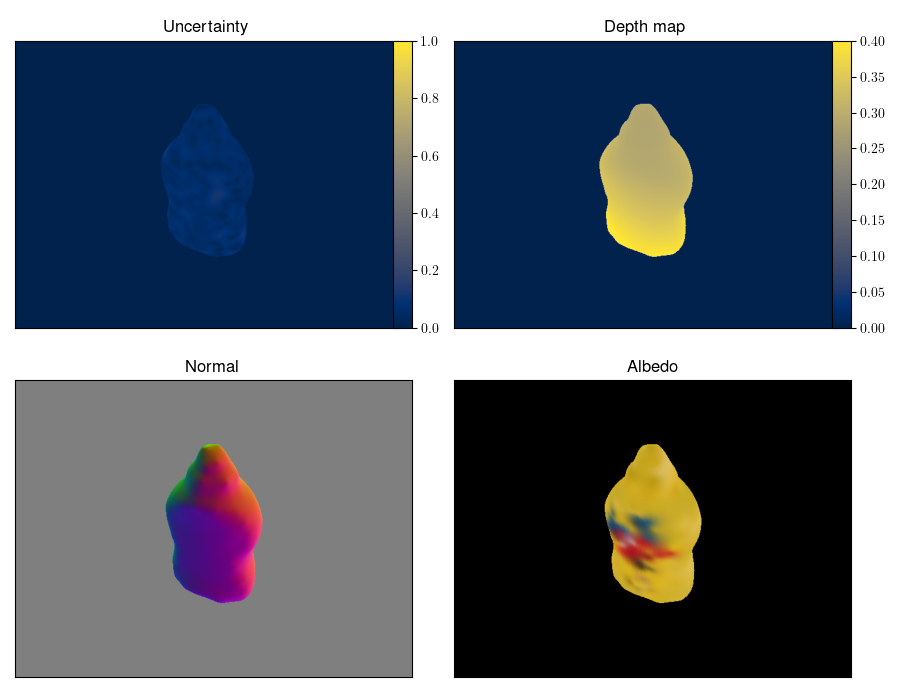}
         \caption{$l=0.10$ (front)}
     \end{subfigure}
     \begin{subfigure}[b]{0.49\textwidth}
         \centering
         \includegraphics[width=\textwidth]{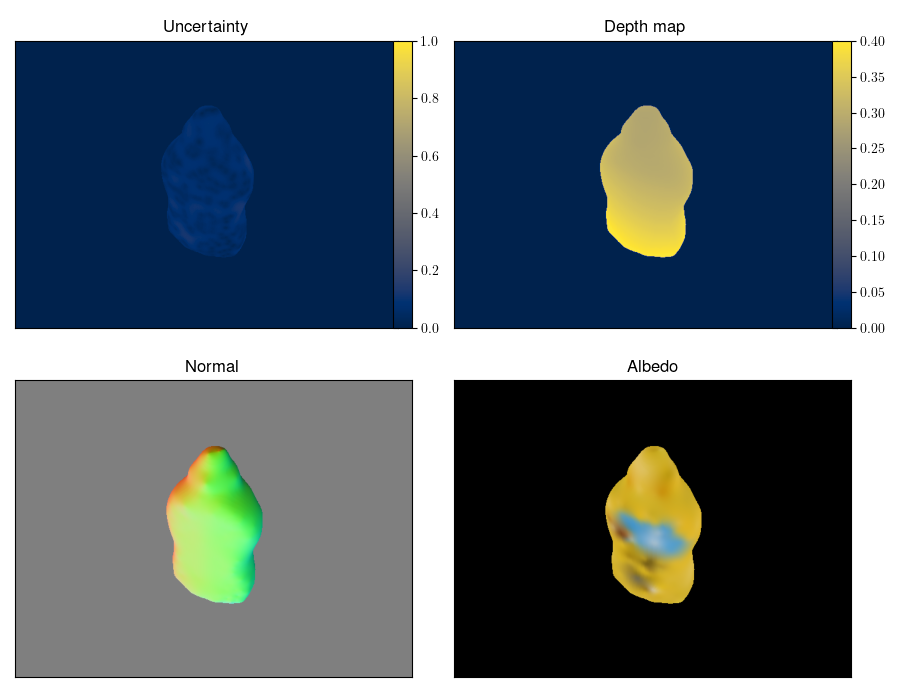}
         \caption{$l=0.10$ (back)}
     \end{subfigure}
\end{figure}
\begin{figure}\ContinuedFloat
     \centering
     \begin{subfigure}[b]{0.49\textwidth}
         \centering
         \includegraphics[width=\textwidth]{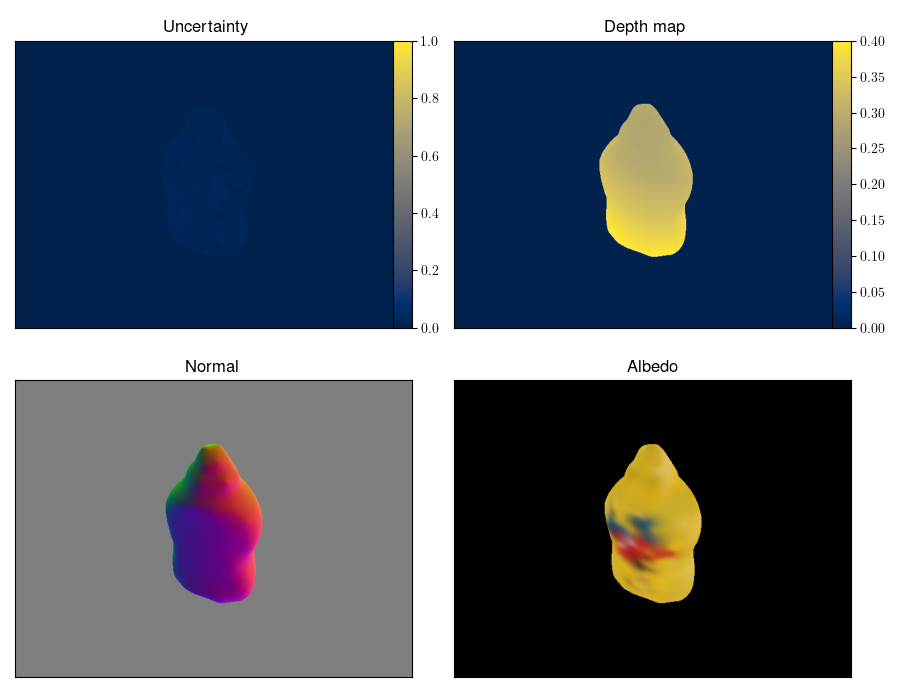}
         \caption{$l=0.30$ (front)}
     \end{subfigure}
     \begin{subfigure}[b]{0.49\textwidth}
         \centering
         \includegraphics[width=\textwidth]{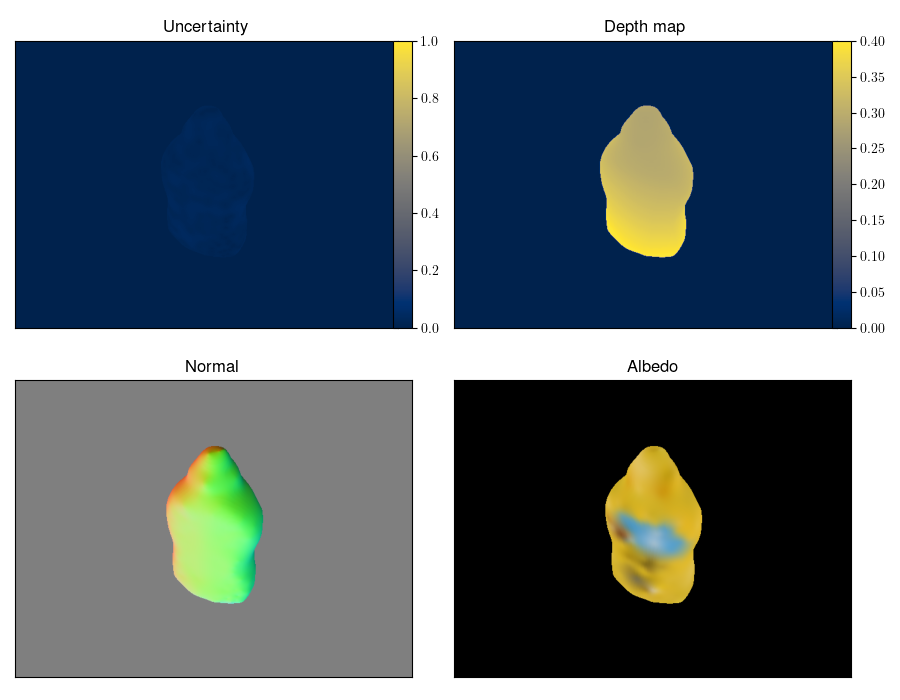}
         \caption{$l=0.30$ (back)}
     \end{subfigure}
     \caption{GPDF results for the whole point cloud of an object}
     \label{fig:GPDF_results_whole}
\end{figure}

\begin{figure}
     \centering
     \begin{subfigure}[b]{0.49\textwidth}
         \centering
         \includegraphics[width=\textwidth]{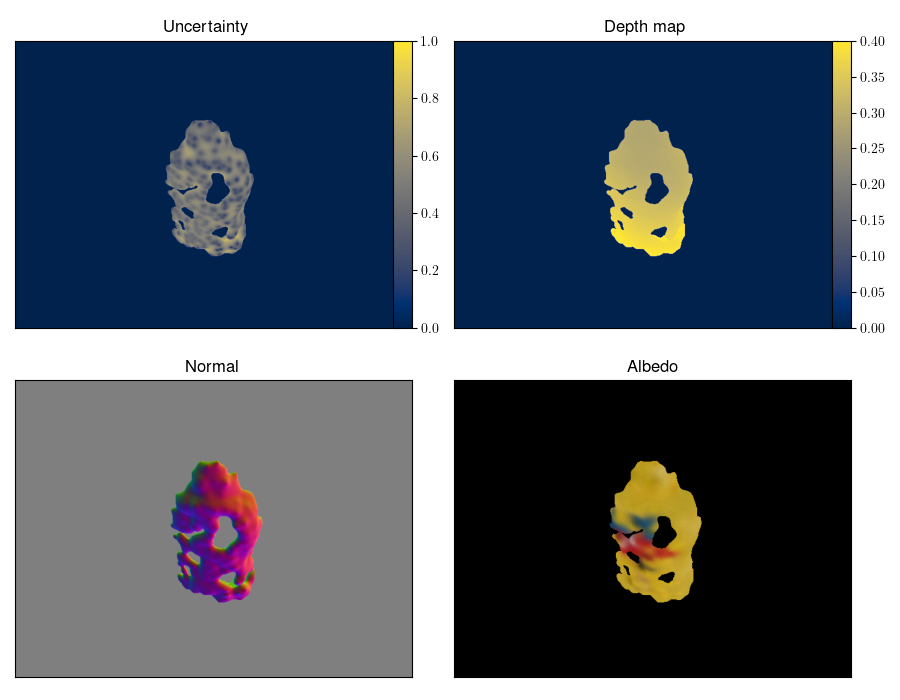}
         \caption{$l=0.01$ (front)}
     \end{subfigure}
     \begin{subfigure}[b]{0.49\textwidth}
         \centering
         \includegraphics[width=\textwidth]{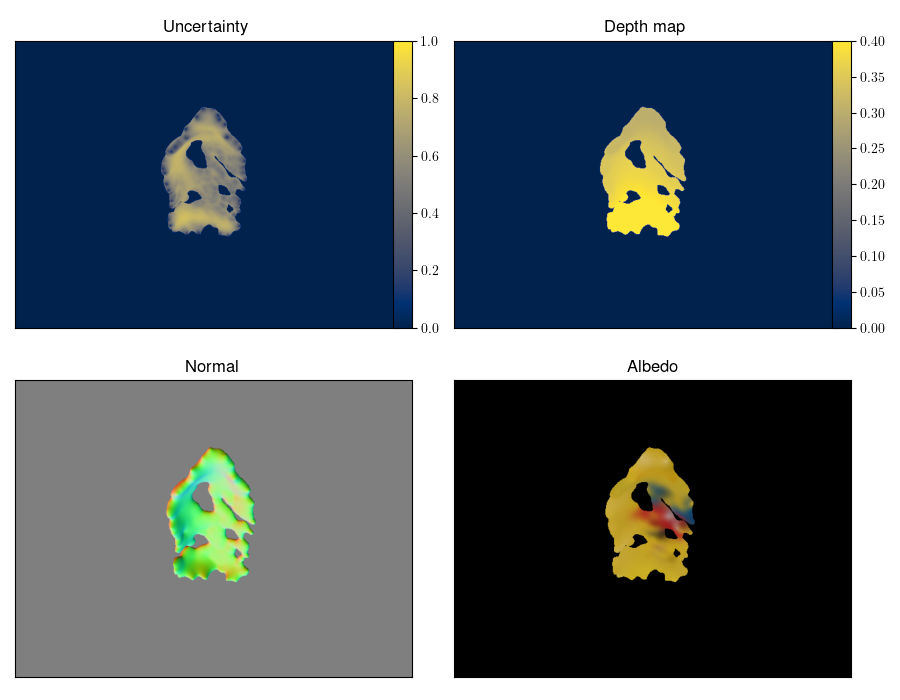}
         \caption{$l=0.01$ (back)}
     \end{subfigure}
\end{figure}
\begin{figure}\ContinuedFloat
     \centering
     \begin{subfigure}[b]{0.49\textwidth}
         \centering
         \includegraphics[width=\textwidth]{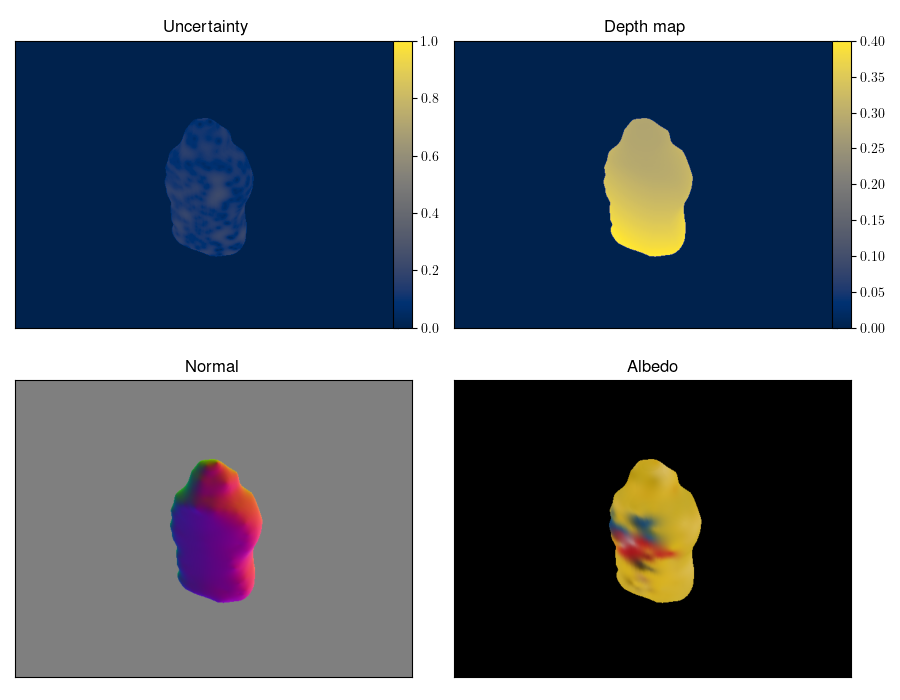}
         \caption{$l=0.05$ (front)}
     \end{subfigure}
     \begin{subfigure}[b]{0.49\textwidth}
         \centering
         \includegraphics[width=\textwidth]{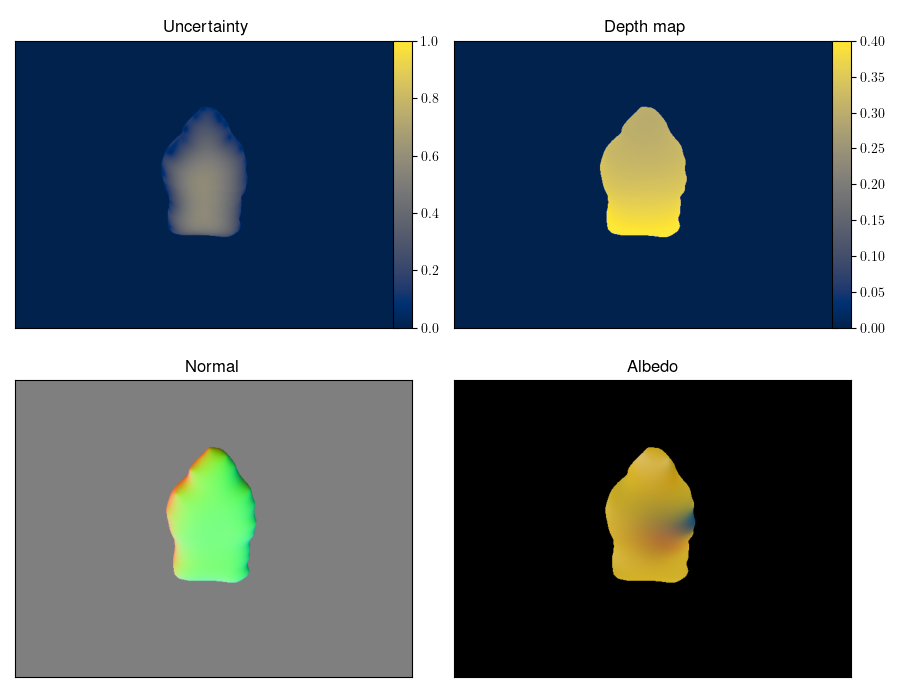}
         \caption{$l=0.05$ (back)}
     \end{subfigure}
\end{figure}
\begin{figure}\ContinuedFloat
     \centering
     \begin{subfigure}[b]{0.49\textwidth}
         \centering
         \includegraphics[width=\textwidth]{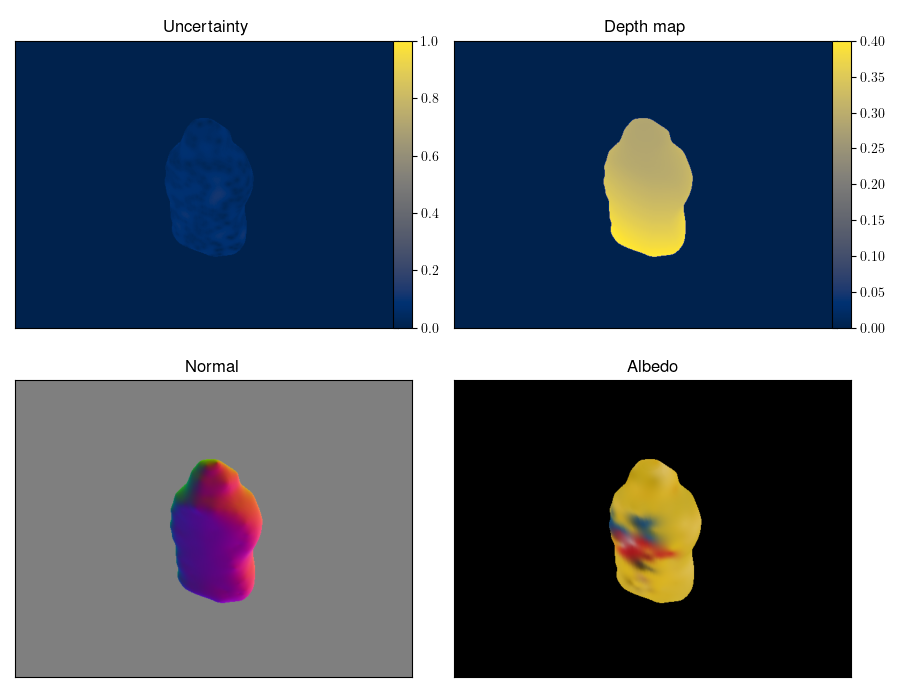}
         \caption{$l=0.10$ (front)}
     \end{subfigure}
     \begin{subfigure}[b]{0.49\textwidth}
         \centering
         \includegraphics[width=\textwidth]{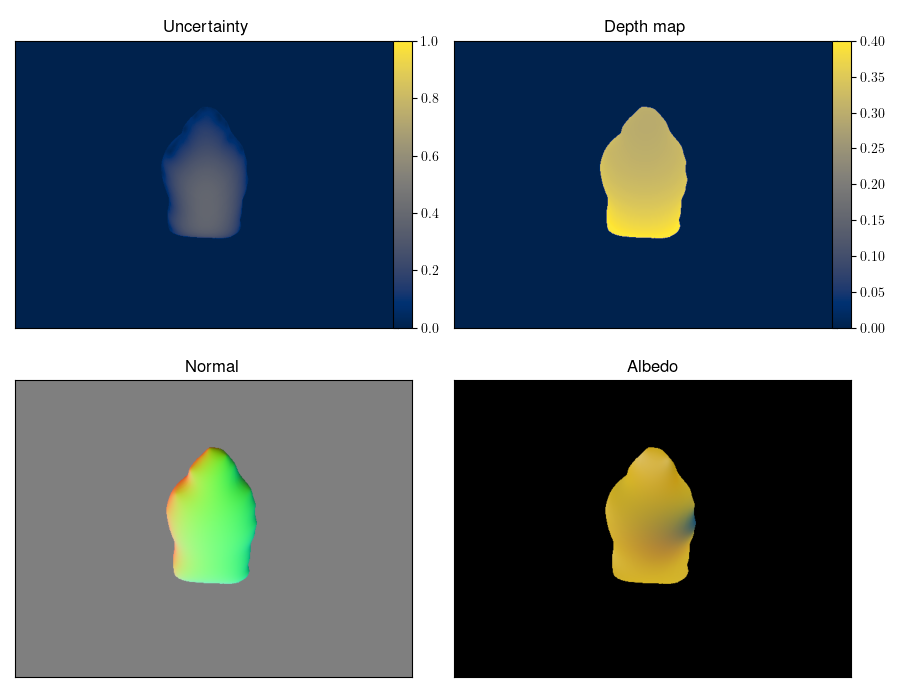}
         \caption{$l=0.10$ (back)}
     \end{subfigure}
\end{figure}
\begin{figure}\ContinuedFloat
     \centering
     \begin{subfigure}[b]{0.49\textwidth}
         \centering
         \includegraphics[width=\textwidth]{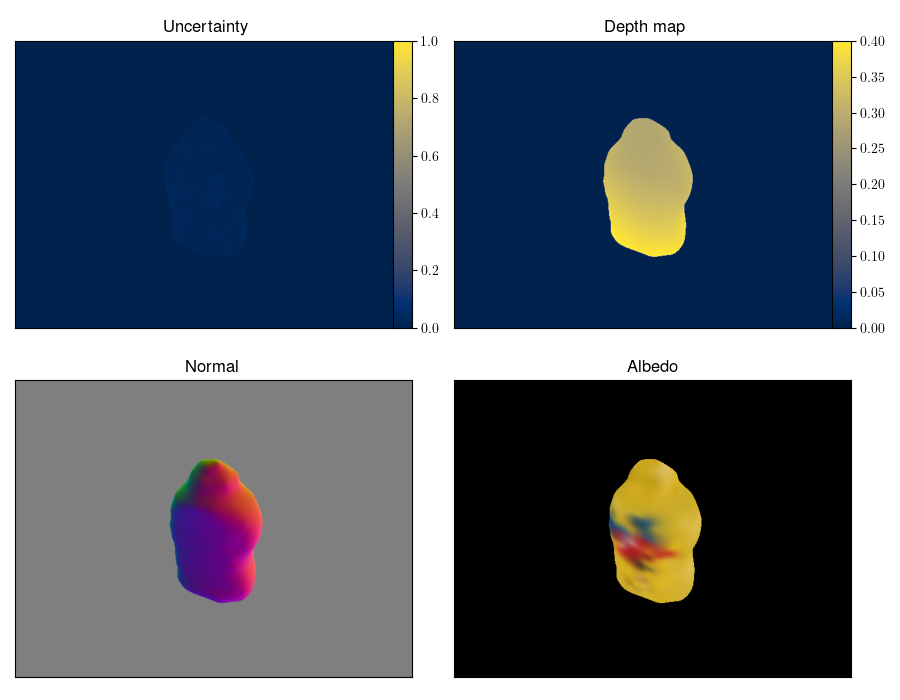}
         \caption{$l=0.30$ (front)}
     \end{subfigure}
     \begin{subfigure}[b]{0.49\textwidth}
         \centering
         \includegraphics[width=\textwidth]{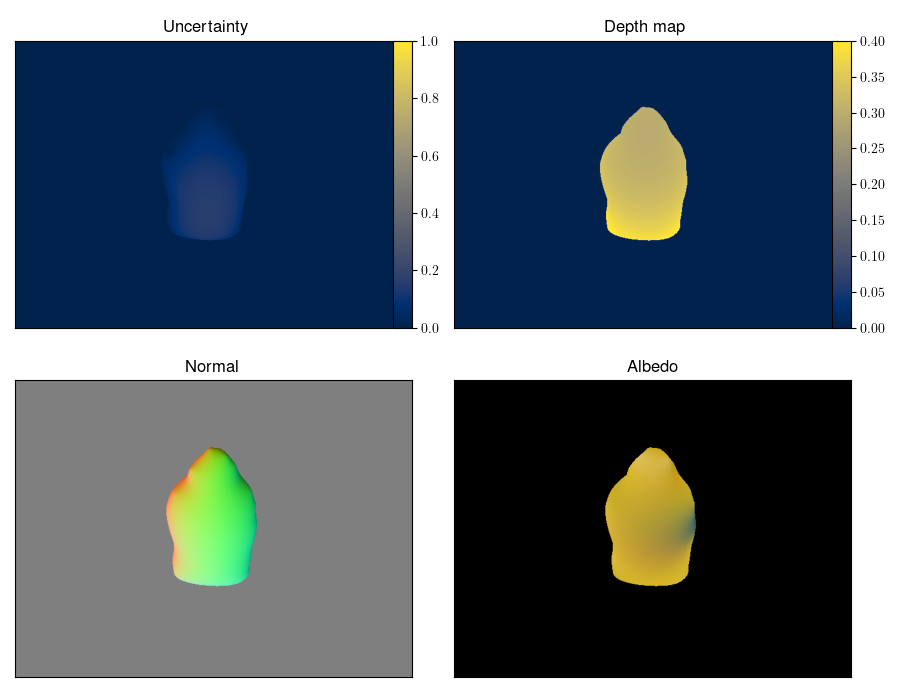}
         \caption{$l=0.30$ (back)}
     \end{subfigure}
     \caption{GPDF results for the partial point cloud of an object}
     \label{fig:GPDF_results_partial}
\end{figure}

\begin{singlespace}
\bibliography{biblio}

\begin{thebibliography}{10}

\bibitem{Gentil2023AccurateGP}
C.~L. Gentil, O.-L. Ouabi, L.~Wu, C.~Pradalier, and T.~Vidal-Calleja, ``Accurate gaussian process distance fields with applications to echolocation and mapping,'' {\em ArXiv}, vol.~abs/2302.13005, 2023.

\bibitem{Ounpraseuth2008GaussianPF}
S.~T. Ounpraseuth, ``Gaussian processes for machine learning,'' {\em Journal of the American Statistical Association}, vol.~103, pp.~429 -- 429, 2008.

\bibitem{Smith2005TheDO}
L.~B. Smith and M.~Gasser, ``The development of embodied cognition: Six lessons from babies,'' {\em Artificial Life}, vol.~11, pp.~13--29, 2005.

\bibitem{Mildenhall2020NeRFRS}
B.~Mildenhall, P.~P. Srinivasan, M.~Tancik, J.~T. Barron, R.~Ramamoorthi, and R.~Ng, ``Nerf: Representing scenes as neural radiance fields for view synthesis,'' {\em Commun. ACM}, vol.~65, pp.~99--106, 2020.

\bibitem{Lambeta2020DIGITAN}
M.~Lambeta, P.~wei Chou, S.~Tian, B.~Yang, B.~Maloon, V.~R. Most, D.~Stroud, R.~Santos, A.~Byagowi, G.~Kammerer, D.~Jayaraman, and R.~Calandra, ``Digit: A novel design for a low-cost compact high-resolution tactile sensor with application to in-hand manipulation,'' {\em IEEE Robotics and Automation Letters}, vol.~5, pp.~3838--3845, 2020.

\bibitem{Myronenko2009PointSR}
A.~Myronenko and X.~B. Song, ``Point set registration: Coherent point drift,'' {\em IEEE Transactions on Pattern Analysis and Machine Intelligence}, vol.~32, pp.~2262--2275, 2009.

\bibitem{Tewari2021AdvancesIN}
A.~Tewari, O.~Fried, J.~Thies, V.~Sitzmann, S.~Lombardi, Z.~Xu, T.~Simon, M.~Nie{\ss}ner, E.~Tretschk, L.~Liu, B.~Mildenhall, P.~Srinivasan, R.~Pandey, S.~Orts-Escolano, S.~Fanello, M.~G. Guo, G.~Wetzstein, J.~y~Zhu, C.~Theobalt, M.~Agrawala, D.~B. Goldman, and M.~Zollh{\"o}fer, ``Advances in neural rendering,'' {\em Computer Graphics Forum}, vol.~41, 2021.

\bibitem{Jakli2000SuperquadricsAT}
A.~Jaklič, A.~Leonardis, and F.~Solina, ``Superquadrics and their geometric properties,'' 2000.

\bibitem{iq_2021}
{iq}, ``Capsule - closest 3d.'' \url{https://www.shadertoy.com/view/stXXzl}, Jul 2021.
\newblock [Online; accessed 9-April-2024].

\bibitem{Choi2023TowardsFD}
H.~J. Choi and N.~Figueroa, ``Towards feasible dynamic grasping: Leveraging gaussian process distance field, se(3) equivariance and riemannian mixture models,'' {\em ArXiv}, vol.~abs/2311.02576, 2023.

\bibitem{Sundaralingam2023CuRoboPC}
B.~Sundaralingam, S.~K.~S. Hari, A.~Fishman, C.~R. Garrett, K.~V. Wyk, V.~Blukis, A.~Millane, H.~Oleynikova, A.~Handa, F.~Ramos, N.~Ratliff, and D.~Fox, ``Curobo: Parallelized collision-free minimum-jerk robot motion generation,'' {\em ArXiv}, vol.~abs/2310.17274, 2023.

\bibitem{Lederman1987HandMA}
S.~J. Lederman and R.~L. Klatzky, ``Hand movements: A window into haptic object recognition,'' {\em Cognitive Psychology}, vol.~19, pp.~342--368, 1987.

\bibitem{DBLP:journals/corr/BajcsyAT16}
R.~Bajcsy, Y.~Aloimonos, and J.~K. Tsotsos, ``Revisiting active perception,'' {\em CoRR}, vol.~abs/1603.02729, 2016.

\bibitem{Bajcsy1988ActiveP}
R.~Bajcsy, ``Active perception,'' {\em Proc. IEEE}, vol.~76, pp.~966--1005, 1988.

\bibitem{Allen1985ObjectRU}
P.~K. Allen and R.~Bajcsy, ``Object recognition using vision and touch,'' in {\em International Joint Conference on Artificial Intelligence}, 1985.

\bibitem{DBLP:journals/corr/abs-1912-00280}
M.~Liu, L.~Sheng, S.~Yang, J.~Shao, and S.~Hu, ``Morphing and sampling network for dense point cloud completion,'' {\em CoRR}, vol.~abs/1912.00280, 2019.

\bibitem{DBLP:journals/corr/abs-1712-07262}
Y.~Yang, C.~Feng, Y.~Shen, and D.~Tian, ``Foldingnet: Interpretable unsupervised learning on 3d point clouds,'' {\em CoRR}, vol.~abs/1712.07262, 2017.

\bibitem{DBLP:journals/corr/abs-1808-00671}
W.~Yuan, T.~Khot, D.~Held, C.~Mertz, and M.~Hebert, ``{PCN:} point completion network,'' {\em CoRR}, vol.~abs/1808.00671, 2018.

\bibitem{Ortiz2022iSDFRN}
J.~Ortiz, A.~Clegg, J.~Dong, E.~Sucar, D.~Novotn{\'y}, M.~Zollhoefer, and M.~Mukadam, ``isdf: Real-time neural signed distance fields for robot perception,'' {\em ArXiv}, vol.~abs/2204.02296, 2022.

\bibitem{Kerbl20233DGS}
B.~Kerbl, G.~Kopanas, T.~Leimkuehler, and G.~Drettakis, ``3d gaussian splatting for real-time radiance field rendering,'' {\em ACM Transactions on Graphics (TOG)}, vol.~42, pp.~1 -- 14, 2023.

\bibitem{Daniilidis1999HandEyeCU}
K.~Daniilidis, ``Hand-eye calibration using dual quaternions,'' {\em The International Journal of Robotics Research}, vol.~18, pp.~286 -- 298, 1999.

\bibitem{DBLP:journals/corr/abs-2010-11487}
L.~Wu, K.~M.~B. Lee, L.~Liu, and T.~A. Vidal{-}Calleja, ``Faithful euclidean distance field from log-gaussian process implicit surfaces,'' {\em CoRR}, vol.~abs/2010.11487, 2020.

\bibitem{lee2022uncertainty}
S.~Lee, C.~Le, W.~Jiahao, A.~Liniger, S.~Kumar, and F.~Yu, ``Uncertainty guided policy for active robotic 3d reconstruction using neural radiance fields,'' {\em IEEE Robotics and Automation Letters}, 2022.

\bibitem{Pan2022ActiveNeRFLW}
X.~Pan, Z.~Lai, S.~Song, and G.~Huang, ``Activenerf: Learning where to see with uncertainty estimation,'' in {\em European Conference on Computer Vision}, 2022.

\bibitem{Jin2023NeUNBVNB}
L.~Jin, X.~Chen, J.~Ruckin, and M.~Popovi'c, ``Neu-nbv: Next best view planning using uncertainty estimation in image-based neural rendering,'' {\em 2023 IEEE/RSJ International Conference on Intelligent Robots and Systems (IROS)}, pp.~11305--11312, 2023.

\bibitem{Zhan2022ActiveRMAPRF}
H.~Zhan, J.~Zheng, Y.~Xu, I.~D. Reid, and H.~Rezatofighi, ``Activermap: Radiance field for active mapping and planning,'' {\em ArXiv}, vol.~abs/2211.12656, 2022.

\bibitem{Jiang2023FisherRFAV}
W.~Jiang, B.~Lei, and K.~Daniilidis, ``Fisherrf: Active view selection and uncertainty quantification for radiance fields using fisher information,'' {\em ArXiv}, vol.~abs/2311.17874, 2023.

\bibitem{He2023ActivePU}
S.~He, C.~D. Hsu, D.~Ong, Y.~Shao, and P.~Chaudhari, ``Active perception using neural radiance fields,'' {\em ArXiv}, vol.~abs/2310.09892, 2023.

\bibitem{DBLP:journals/corr/abs-2108-00737}
E.~Safronov, N.~A. Piga, M.~Colledanchise, and L.~Natale, ``Active perception for ambiguous objects classification,'' {\em CoRR}, vol.~abs/2108.00737, 2021.

\bibitem{Hausman2012SegmentationOC}
K.~Hausman, C.~Bersch, D.~Pangercic, S.~Osentoski, Z.-C. M{\'a}rton, and M.~Beetz, ``Segmentation of cluttered scenes through interactive perception,'' 2012.

\bibitem{Radford2021LearningTV}
A.~Radford, J.~W. Kim, C.~Hallacy, A.~Ramesh, G.~Goh, S.~Agarwal, G.~Sastry, A.~Askell, P.~Mishkin, J.~Clark, G.~Krueger, and I.~Sutskever, ``Learning transferable visual models from natural language supervision,'' in {\em International Conference on Machine Learning}, 2021.

\bibitem{Shirai2023TactileTM}
Y.~Shirai, D.~K. Jha, A.~U. Raghunathan, and D.~W. Hong, ``Tactile tool manipulation /author=shirai, yuki; jha, devesh k.; raghunathan, arvind; hong, dennis /creationdate=may 2, 2023 /subject=robotics,'' 2023.

\bibitem{Hogan2020TactileDM}
F.~R. Hogan, J.~Ballester, S.~Dong, and A.~Rodriguez, ``Tactile dexterity: Manipulation primitives with tactile feedback,'' {\em 2020 IEEE International Conference on Robotics and Automation (ICRA)}, pp.~8863--8869, 2020.

\bibitem{She2019CableMW}
Y.~She, S.~Wang, S.~Dong, N.~Sunil, A.~Rodriguez, and E.~H. Adelson, ``Cable manipulation with a tactile-reactive gripper,'' {\em The International Journal of Robotics Research}, vol.~40, pp.~1385 -- 1401, 2019.

\bibitem{Ding2017TactilePO}
S.~Ding, Y.~Pan, M.~Tong, and X.~Zhao, ``Tactile perception of roughness and hardness to discriminate materials by friction-induced vibration,'' {\em Sensors (Basel, Switzerland)}, vol.~17, 2017.

\bibitem{Yuan2016EstimatingOH}
W.~Yuan, M.~A. Srinivasan, and E.~H. Adelson, ``Estimating object hardness with a gelsight touch sensor,'' {\em 2016 IEEE/RSJ International Conference on Intelligent Robots and Systems (IROS)}, pp.~208--215, 2016.

\bibitem{Cao2023MultimodalZL}
G.~Cao, J.~Jiang, D.~Bollegala, M.~Li, and S.~Luo, ``Multimodal zero-shot learning for tactile texture recognition,'' {\em ArXiv}, vol.~abs/2306.12705, 2023.

\bibitem{Aoyama2023FewShotLO}
M.~Y. Aoyama, J.~Moura, N.~Saito, and S.~Vijayakumar, ``Few-shot learning of force-based motions from demonstration through pre-training of haptic representation,'' {\em ArXiv}, vol.~abs/2309.04640, 2023.

\bibitem{Kappasov2015TactileSI}
Z.~Kappasov, J.~A. Corrales, and V.~Perdereau, ``Tactile sensing in dexterous robot hands - review,'' {\em Robotics Auton. Syst.}, vol.~74, pp.~195--220, 2015.

\bibitem{Roberts2021SoftTS}
P.~Roberts, M.~Zadan, and C.~Majidi, ``Soft tactile sensing skins for robotics,'' {\em Current Robotics Reports}, vol.~2, pp.~343 -- 354, 2021.

\bibitem{Yuan2017GelSightHR}
W.~Yuan, S.~Dong, and E.~H. Adelson, ``Gelsight: High-resolution robot tactile sensors for estimating geometry and force,'' {\em Sensors (Basel, Switzerland)}, vol.~17, 2017.

\bibitem{Taylor2021GelSlim3H}
I.~H. Taylor, S.~Dong, and A.~Rodriguez, ``Gelslim 3.0: High-resolution measurement of shape, force and slip in a compact tactile-sensing finger,'' {\em 2022 International Conference on Robotics and Automation (ICRA)}, pp.~10781--10787, 2021.

\bibitem{Lepora2021SoftBO}
N.~F. Lepora, ``Soft biomimetic optical tactile sensing with the tactip: A review,'' {\em IEEE Sensors Journal}, vol.~21, pp.~21131--21143, 2021.

\bibitem{Yin2022MultimodalPA}
J.~Yin, G.~M. Campbell, J.~H. Pikul, and M.~H. Yim, ``Multimodal proximity and visuotactile sensing with a selectively transmissive soft membrane,'' {\em 2022 IEEE 5th International Conference on Soft Robotics (RoboSoft)}, pp.~802--808, 2022.

\bibitem{XELA_Robotics_2018}
{XELA Robotics}, ``Tactile sensors.'' \url{https://www.xelarobotics.com/tactile-sensors}, 2018.
\newblock [Online; accessed 9-April-2024].

\bibitem{Contactile_2019}
{Contactile}, ``Tactile sensors.'' \url{https://contactile.com/products/#tactile_sensors}, 2019.
\newblock [Online; accessed 9-April-2024].

\bibitem{Masterjohn2021VelocityLA}
J.~G. Masterjohn, D.~Guoy, J.~Shepherd, and A.~M. Castro, ``Velocity level approximation of pressure field contact patches,'' {\em IEEE Robotics and Automation Letters}, vol.~7, pp.~11593--11600, 2021.

\bibitem{8793773}
D.~Driess, D.~Hennes, and M.~Toussaint, ``Active multi-contact continuous tactile exploration with gaussian process differential entropy,'' in {\em 2019 International Conference on Robotics and Automation (ICRA)}, pp.~7844--7850, May 2019.

\bibitem{KHADIVAR2023104461}
F.~Khadivar, K.~Yao, X.~Gao, and A.~Billard, ``Online active and dynamic object shape exploration with a multi-fingered robotic hand,'' {\em Robotics and Autonomous Systems}, vol.~166, p.~104461, 2023.

\bibitem{DBLP:journals/corr/abs-2103-00655}
C.~de~Farias, N.~Marturi, R.~Stolkin, and Y.~Bekiroglu, ``Simultaneous tactile exploration and grasp refinement for unknown objects,'' {\em CoRR}, vol.~abs/2103.00655, 2021.

\bibitem{DBLP:journals/corr/abs-2109-09884}
S.~Suresh, Z.~Si, J.~G. Mangelson, W.~Yuan, and M.~Kaess, ``Efficient shape mapping through dense touch and vision,'' {\em CoRR}, vol.~abs/2109.09884, 2021.

\bibitem{DBLP:journals/corr/abs-2011-07044}
S.~Suresh, M.~Bauz{\'{a}}, K.~Yu, J.~G. Mangelson, A.~Rodriguez, and M.~Kaess, ``Tactile {SLAM:} real-time inference of shape and pose from planar pushing,'' {\em CoRR}, vol.~abs/2011.07044, 2020.

\bibitem{Suresh2023NeuralFW}
S.~Suresh, H.~Qi, T.~Wu, T.~Fan, L.~Pineda, M.~Lambeta, J.~Malik, M.~Kalakrishnan, R.~Calandra, M.~Kaess, J.~Ortiz, and M.~Mukadam, ``Neural feels with neural fields: Visuo-tactile perception for in-hand manipulation,'' 2023.

\bibitem{Swann2024TouchGSVS}
A.~Swann, M.~Strong, W.~K. Do, G.~S. Camps, M.~Schwager, and M.~Kennedy, ``Touch-gs: Visual-tactile supervised 3d gaussian splatting,'' 2024.

\bibitem{Caccamo2016ActivePA}
S.~Caccamo, P.~G{\"u}ler, H.~Kjellstr{\"o}m, and D.~Kragic, ``Active perception and modeling of deformable surfaces using gaussian processes and position-based dynamics,'' {\em 2016 IEEE-RAS 16th International Conference on Humanoid Robots (Humanoids)}, pp.~530--537, 2016.

\bibitem{Wang2021NeuSLN}
P.~Wang, L.~Liu, Y.~Liu, C.~Theobalt, T.~Komura, and W.~Wang, ``Neus: Learning neural implicit surfaces by volume rendering for multi-view reconstruction,'' {\em ArXiv}, vol.~abs/2106.10689, 2021.

\bibitem{8768489}
M.~S. Ahn, H.~Chae, D.~Noh, H.~Nam, and D.~Hong, ``Analysis and noise modeling of the intel realsense d435 for mobile robots,'' in {\em 2019 16th International Conference on Ubiquitous Robots (UR)}, pp.~707--711, 2019.

\bibitem{Tsai1988ANT}
R.~Y. Tsai and R.~K. Lenz, ``A new technique for fully autonomous and efficient 3d robotics hand/eye calibration,'' {\em IEEE Trans. Robotics Autom.}, vol.~5, pp.~345--358, 1988.

\bibitem{Smooth-On}
{Smooth-On}, ``Products.'' \url{https://www.smooth-on.com/products/}.
\newblock [Online; accessed 9-April-2024].

\bibitem{Azulay2023AllSightAL}
O.~Azulay, N.~Curtis, R.~Sokolovsky, G.~Levitski, D.~Slomovik, G.~Lilling, and A.~Sintov, ``Allsight: A low-cost and high-resolution round tactile sensor with zero-shot learning capability,'' {\em IEEE Robotics and Automation Letters}, vol.~9, pp.~483--490, 2023.

\bibitem{Kuhn1955TheHM}
H.~W. Kuhn, ``The hungarian method for the assignment problem,'' {\em Naval Research Logistics (NRL)}, vol.~52, 1955.

\bibitem{Yuille1989AMA}
A.~L. Yuille and N.~M. Grzywacz, ``A mathematical analysis of the motion coherence theory,'' {\em International Journal of Computer Vision}, vol.~3, pp.~155--175, 1989.

\bibitem{Kaslin2018TowardsAP}
R.~Kaslin, H.~Kolvenbach, L.~Paez, K.~Lika, and M.~Hutter, ``Towards a passive adaptive planar foot with ground orientation and contact force sensing for legged robots,'' {\em 2018 IEEE/RSJ International Conference on Intelligent Robots and Systems (IROS)}, pp.~2707--2714, 2018.

\bibitem{Tracy2022DifferentiableCD}
K.~Tracy, T.~A. Howell, and Z.~Manchester, ``Differentiable collision detection for a set of convex primitives,'' {\em 2023 IEEE International Conference on Robotics and Automation (ICRA)}, pp.~3663--3670, 2022.

\bibitem{Narang2022FactoryFC}
Y.~S. Narang, K.~Storey, I.~Akinola, M.~Macklin, P.~Reist, L.~Wawrzyniak, Y.~Guo, {\'A}.~Morav{\'a}nszky, G.~State, M.~Lu, A.~Handa, and D.~Fox, ``Factory: Fast contact for robotic assembly,'' {\em ArXiv}, vol.~abs/2205.03532, 2022.

\bibitem{Li2023RepresentingRG}
Y.~Li, Y.~Zhang, A.~Razmjoo, and S.~Calinon, ``Representing robot geometry as distance fields: Applications to whole-body manipulation,'' 2023.

\bibitem{Koptev2023NeuralJS}
M.~Koptev, N.~Figueroa, and A.~Billard, ``Neural joint space implicit signed distance functions for reactive robot manipulator control,'' {\em IEEE Robotics and Automation Letters}, vol.~8, pp.~480--487, 2023.

\bibitem{Zhang2023ContinuousIS}
T.~Zhang, J.~Wang, C.~Xu, A.~Gao, and F.~Gao, ``Continuous implicit sdf based any-shape robot trajectory optimization,'' {\em 2023 IEEE/RSJ International Conference on Intelligent Robots and Systems (IROS)}, pp.~282--289, 2023.

\bibitem{Bajcsy1987ThreeDO}
R.~Bajcsy and F.~Solina, ``Three dimensional object representation revisited,'' 1987.

\bibitem{Liu2021RobustAA}
W.~Liu, Y.~Wu, S.~Ruan, and G.~S. Chirikjian, ``Robust and accurate superquadric recovery: a probabilistic approach,'' {\em 2022 IEEE/CVF Conference on Computer Vision and Pattern Recognition (CVPR)}, pp.~2666--2675, 2021.

\bibitem{Merwe2019LearningC3}
M.~V. der Merwe, Q.~Lu, B.~Sundaralingam, M.~Matak, and T.~Hermans, ``Learning continuous 3d reconstructions for geometrically aware grasping,'' {\em 2020 IEEE International Conference on Robotics and Automation (ICRA)}, pp.~11516--11522, 2019.

\bibitem{Chou2022GenSDFTL}
G.~Chou, I.~Chugunov, and F.~Heide, ``Gensdf: Two-stage learning of generalizable signed distance functions,'' {\em ArXiv}, vol.~abs/2206.02780, 2022.

\bibitem{Chibane2020NeuralUD}
J.~Chibane, A.~Mir, and G.~Pons-Moll, ``Neural unsigned distance fields for implicit function learning,'' {\em ArXiv}, vol.~abs/2010.13938, 2020.

\bibitem{Seeger2004GaussianPF}
M.~W. Seeger, ``Gaussian processes for machine learning,'' {\em International journal of neural systems}, vol.~14 2, pp.~69--106, 2004.

\bibitem{williams2007gaussian}
O.~Williams and A.~Fitzgibbon, ``Gaussian process implicit surfaces,'' in {\em Gaussian Processes in Practice}, April 2007.

\bibitem{NIPS2011_a8e864d0}
A.~Mchutchon and C.~Rasmussen, ``Gaussian process training with input noise,'' in {\em Advances in Neural Information Processing Systems} (J.~Shawe-Taylor, R.~Zemel, P.~Bartlett, F.~Pereira, and K.~Weinberger, eds.), vol.~24, Curran Associates, Inc., 2011.

\bibitem{Stanton2021KernelIF}
S.~Stanton, W.~J. Maddox, I.~A. Delbridge, and A.~G. Wilson, ``Kernel interpolation for scalable online gaussian processes,'' in {\em International Conference on Artificial Intelligence and Statistics}, 2021.

\bibitem{RiutortMayol2020PracticalHS}
G.~Riutort-Mayol, P.-C. Burkner, M.~R. Andersen, A.~Solin, and A.~Vehtari, ``Practical hilbert space approximate bayesian gaussian processes for probabilistic programming,'' {\em Statistics and Computing}, vol.~33, 2020.

\bibitem{Yuan2023UniFusionUC}
Y.~Yuan and A.~Nuechter, ``Uni-fusion: Universal continuous mapping,'' {\em ArXiv}, vol.~abs/2303.12678, 2023.

\bibitem{Galin2020SegmentTU}
E.~Galin, {\'E}.~Gu{\'e}rin, A.~Paris, and A.~Peytavie, ``Segment tracing using local lipschitz bounds,'' {\em Computer Graphics Forum}, vol.~39, 2020.

\bibitem{Aydinlilar2023ForwardIF}
M.~Aydinlilar and C.~Zanni, ``Forward inclusion functions for ray-tracing implicit surfaces,'' {\em Comput. Graph.}, vol.~114, pp.~190--200, 2023.

\bibitem{Shen2023DistilledFF}
B.~W. Shen, G.~Yang, A.~Yu, J.~R. Wong, L.~P. Kaelbling, and P.~Isola, ``Distilled feature fields enable few-shot language-guided manipulation,'' {\em ArXiv}, vol.~abs/2308.07931, 2023.

\bibitem{alli2017YaleCMUBerkeleyDF}
B.~Çalli, A.~Singh, J.~Bruce, A.~Walsman, K.~Konolige, S.~S. Srinivasa, P.~Abbeel, and A.~M. Dollar, ``Yale-cmu-berkeley dataset for robotic manipulation research,'' {\em The International Journal of Robotics Research}, vol.~36, pp.~261 -- 268, 2017.

\bibitem{Salehian2018AUF}
S.~S.~M. Salehian, N.~Figueroa, and A.~Billard, ``A unified framework for coordinated multi-arm motion planning,'' {\em The International Journal of Robotics Research}, vol.~37, pp.~1205 -- 1232, 2018.

\bibitem{Richardson2022LearningTF}
B.~A. Richardson, Y.~Vardar, C.~Wallraven, and K.~J. Kuchenbecker, ``Learning to feel textures: Predicting perceptual similarities from unconstrained finger-surface interactions,'' {\em IEEE Transactions on Haptics}, vol.~15, pp.~705--717, 2022.

\bibitem{Khojasteh2024RobustSR}
B.~Khojasteh, Y.~Shao, and K.~J. Kuchenbecker, ``Robust surface recognition with the maximum mean discrepancy: Degrading haptic-auditory signals through bandwidth and noise,'' {\em IEEE Transactions on Haptics}, vol.~17, pp.~58--65, 2024.

\bibitem{Borovitskiy2020MaternGP}
V.~Borovitskiy, A.~Terenin, P.~Mostowsky, and M.~P. Deisenroth, ``Matern gaussian processes on riemannian manifolds,'' {\em ArXiv}, vol.~abs/2006.10160, 2020.

\end{thebibliography}
\bibliographystyle{ieeetr}
\end{singlespace}

\end{document}